\pdfoutput=1
\documentclass[10pt,numbers]{cardpreprint}
\usepackage{pifont}
\usepackage{tikz}
\usetikzlibrary{positioning,fit,calc,arrows.meta,backgrounds}
\definecolor{archslate}{HTML}{8DA2BC}
\definecolor{archtrain}{HTML}{D5DEEA}
\definecolor{archfrozen}{HTML}{ECEDEF}
\definecolor{archloss}{HTML}{FFDDBB}
\definecolor{archink}{HTML}{33363B}
\definecolor{archgray}{HTML}{6E737B}
\tikzset{
  arch/.style={font=\footnotesize\sffamily, >={Stealth[length=4pt,width=3.5pt]}, line width=0.5pt, draw=archink, text=archink},
  data/.style={draw=archgray!70, fill=white, rounded corners=2pt, minimum height=5.5mm, inner xsep=3pt, align=center},
  module/.style={draw=archslate!90!black, fill=archtrain, rounded corners=2pt, minimum height=6.5mm, inner xsep=4pt, align=center},
  frozen/.style={module, fill=archfrozen, draw=archgray!80, dashed},
  loss/.style={draw=orange!60!black, fill=archloss, rounded corners=2pt, minimum height=5.5mm, inner xsep=3pt, align=center},
  op/.style={circle, draw=archink, fill=white, inner sep=0pt, minimum size=3.6mm, font=\scriptsize\sffamily},
  flow/.style={->, draw=archink},
  cond/.style={->, draw=archslate!80!black},
  dims/.style={font=\scriptsize\sffamily, text=archgray},
  stage/.style={font=\scriptsize\sffamily\scshape, text=archgray},
  group/.style={draw=archslate!60, fill=archslate!8, rounded corners=4pt, inner sep=4pt},
}

\newcommand{\proteus}{Proteus\xspace}

\definecolor{myred}{rgb}{0.8,0,0}
\definecolor{mygreen}{rgb}{0,0.8,0}
\newcommand{\good}{{\color{mygreen}\ding{51}}}
\newcommand{\bad}{{\color{myred}\ding{55}}}

\newcommand{\fst}[1]{\best{#1}}
\newcommand{\snd}[1]{\second{#1}}
\newcommand{\trd}[1]{\third{#1}}
\newcommand{\ffst}[1]{\swatch{cardbest}{\textbf{#1}}}
\newcommand{\fsnd}[1]{\swatch{cardsecond}{#1}}
\newcommand{\ftrd}[1]{\swatch{cardthird}{#1}}

\title{\proteus: A Scalable Data Engine for Template-Aligned Clothed Humans}

\author[1,2]{Anton Zubekhin}
\author[1,2]{Heming Zhu}
\author[1]{Briac Toussaint}
\author[3]{Adam Kortylewski}
\author[4]{Hugo Bertiche}
\author[4]{Thabo Beeler}
\author[1,2]{Marc Habermann}
\author[1,2]{Christian Theobalt}
\affiliation[1]{Max Planck Institute for Informatics, Saarland Informatics Campus}
\affiliation[2]{Saarbr\"ucken Research Center for Visual Computing, Interaction and Artificial Intelligence}
\affiliation[3]{University of Freiburg}
\affiliation[4]{Google}
\correspondence{\email{azubekhi@mpi-inf.mpg.de}}
\metadata[Project page]{\url{https://vcai.mpi-inf.mpg.de/projects/proteus/}}

\abstract{
Template-aligned clothed human geometry shares the topology and skeletal rig of a parametric body model, establishing dense surface correspondence across identities and making assets directly animatable.
This unified representation is well suited for learning clothed human reconstruction, registration, and body estimation models, yet obtaining such data at scale requires tedious and expensive human scanning followed by accurate non-rigid registration.
Recent image and 3D generative models can synthesize diverse clothed humans in large numbers, but their outputs have arbitrary topology and lack correspondence to an underlying body.
We show that, when carefully annotated and registered to a common body template, generated humans provide strong supervision for a wide range of downstream tasks.
To this end, we introduce \proteus, a scalable data engine that converts text descriptions into clothed human assets in a shared, high-resolution SMPL-X topology (SMPL-X-D).
\proteus generates full-body images from structured prompts, lifts them to textured 3D assets, annotates their outer surfaces with multi-view semantic labels and 3D body landmarks, and deforms the body template to each surface via coarse-to-fine non-rigid registration.
The resulting dataset pairs text, images, textured 3D assets, semantic annotations, and template-aligned clothed surfaces.
To validate this supervision, we train downstream models exclusively on \proteus and evaluate them on real-world scans: they generate SMPL-X-D meshes from images and from partial point clouds, acting as feed-forward registration, segment clothed point clouds, and estimate the SMPL-X body from point clouds and from dense template correspondences in images.
Across these tasks, models trained only on \proteus perform on par with, and in several settings outperform, existing task-specific methods.
}

\hypersetup{pdftitle={Proteus: A Scalable Data Engine for Template-Aligned Clothed Humans}}

\begin{document}
\maketitle

\addtocontents{toc}{\protect\setcounter{tocdepth}{-1}}

\begin{figure}[h!]
  \centering
  \includegraphics[width=\textwidth]{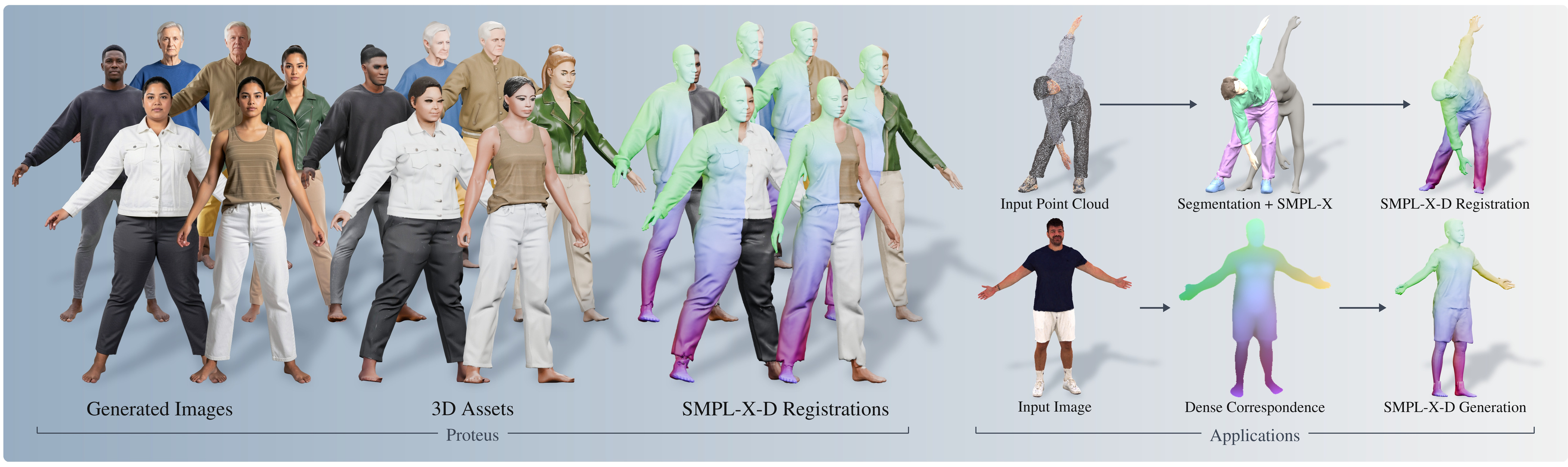}
  \caption{\textbf{Overview of \proteus and its applications.} \proteus pairs generated images of clothed people with lifted 3D assets and dense SMPL-X-D registrations in shared SMPL-X topology. On each registration, the left half is colored by canonical template position to show cross-subject correspondence, and the right half shows the transferred RGB texture. \proteus data supervises a variety of downstream tasks, which work with image and 3D inputs.}
  \label{fig:teaser}
\end{figure}

\section{Introduction}
\label{sec:intro}
Parametric human body models~\cite{loper2015smpl,pavlakos2019smplx,bregier2025humanmeshmodelinganny,MHR:2025,soma2026}
provide a shared topology and skeletal rig across identities,
making them standard foundations for human reconstruction, animation, and generative modeling.
However, these models only represent the minimally clothed body,
and extending them to clothing typically requires giving up template alignment.

Implicit template-free representations~\cite{saito2020pifuhd,xiu2022icon,xiu2023econ,ho2024sith}
capture detailed clothed geometry, while layered representations~\cite{ponsmoll2017clothcap,jiang2020bcnet,corona2021smplicit,zhang2025disclothavatar,nam2025decloth}
model individual garments separately for physics simulation.
For large-scale learning and downstream applications, a compelling alternative is a unified representation:
a single clothed surface aligned to the body template, modeled as per-vertex offsets (SMPL-D)~\cite{alldieck2018videobasedreconstruction,bhatnagar2019mgn,alldieck2019tex2shape,bhatnagar2020ipnet,ma2020cape}.
As every sample shares the same mesh connectivity and UV layout, SMPL-D provides strict vertex-level correspondence across subjects.
Yet this data is largely absent at scale: building a diverse corpus requires extensive multi-view scanning~\cite{tao2021function4d,han2023high,ho2023custom,shen2023xavatar,isik2023humanrf,cheng2023dna,xiong2024mvhumannet}
and reliable non-rigid registration to the template~\cite{amberg2007optimal,bogo2014faust,ponsmoll2017clothcap,groueix2018coded,bhatnagar2020loopreg,marin2024nicp},
which makes data acquisition tedious and expensive.

This scarcity motivates a synthetic approach.
Recent synthetic human datasets~\cite{yang2025sigman,zhuang2025idol,xue2025infinihuman,prospero2026posedreamerscalablephotorealistichuman,hu2026humannova}
demonstrate that modern image~\cite{labs2025flux1kontextflowmatching,wu2025qwenimagetechnicalreport,flux-2-2025} and 3D generative models~\cite{xiang2025trellis,hunyuan3d2025hunyuan3d,xiang2025trellis2}
already possess strong appearance and geometry priors.
However, their outputs cannot directly supervise template-aligned models: generated 3D assets have arbitrary surface topology,
often contain internal geometry layers, and lack any correspondence to a parametric body.
Bringing these unconstrained generative priors into a shared template topology requires careful geometric processing, semantic annotation, and registration.
To address this, we introduce \textit{\proteus}, a scalable synthetic data engine that generates data in strict correspondence across samples.
Starting from structured descriptions of people and their outfits, \proteus generates full-body images, lifts them to textured 3D assets,
and annotates them with semantic labels and skeletal landmarks through multi-view analysis~\cite{khirodkar2026sapiens2,jiang2023rtmlib}.
Then, it fits an SMPL-X body and registers a high-resolution body template to the clothed surface.
Geometric and rendering losses drive the registration, and deformation regularizations~\cite{sorkine2007arap} control how the template adapts to clothing and exposed body regions.

\proteus targets garments whose outer surfaces can be represented as deformations of the body template, excluding skirts, dresses, long open coats, accessories, and shoes.
The output is a single clothed body surface with shared SMPL-X topology, which we denote SMPL-X-D.
To the best of our knowledge, \proteus is the first dataset that pairs textured 3D human assets with dense SMPL-X-D registrations at scale.
We demonstrate the utility of this data by training models exclusively on \proteus for a diverse set of 2D and 3D tasks and evaluating them directly on real-world scans, where our models outperform existing task-specific methods on multiple tasks.

\noindent Our contributions are:
\begin{itemize}
    \item A scalable synthetic data engine, \proteus, that converts generated human assets into clothed surfaces with shared SMPL-X-D topology through multi-view annotation and our novel coarse-to-fine non-rigid registration method.
    \item A large-scale multimodal dataset pairing text descriptions, generated images, textured 3D meshes, semantic annotations, SMPL-X fits, and dense SMPL-X-D registrations.
    \item Downstream models trained only on \proteus that generalize to real scans across four tasks: SMPL-X-D generation from images, feed-forward registration of partial point clouds, semantic segmentation and SMPL-X estimation on raw point clouds, and dense image correspondence prediction.
\end{itemize}

\section{Related Work}
\label{sec:related-work}
\subsection{Clothed Human Geometry}
Template-aligned methods represent clothing as offsets or attached points on a parametric body~\cite{alldieck2019tex2shape,ma2020cape,patel20tailornet,ma2021scale,ma2021pop},
preserving cross-subject correspondence and the skeletal rig of the underlying template.
However, these methods rely on registered 3D scans for supervision, which limits their scalability as such data is scarce and costly to acquire.
A separate line of work~\cite{ponsmoll2017clothcap,bhatnagar2019mgn,jiang2020bcnet,zhu2020deep,corona2021smplicit,zhu2022registering,moon2022clothwild,zhang2025disclothavatar,nam2025decloth,vuran2025BMVC,li2025patterncoordinates}
models garments as separate layers over the body to support garment editing and physics simulation.
Meanwhile, monocular clothed human reconstruction has been largely driven by template-free implicit methods~\cite{saito2020pifuhd,pamir2020,xiu2022icon,xiu2023econ},
often paired with generative priors to complete unobserved regions~\cite{ho2024sith,sanyal2024sculpt,kolotouros2024avatarpopup,tang2025generativehumangeometrydistribution}.
While geometrically detailed, their outputs are unstructured and require post-hoc registration to establish correspondence with a parametric body.
\proteus combines the geometric fidelity of state-of-the-art 3D generative models~\cite{hunyuan3d2025hunyuan3d,xiang2025trellis2,li2026pixal3d}
with a shared SMPL-X-D topology at virtually unlimited scale, enabling a wide range of downstream applications (\cref{sec:application}).
\subsection{Large-scale Human Captures}
Early multi-view datasets such as PeopleSnapshot~\cite{alldieck2019tex2shape}, ZJU-MoCap~\cite{peng2021neural}, and ActorsHQ~\cite{isik2023humanrf} remain standard benchmarks for per-subject avatar reconstruction, but contain few subjects and outfits.
Static~\cite{tao2021function4d,han2023high} and dynamic~\cite{ma2020cape,shen2023xavatar,wang20244ddress} 3D scan datasets, along with large studio captures such as DNA-Rendering~\cite{cheng2023dna} and MVHumanNet~\cite{xiong2024mvhumannet}, offer greater diversity for training generalizable models.
Scans typically supervise single-image reconstruction~\cite{xiu2022icon,xiu2023econ,zhang2024sifu,ho2024sith,xue2024human3diffusion} and sparse-view novel view synthesis~\cite{kwon2024generalizable},
whereas multi-view captures support feed-forward reconstruction~\cite{zubekhin2025giga,liu2025hgm,hu2026forge4d}, multi-view and 4D diffusion~\cite{wang2025meat,jin2025diffuman4d}, and single-image avatar generation~\cite{huang2025adahuman,wang2025geneman,yang2025sigman}.
Despite their scale, these datasets typically only provide minimally clothed SMPL-X fits, leaving the clothed surface unregistered.
Adding non-rigid clothing registration to real captures requires per-scan optimization and remains fundamentally bounded by studio acquisition complexity.
\proteus instead synthesizes and registers clothed human geometry in SMPL-X-D format on demand; we reserve real scans purely for evaluation.
\subsection{Synthetic Human Data Generation}
Rendered synthetic humans with ground-truth body annotations are widely used to supervise pose, shape, and motion estimation~\cite{varol2017surreal,patel2021agora,yang2023synbody,black2023bedlam,tesch2025bedlam2,prospero2026posedreamerscalablephotorealistichuman}.
More recently, 3D avatar reconstruction and generation methods have been supervised using synthetic data from generative models~\cite{yang2025sigman,zhuang2025idol,xue2025infinihuman} or re-rendered 3D assets and 4DGS~\cite{kerbl2023gaussian,wu2024_4dgs} captures~\cite{hu2026humannova}.
However, none of the existing synthetic datasets register clothed geometry to a common body template as we do in \proteus.

\section{\proteus{}: A Scalable Data Engine for Parametric Clothed Humans}
\label{sec:data}
\begin{figure}[t]
  \centering
  \includegraphics[width=\textwidth]{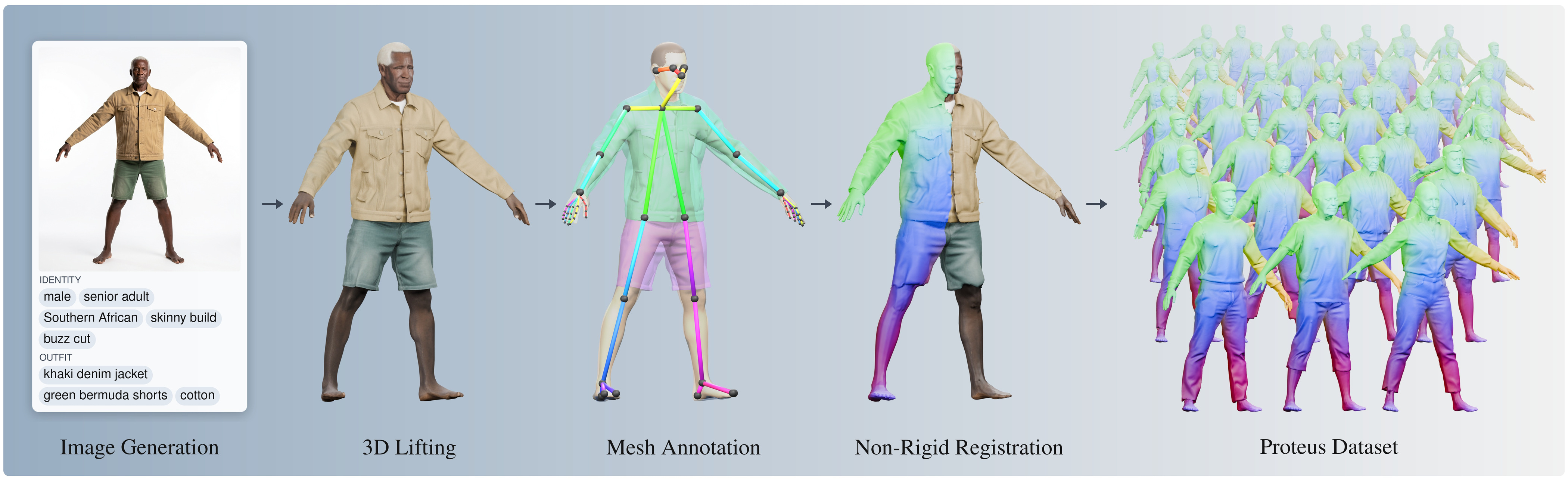}
  \caption{\textbf{The \proteus data engine.} Structured identity and outfit attributes prompt a frontal full-body image, which is lifted to a textured 3D asset. We extract the visible outer surface, annotate it with multi-view semantic labels and triangulated 3D keypoints, fit an SMPL-X body, and deform a high-resolution SMPL-X template onto the surface via coarse-to-fine non-rigid registration. As in Fig.~\ref{fig:teaser}, registered SMPL-X-D meshes show canonical template coordinates on the left and transferred RGB textures on the right.}
  \label{fig:overview}
\end{figure}

Our goal is to construct a large-scale dataset of textured 3D clothed humans registered to a shared, high-resolution SMPL-X topology (SMPL-X-D) without relying on 3D human scans.
\proteus converts text descriptions (\cref{sec:prompts}) into textured clothed humans with a shared SMPL-X topology (Fig.~\ref{fig:overview}).
Raw outputs from image and 3D generative models lack consistent structure and body alignment; \proteus guides generation (\cref{sec:t2i}, \cref{sec:lifting}) and processes each asset through a staged annotation (\cref{sec:annotation}), registration (\cref{sec:registration}), and filtering pipeline (\cref{sec:filtering}).
Each resulting sample contains a prompt $p$, a full-body image $I$, a generated mesh $\mathcal{M}^{\mathrm{gen}}$, its cleaned outer surface $\mathcal{M}^{\mathrm{outer}}$, per-vertex semantic labels $\boldsymbol\ell$, triangulated 3D joints $\mathbf{J}^{\mathrm{op}}$, fitted SMPL-X parameters, and a registered clothed mesh $\mathcal{M}^{\mathrm{reg}}$.
With \proteus, we generate a total of $150$K synthetic clothed humans to supervise our downstream models; \cref{supp:dataset,supp:impl,supp:data-bench} list prompt attributes, benchmark the generation stages, and detail SMPL-X fitting, registration, and statistical filtering.
\subsection{Textual Description Mining}
\label{sec:prompts}
To ensure broad coverage of subjects and wardrobe while keeping poses amenable to registration, we construct each prompt $p$ by sampling attributes from a structured bank: identity (gender, age, ethnicity, body shape), outfit (upper- and lower-body garment types, material properties, and fit), a relaxed A-pose description, and environment (indoor/outdoor, background, lighting).
\subsection{Image Generation}
\label{sec:t2i}
A text-to-image model synthesizes a frontal full-body image $I$ from each prompt $p$.
We use FLUX2-klein-9b~\cite{flux-2-2025}, selected via a benchmark of fast open-source generators against Boogu-Image-Turbo~\cite{chen2026booguimage} and Krea-2-Turbo~\cite{krea2026krea2} (see \cref{supp:data-bench}).
FLUX2-klein reliably follows complex garment descriptions and introduces natural pose variation, whereas the alternatives either exaggerate the A-pose or produce torso close-ups despite explicit full-body prompts.
To guard against generation failures before 3D lifting, a Grounded-SAM2 detector~\cite{ravi2024sam2segmentimages, ren2024grounded} filters out roughly $0.5\%$ of images exhibiting anatomical artifacts such as extra limbs or cropped legs.
\subsection{3D Lifting}
\label{sec:lifting}
We lift each filtered image to a 3D asset $\mathcal{M}^{\mathrm{gen}}$ using one of three state-of-the-art image-to-3D models, each contributing one third of the dataset: TRELLIS.2~\cite{xiang2025trellis2}, Pixal3D~\cite{li2026pixal3d}, and Hunyuan3D-2.1~\cite{hunyuan3d2025hunyuan3d}.
TRELLIS.2 and Pixal3D use an o-voxel representation, producing non-watertight meshes with arbitrary topology and PBR materials that do not strictly replicate the input image appearance.
Hunyuan3D-2.1 extracts watertight meshes from a neural SDF via marching cubes~\cite{lorensen1987marchingcubes}, bakes in the input appearance, and synthesizes occluded regions with multi-view diffusion.
In our benchmark on frontal renders of real scans (see \cref{supp:data-bench}), all three models recover plausible 3D humans that differ primarily in surface triangulation and texture style; combining them mitigates single-generator bias.
\paragraph{Outer surface extraction.}
O-voxel generators frequently synthesize internal geometry layers that are occluded from the outside but would corrupt surface registration.
We render $\mathcal{M}^{\mathrm{gen}}$ from multiple views with a depth-based visibility buffer and discard faces that are never visible or face inward, yielding a clean outer surface $\mathcal{M}^{\mathrm{outer}}$ with vertices $\mathbf{V}^{\mathrm{outer}}$ and faces $\mathbf{F}^{\mathrm{outer}}$.
This step removes $40\%$ of mesh faces on average while preserving the visible clothed geometry.
\subsection{Mesh Annotation}
\label{sec:annotation}
Aligning a parametric body template to an unstructured 3D asset requires both skeletal landmarks for pose initialization and semantic boundaries to distinguish clothing from exposed skin.
We render $\mathcal{M}^{\mathrm{outer}}$ from multiple views, estimate 2D OpenPose-134 keypoints~\cite{cao2018openpose} with RTMlib~\cite{jiang2023rtmlib}, and predict 2D semantic masks with Sapiens2~\cite{khirodkar2026sapiens2}.
We then triangulate the multi-view keypoints into 3D joints $\mathbf{J}^{\mathrm{op}}\in\mathbb{R}^{134\times3}$ and back-project the segmentation logits onto the mesh vertices using depth and visibility weights, assigning a semantic label $\boldsymbol\ell=(\ell_k)_{k=1}^{N_T}\in\{0,\ldots,28\}^{N_T}$ to each of the $N_T$ vertices of $\mathcal{M}^{\mathrm{outer}}$.
\subsection{Non-Rigid Clothed Body Registration}
\label{sec:registration}
Our registration deforms a high-resolution SMPL-X template onto the annotated surface $\mathcal{M}^{\mathrm{outer}}$ in two steps.
First, we fit the SMPL-X body shape $\hat{\boldsymbol\beta}\in\mathbb{R}^{10}$ and axis-angle pose $\hat{\boldsymbol\theta}\in\mathbb{R}^{55\times3}$ to the triangulated skeleton $\mathbf{J}^{\mathrm{op}}$ and target surface $\mathcal{M}^{\mathrm{outer}}$.
Second, keeping the body fit fixed, we optimize dense canonical vertex offsets $\Delta^s \in \mathbb{R}^{N_s \times 3}$ over the clothed body and head (excluding hands and feet) across three progressively subdivided SMPL-X topologies: base (coarse, $N_{s=0}=10$K), once-subdivided (mid, $N_{s=1}=41$K), and twice-subdivided (fine, $N_{s=2}=167$K).
Each stage balances data terms that pull the template toward the target surface, deformation regularizers that maintain local rigidity and smoothness, and a semantic regularizer that restricts offsets in exposed skin and hair regions.
Finally, we transfer the RGB texture from $\mathcal{M}^{\mathrm{outer}}$ to the registered SMPL-X-D mesh via nearest-neighbor matching.
\finding[find:1]{A twice-subdivided SMPL-X surface can represent a wide range of clothing types as a single layer.}

\paragraph{Forward deformation.}
Let $\mathbf{v}_i$ be a canonical template vertex and $\Delta_i$ its learnable offset, with $\Delta^s = \left(\Delta_i\right)_{i=1}^{N_s}$; we omit the stage superscript $s$ below for readability.
Given the fitted SMPL-X parameters $\hat\phi$ (shape $\hat{\boldsymbol\beta}$, pose $\hat{\boldsymbol\theta}$, and global transform), we pose the displaced template via linear blend skinning $\mathcal{W}$:
\begin{equation}
\mathbf{x}_i = \mathcal{W}(\mathbf{v}_i+\Delta_i;\hat\phi).
\label{eq:canonical-offset-template}
\end{equation}
\paragraph{Stage objective.}
Each registration stage minimizes a weighted objective:
\begin{align}
\mathcal{L}_{\mathrm{stage}}
= {}&
\lambda_{\mathrm{p2p}} \mathcal{L}_{\mathrm{p2p}}
 + \lambda_{\mathrm{render}} \mathcal{L}_{\mathrm{render}}
 + \lambda_{\mathrm{offlap}} \mathcal{L}_{\mathrm{offlap}} \notag \\
&+ \lambda_{\mathrm{arap}} \mathcal{L}_{\mathrm{arap}}
+ \lambda_{\mathrm{edge}} \mathcal{L}_{\mathrm{edge}}
 + \lambda_{\mathrm{sel}} \mathcal{L}_{\mathrm{sel}}.
\label{eq:stage-objective}
\end{align}
\paragraph{Geometric data fitting.}
We match posed source vertices $\mathbf{x}_i$ to target points $\mathbf{y}_i$ along their surface normals, rejecting distant matches and pairs with incompatible normals.
Over the set of valid correspondences $\mathcal{C}$, with displacement $\mathbf{d}_i=\mathbf{y}_i-\mathbf{x}_i$ and source and target normals $\mathbf{n}_i,\mathbf{m}_i$, the symmetric point-to-plane loss is
\begin{equation}
\mathcal{L}_{\mathrm{p2p}}
=\frac{1}{2|\mathcal{C}|}\sum_{i\in\mathcal{C}}
\left(|\mathbf{d}_i^\top\mathbf{n}_i|+|\mathbf{d}_i^\top\mathbf{m}_i|\right).
\label{eq:symmetric-ptp}
\end{equation}
To capture silhouette boundaries and fine surface orientation, we also penalize discrepancies between rendered normal maps $\mathbf{N}$ and alpha masks $\mathbf{A}$ of the source ($S$) and target ($T$):
\begin{equation}
\mathcal{L}_{\mathrm{render}}
=\lambda_{\mathbf{N}}\|\mathbf{N}_S-\mathbf{N}_T\|^2
+\lambda_{\mathbf{A}}\|\mathbf{A}_S-\mathbf{A}_T\|^2,
\label{eq:render-loss}
\end{equation}
where $\|\cdot\|^2$ denotes the mean squared error over pixels, channels, and four randomly sampled views (see \cref{supp:impl} for $\lambda_{\mathbf{N}}$ and $\lambda_{\mathbf{A}}$).
\paragraph{Deformation regularization.}
We regularize the canonical offset field with the uniform graph Laplacian $L$, which measures the deviation of each vertex's offset from the mean of its 1-ring neighbors:
\begin{align}
\mathcal{L}_{\mathrm{offlap}}
=
\frac{1}{N_s}\|L\Delta\|_F^2 .
\label{eq:offset-laplacian-loss}
\end{align}
Let $\mathbf{e}_{ij}$ and $\tilde{\mathbf{e}}_{ij}=\mathbf{x}_j-\mathbf{x}_i$ denote reference and deformed edges on the active edge set $\mathcal{E}$, where reference edges are measured on the posed template prior to the current stage's offsets.
An as-rigid-as-possible (ARAP) term~\cite{sorkine2007arap} penalizes non-rigid local distortions, and an edge-length term prevents excessive stretching or compression, akin to the isometry loss in DDC~\cite{habermann2021real}:
\begin{align}
\mathcal{L}_{\mathrm{arap}}
&=\frac{1}{|\mathcal{E}|}\sum_{(i,j)\in\mathcal{E}}
 w_{ij}\|\tilde{\mathbf{e}}_{ij}-R_i\mathbf{e}_{ij}\|_2^2,
\label{eq:arap-loss}\\
\mathcal{L}_{\mathrm{edge}}
&=\frac{1}{|\mathcal{E}|}\sum_{(i,j)\in\mathcal{E}}
(\|\tilde{\mathbf{e}}_{ij}\|_2-\|\mathbf{e}_{ij}\|_2)^2,
\label{eq:edge-loss}
\end{align}
where $w_{ij}$ are non-negative cotangent weights and the local rotations $R_i$ are obtained via polar decomposition of edge covariances.
\paragraph{Skin and hair regularization.}
To prevent garment deformations from pulling adjacent body regions out of place, $\mathcal{L}_{\mathrm{sel}}$ penalizes canonical offsets on template vertices matched to target skin or hair labels (see \cref{supp:impl} for details).
\subsection{Registration Filtering}
\label{sec:filtering}
Manual quality control is infeasible at the scale of $100$K+ assets, so we apply automated statistical filtering to discard suboptimal registrations.
We evaluate each registered sample across multiple criteria: surface distance to the target, offset field distortion, correspondence coverage, and multi-view normal and silhouette agreement.
Data is generated in chunks of $8\text{--}10$K samples; within each chunk, samples are ranked by their worst-performing metric and the bottom $5\%$ are rejected (exact metrics, rejection thresholds, and filtered artifact examples are in \cref{supp:reg-filter}).
\section{Results}
\label{sec:results}
We evaluate the generated 3D assets and our registration pipeline against real 3D scans, testing robustness to both target geometry and SMPL-X initialization as well as comparing against existing non-rigid registration baselines (per-vertex error maps, qualitative comparisons, and extended ablations are in \cref{supp:data-bench}).
\paragraph{Evaluation Protocol.} We curate 290 diverse clothed human scans from four public datasets, CloSe-Di~\cite{antic2024close}, 2K2K~\cite{han2023high}, 4D-DRESS~\cite{wang20244ddress}, and THuman2.1~\cite{tao2021function4d}, along with a fifth set captured in-house and reconstructed with ProbeSDF~\cite{toussaint2025probesdf}.
We report Chamfer distance (CD), F1-score, and normal consistency (NC) as geometric quality measures; further information is provided in \cref{supp:data-bench}.
\subsection{End-to-End Evaluation}
\begin{figure}[t]
  \centering
  \includegraphics[width=0.62\linewidth]{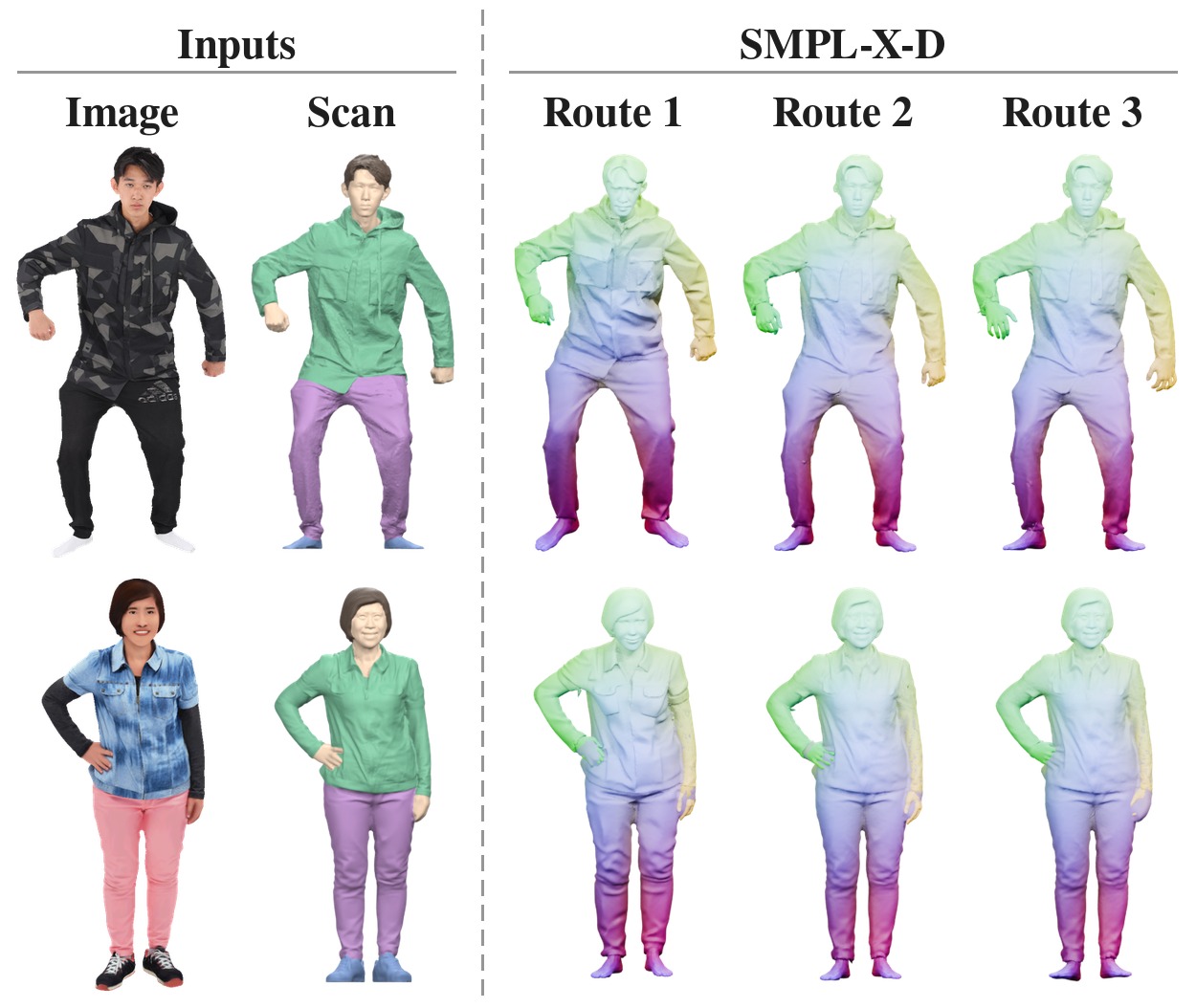}
  \caption{\textbf{Qualitative comparison of the three evaluation routes on real scans.} Route~1 lifts a frontal render of the scan to 3D and registers SMPL-X-D to the lifted mesh. Routes~2 and~3 register SMPL-X-D directly to the ground-truth scan, initialized from our predicted SMPL-X fit (Route~2) or the dataset-provided SMPL-X fit (Route~3). Colors encode template correspondence.}
  \label{fig:end-to-end-evaluation}
\end{figure}

\begin{table}[t]
  \centering
  \caption{\textbf{End-to-end error decomposition on 290 real scans.}
  Routes R1--R3 evaluate each registration against its respective target surface (lifted mesh for R1, real scan for R2 and R3, where R3 uses native SMPL-X fits available for 116 scans).
  \emph{Lifting} compares the raw 3D-lifted assets directly against the ground-truth scans.
  }
  \label{tab:real-scan-routes-main}
  \setlength{\tabcolsep}{4pt}
  \small
  \begin{tabular}{lllrrr}
    \toprule
    \thead{Route} & \thead{Target} & \thead{SMPL-X} & \thead{CD ($\times 10^4$) $\downarrow$} & \thead{F1@0.05$\uparrow$} & \thead{NC$\uparrow$} \\
    \midrule
    R1 & Lifted & Ours   & 18.25 & 0.956 & 0.922 \\
    R2 & Scan   & Ours   & 14.67 & 0.958 & 0.915 \\
    R3 & Scan   & Native & 13.47 & 0.963 & 0.924 \\
    \midrule
    Lifting & Scan & --- & 81.14 & 0.764 & 0.812 \\
    \bottomrule
  \end{tabular}
\end{table}

Evaluating \proteus requires disentangling the error introduced by 3D generative lifting from the error of our SMPL-X fitting and non-rigid registration.
We therefore evaluate the pipeline along three routes on the real scans:
\begin{itemize}
    \item \textbf{Route 1 (Full pipeline on lifted assets):} lifts a frontal render of a real scan to 3D using an image-to-3D generator (\cref{sec:lifting}) and registers SMPL-X-D to the lifted asset, mirroring \proteus generation (\cref{sec:data}).
    \item \textbf{Route 2 (Registration on real scans):} bypasses image generation and 3D lifting, running our multi-view annotation (\cref{sec:annotation}), SMPL-X fitting, and SMPL-X-D registration (\cref{sec:registration}) directly on the ground-truth scan. Comparing Routes~1 and~2 isolates how target origin (generated asset vs.\ real scan) affects registration.
    \item \textbf{Route 3 (Oracle SMPL-X initialization):} follows Route~2 on the real scan, but initializes registration from the dataset-provided ground-truth SMPL-X fit where available, testing sensitivity to SMPL-X initialization.
\end{itemize}
We additionally evaluate the raw lifted 3D assets directly against the ground-truth scans (``Lifting'').

Fig.~\ref{fig:end-to-end-evaluation} illustrates qualitative results across all three routes, and Tab.~\ref{tab:real-scan-routes-main} breaks down the geometric error by stage.
Our registration achieves comparable surface fidelity to its target across all three routes, demonstrating robustness to both generated geometries and predicted SMPL-X initializations.
In contrast, the lifted 3D assets deviate further from the ground-truth scans: as single-image 3D generators synthesize plausible but stochastic geometry in unobserved regions, the 3D lifting stage accounts for the majority of the end-to-end discrepancy against the reference scan.
\finding[find:2]{Our registration retains most of the target surface details, whether the target is a real scan or a generated asset.}

\subsection{Registration Evaluation}
\begin{figure}[t]
  \centering
  \includegraphics[width=0.62\linewidth]{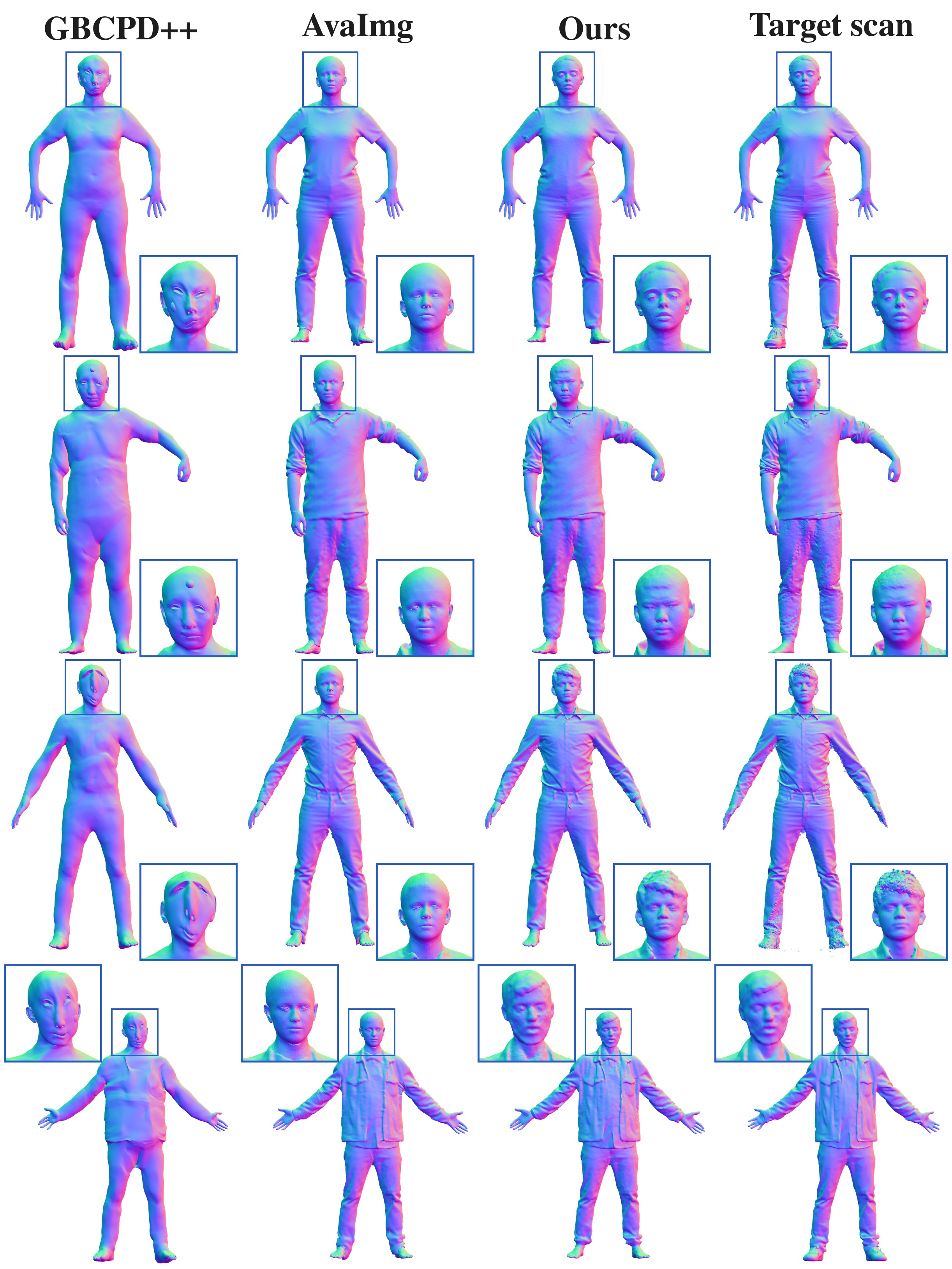}
  \caption{\textbf{Qualitative comparison of non-rigid registration methods on real scans.} We show surface normal renders with zoomed-in insets of the head and collar regions. Our method captures sharp facial features and garment folds within the SMPL-X topology (see \cref{fig:supp-registration-comparison} for all baselines).}
  \label{fig:registration-evaluation}
\end{figure}

\begin{table}[t]
  \centering
  \caption{\textbf{Comparison of non-rigid registration methods on real scans.}
  We report 3D geometric accuracy and multi-view surface normal agreement. Across all tables, we highlight the \ffst{best}, \fsnd{second-best}, and \ftrd{third-best} results.}
  \label{tab:registration-comparison}
  \footnotesize
  \setlength{\tabcolsep}{3pt}
  \fitwidth{%
    \begin{tabular}{lrrrrr}
      \toprule
      & \multicolumn{3}{c}{\thead{3D Metrics}} & \multicolumn{2}{c}{\thead{Multi-view Normals}} \\
      \cmidrule(lr){2-4} \cmidrule(lr){5-6}
      \thead{Method}
      & \multicolumn{1}{c}{\thead{CD ($\times 10^4$) $\downarrow$}}
      & \multicolumn{1}{c}{\thead{F1@0.005 $\uparrow$}}
      & \multicolumn{1}{c}{\thead{NC $\uparrow$}}
      & \thead{SSIM $\uparrow$}
      & \thead{LPIPS $\downarrow$} \\
      \midrule
      Amberg et al.~\cite{amberg2007optimal} & 0.2548 & 0.9032 & \trd{0.9050} & \trd{0.9256} & \trd{0.0761} \\
      GBCPD++~\cite{hirose2023geodesicbcpd} & \fst{0.0876} & \snd{0.9364} & 0.8899 & 0.9190 & 0.0770 \\
      NICP+D~\cite{marin2024nicp} & 0.7511 & 0.7266 & 0.8962 & 0.9203 & 0.0804 \\
      OAReg~\cite{zhao2025oareg} & 0.6330 & 0.6053 & 0.8700 & 0.9078 & 0.1067 \\
      AvaImg~\cite{kostyrko2026avaimg} & \snd{0.1098} & \trd{0.9346} & \snd{0.9414} & \snd{0.9545} & \snd{0.0428} \\
      \textbf{Ours} & \trd{0.1099} & \fst{0.9493} & \fst{0.9469} & \fst{0.9670} & \fst{0.0303} \\
      \bottomrule
    \end{tabular}%
  }
\end{table}

Thanks to our coarse-to-fine registration, \proteus produces SMPL-X-aligned meshes whose surface fidelity closely matches the underlying unconstrained 3D assets.
Without ground-truth SMPL-X-D registrations, we compare our method with non-rigid ICP~\cite{amberg2007optimal}, GBCPD++~\cite{hirose2023geodesicbcpd}, OAReg~\cite{zhao2025oareg}, the Chamfer-based SMPL-X-D registration of NICP~\cite{marin2024nicp}, and Avatar-as-Image (AvaImg)~\cite{kostyrko2026avaimg}.

Fig.~\ref{fig:registration-evaluation} shows GBCPD++, AvaImg, and our method, and Tab.~\ref{tab:registration-comparison} reports quantitative results.
All baselines except AvaImg produce overly smooth surfaces that miss fine target details, whereas our method captures them within the SMPL-X topology.
AvaImg registers body details similarly to our method but over-regularizes the high-resolution head.
Notably, while GBCPD++ attains the lowest Chamfer distance, which favors smooth surfaces that stay close to the target \textit{on average}, it erases high-frequency garment wrinkles; the remaining metrics better reflect detail registration and favor our method.

\section{Applications}
\label{sec:application}
Each \proteus sample pairs a photorealistic image with a textured 3D asset, per-vertex semantic labels, an SMPL-X fit, and a dense SMPL-X-D registration (\cref{sec:data}).
We use this corpus to supervise models across four downstream applications: image-conditioned SMPL-X-D generation (\cref{sec:app-image,sec:app-architectures}), feed-forward registration of point clouds (\cref{sec:app-pointcloud}), semantic segmentation and SMPL-X estimation on raw point clouds (\cref{sec:app-annotation}), and dense SMPL-X correspondence prediction in images (\cref{sec:app-dense}).
Every model is trained exclusively on synthetic \proteus data and evaluated zero-shot on real 3D scans (\cref{supp:models} details data preparation, architectures, training protocols, and baselines; see also the video on our project page).
\paragraph{Training data.}
We use canonical assets generated with the pipeline of \cref{sec:data}.
For image-conditioned SMPL-X-D generation, we render each asset from four random views; this task assumes a relaxed input pose with separated limbs, matching the A-pose registrations produced by \proteus.
Since point-cloud segmentation, SMPL-X estimation, and dense image correspondence prediction require diverse body poses, we re-pose each SMPL-X-D asset with up to five AMASS~\cite{mahmood2019amass} poses via linear blend skinning.
Each SMPL-X-D mesh carries per-vertex semantic labels transferred from $\mathcal{M}^{\mathrm{outer}}$ alongside the RGB texture, and we rasterize pixel-level canonical SMPL-X coordinates across rendered views.
\subsection{Image-Conditioned SMPL-X-D Generation}
\label{sec:app-image}
\begin{table}[t]
  \centering
  \caption{\textbf{Single-image clothed human reconstruction on real scans.}
  Body-guided methods (\good) receive the Route~2 SMPL-X fit alongside the input image, whereas body-free methods (\bad) use the image alone.
  For our model, we report the best of 16 samples (lowest CD per scan).
  }
  \label{tab:image-to-smplxd}
  \footnotesize
  \setlength{\tabcolsep}{3pt}
  \fitwidth{%
    \begin{tabular}{llrrrrrr}
      \toprule
      \thead{Method}
      & \thead{Body}
      & \multicolumn{1}{c}{\thead{CD ($\times 10^4$) $\downarrow$}}
      & \multicolumn{1}{c}{\thead{F1@0.01 $\uparrow$}}
      & \multicolumn{1}{c}{\thead{PSNR $\uparrow$}}
      & \multicolumn{1}{c}{\thead{SSIM $\uparrow$}}
      & \multicolumn{1}{c}{\thead{LPIPS $\downarrow$}}
      & \multicolumn{1}{c}{\thead{FID-CLIP $\downarrow$}} \\
      \midrule
      Human3Diffusion~\cite{xue2024human3diffusion} & \bad & 103.4 & 0.628 & 14.18 & 0.583 & 0.344 & 59.7 \\
      HumanNOVA~\cite{hu2026humannova} & \bad & 103.8 & 0.660 & 14.28 & 0.590 & 0.329 & 51.6 \\
      SyncHuman~\cite{chen2025synchuman} & \bad & 69.7 & 0.785 & \trd{15.73} & \trd{0.668} & \fst{0.251} & \fst{16.9} \\
      \midrule
      SiTH~\cite{ho2024sith} & \good & 72.2 & 0.762 & 15.28 & 0.646 & \trd{0.280} & \snd{17.5} \\
      SIFU~\cite{zhang2024sifu} & \good & \snd{61.2} & \snd{0.813} & \snd{15.74} & \trd{0.668} & 0.286 & 30.5 \\
      SHERT~\cite{zhan2024shert} & \good & 69.6 & 0.783 & 15.11 & 0.630 & 0.305 & 34.8 \\
      VS~\cite{liu2024vs} & \good & \trd{62.3} & \trd{0.811} & \snd{15.74} & \snd{0.675} & \trd{0.280} & 32.3 \\
      \midrule
      \textbf{Ours} & \good & \fst{56.4} & \fst{0.848} & \fst{16.14} & \fst{0.677} & \snd{0.260} & \trd{21.7} \\
      \bottomrule
    \end{tabular}%
  }
\end{table}

\begin{figure}[t]
  \centering
  \includegraphics[width=\textwidth]{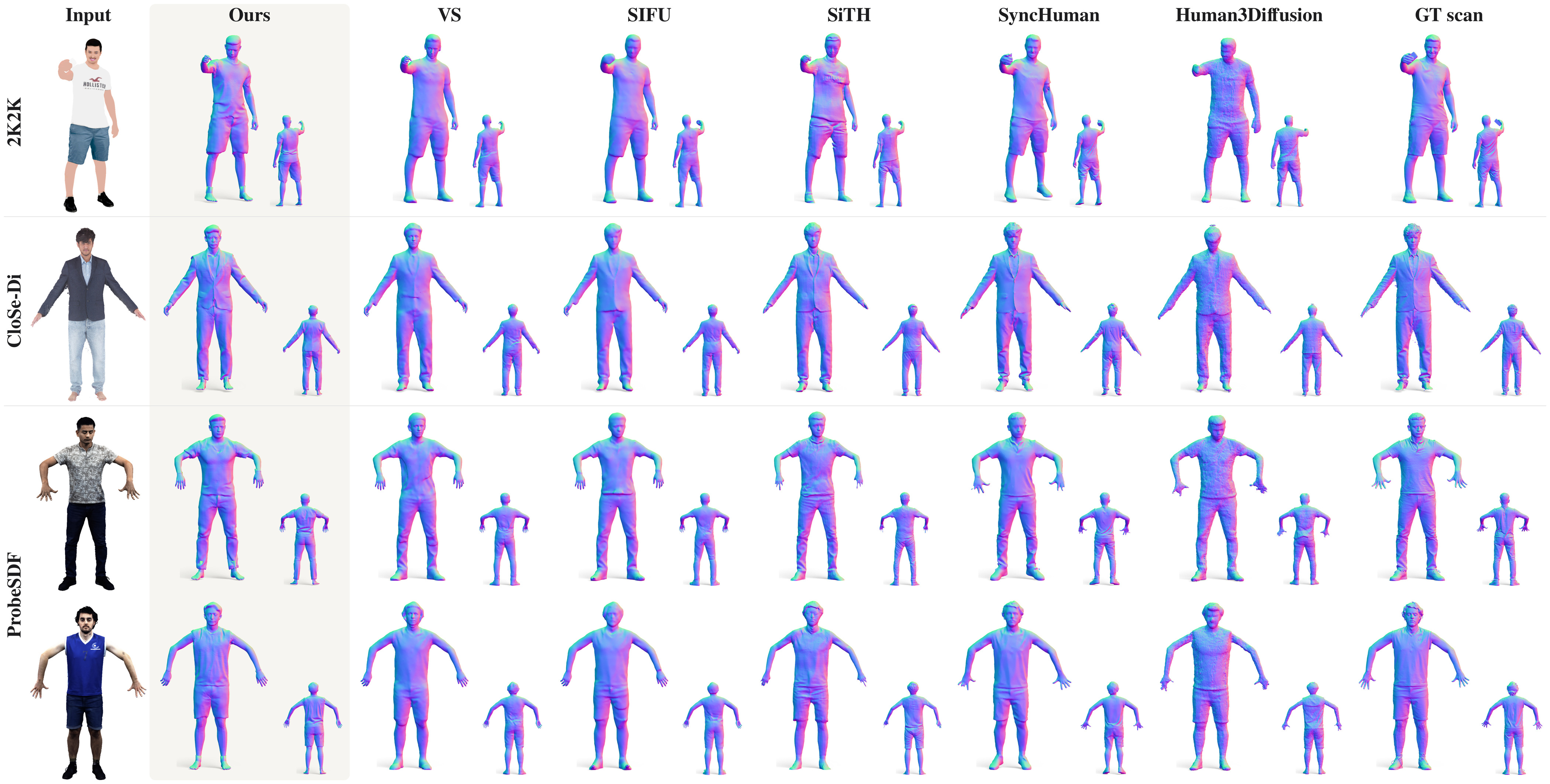}
  \caption{\textbf{Qualitative comparison of single-image clothed human reconstruction on real scans.} Each row shows the frontal input image, an output of our method, the body-guided (VS, SIFU, SiTH) and body-free (SyncHuman, Human3Diffusion) baselines, and the ground-truth scan, rendered as camera-space normals from the front and back. Our samples recover garment details such as lapels, pocket flaps, and trouser creases while remaining registered to SMPL-X.}
  \label{fig:image-comparison}
\end{figure}

Given an image of a person and an SMPL-X body fit, our first model generates SMPL-X-D offsets that dress the parametric body.
Taking advantage of the shared template topology, we bake the fine-stage canonical offsets $\Delta^{s=2}$ into a $256\times256$ UV displacement map and train a conditional rectified-flow model~\cite{lipman2023flowmatching,liu2023rectifiedflow} directly in texel space, following the PixelDiT architecture~\cite{yu2026pixeldit}.
Conditioning features are extracted from the input image with a frozen DINOv3~\cite{simeoni2026dinov3} encoder and injected into the MM-DiT blocks~\cite{esser2024scaling}, while the body shape $\boldsymbol\beta$ is embedded via a lightweight MLP to modulate the blocks via AdaIN~\cite{huang2017adain}.

We evaluate on the 290 real scans introduced in \cref{sec:results}, using a frontal render of each scan and its Route~2 SMPL-X fit as input.
We measure 3D geometric accuracy with Chamfer distance (CD) and F1 score, multi-view normal consistency with PSNR, SSIM, and LPIPS~\cite{zhang2018lpips}, and distributional realism with FID~\cite{heusel2017fid} in CLIP~\cite{radford2021clip} feature space.
Prior to evaluation, we account for each method's coordinate convention using a single fixed rotation and scale across all scans, followed by translation-only ICP alignment.
\paragraph{Baselines.}
We compare against two families of single-image clothed human reconstruction methods: body-guided approaches, SiTH~\cite{ho2024sith}, SIFU~\cite{zhang2024sifu}, SHERT~\cite{zhan2024shert}, and VS~\cite{liu2024vs}, which condition on an SMPL-X estimate, and body-free approaches, Human3Diffusion~\cite{xue2024human3diffusion}, HumanNOVA~\cite{hu2026humannova}, and SyncHuman~\cite{chen2025synchuman}, which reconstruct geometry from the image alone.
General-purpose image-to-3D generators such as TRELLIS.2~\cite{xiang2025trellis2} and Hunyuan3D-2.1~\cite{hunyuan3d2025hunyuan3d} are omitted here as \proteus uses them during data synthesis; their reconstruction performance on the same benchmark is reported in \cref{tab:real-scan-routes-main}.

\finding[find:3]{A model trained only on synthetic \proteus data is competitive with dedicated reconstruction methods when evaluated on real scans.}

Conditioned on the input image and body shape, our model achieves the best 3D geometric accuracy (CD and F1) as well as the highest PSNR and SSIM across all methods (\cref{tab:image-to-smplxd}), while producing outputs that are already registered to SMPL-X.
Body-guided baselines (SiTH, SIFU, SHERT, VS) anchor their predictions to the input SMPL-X fit and refine the surface against predicted front-and-back normal maps via test-time optimization, which aids their normal-map metrics.
Body-free methods (Human3Diffusion, HumanNOVA, SyncHuman) leverage strong multi-view diffusion priors to synthesize detailed geometry, but operate in uncalibrated coordinate frames that require global alignment.
Crucially, every external baseline outputs an unstructured mesh that requires a separate post-hoc registration step to establish correspondence with SMPL-X, whereas our model is feed-forward, template-aligned by construction, and matches or outperforms dedicated reconstruction pipelines.
We provide qualitative comparisons for selected methods in \cref{fig:image-comparison}.

\subsection{Same-Data Comparison of Template-Aligned Models}
\label{sec:app-architectures}
\begin{table}[t]
  \centering
  \caption{\textbf{Comparison of template-aligned architectures trained on \proteus.}
  For stochastic models, paired metrics are reported as the average (\emph{Mean}) and per-metric best (\emph{Best}) over 16 samples per scan; deterministic regressors produce a single prediction.
  LPIPS, FID, and CMMD are evaluated on 8-view surface normal renders.}
  \label{tab:same-data-models}
  \footnotesize
  \setlength{\tabcolsep}{2.5pt}
  \fitwidth{%
    \begin{tabular}{llrrrrrrrr}
      \toprule
      & & \multicolumn{2}{c}{\thead{CD ($\times 10^4$) $\downarrow$}} & \multicolumn{2}{c}{\thead{F1@0.01 $\uparrow$}} & \multicolumn{2}{c}{\thead{LPIPS $\downarrow$}} & & \\
      \cmidrule(lr){3-4} \cmidrule(lr){5-6} \cmidrule(lr){7-8}
      \thead{Method}
      & \thead{Type}
      & \multicolumn{1}{c}{\thead{Mean}}
      & \multicolumn{1}{c}{\thead{Best}}
      & \multicolumn{1}{c}{\thead{Mean}}
      & \multicolumn{1}{c}{\thead{Best}}
      & \multicolumn{1}{c}{\thead{Mean}}
      & \multicolumn{1}{c}{\thead{Best}}
      & \multicolumn{1}{c}{\thead{FID $\downarrow$}}
      & \multicolumn{1}{c}{\thead{CMMD $\downarrow$}} \\
      \midrule
      Tex2Shape~\cite{alldieck2019tex2shape} & Reg. & \trd{93.7} & 93.7 & \trd{0.632} & 0.632 & \snd{0.303} & 0.303 & 40.1 & 0.521 \\
      IDOL~\cite{zhuang2025idol} & Reg. & 96.9 & 96.9 & 0.608 & 0.608 & 0.310 & 0.310 & \trd{38.9} & \trd{0.500} \\
      CAPE~\cite{ma2020cape} & CVAE & 95.2 & \trd{79.5} & 0.626 & \trd{0.705} & \trd{0.304} & \trd{0.286} & 49.4 & 0.666 \\
      SCULPT~\cite{sanyal2024sculpt} & GAN & \snd{89.5} & \snd{72.9} & \snd{0.651} & \snd{0.742} & 0.312 & \snd{0.284} & \fst{21.6} & \fst{0.247} \\
      \midrule
      \textbf{Ours} & Flow & \fst{73.6} & \fst{62.9} & \fst{0.734} & \fst{0.802} & \fst{0.294} & \fst{0.274} & \snd{23.9} & \snd{0.275} \\
      \bottomrule
    \end{tabular}%
  }
\end{table}

The baselines in \cref{sec:app-image} differ in their training corpora, generative priors, test-time refinement, and output representations.
To isolate the impact of the model architecture under identical supervision, we benchmark representative template-aligned methods trained on the same \proteus data and adapted to the SMPL-X-D representation: Tex2Shape~\cite{alldieck2019tex2shape}, IDOL~\cite{zhuang2025idol}, CAPE~\cite{ma2020cape}, and SCULPT~\cite{sanyal2024sculpt} (see \cref{supp:models-samedata} for implementation details).
In our unified setup, Tex2Shape and IDOL act as deterministic regressors of the UV displacement texture, CAPE is a graph-based conditional VAE over vertex offsets, and SCULPT is a conditional StyleGAN3~\cite{karras2021stylegan3} generator in UV space, compared against our rectified-flow model from \cref{sec:app-image}.
\paragraph{Metrics.}
Alongside 3D CD and F1, we evaluate 8-view normal renders using paired LPIPS~\cite{zhang2018lpips} as well as distributional FID~\cite{heusel2017fid} and CMMD~\cite{jayasumana2024cmmd} in CLIP~\cite{radford2021clip} feature space.
Since paired metrics inherently favor smooth predictions near the conditional mean, we report both the per-scan average over 16 drawn samples (\textit{Mean}) and the best sample per metric (\textit{Best}) for all stochastic models.
\finding[find:4]{\proteus effectively supervises regression, variational, generative adversarial, and flow-based architectures alike.}

Quantitative results are summarized in \cref{tab:same-data-models}.
The deterministic regressors (Tex2Shape and IDOL) collapse toward the conditional mean, yielding overly smooth surfaces whose surface normal distributions diverge from real scans.
Despite paired metrics favoring smooth averages, both regressors trail behind even the mean sample of our flow model.
Among the generative baselines, CAPE exhibits limited sample diversity, whereas adversarial training in SCULPT yields sharp surface detail and the lowest FID/CMMD scores.
Our rectified-flow model nearly matches SCULPT in distributional sharpness while outperforming all baselines in 3D geometric accuracy and paired LPIPS, both on average and for the best sample.
Qualitative comparisons across all five architectures are provided in \cref{fig:supp-samedata}.
\subsection{Feed-Forward Registration of Point Clouds}
\label{sec:app-pointcloud}
\begin{figure}[t]
  \centering
  \includegraphics[width=\linewidth]{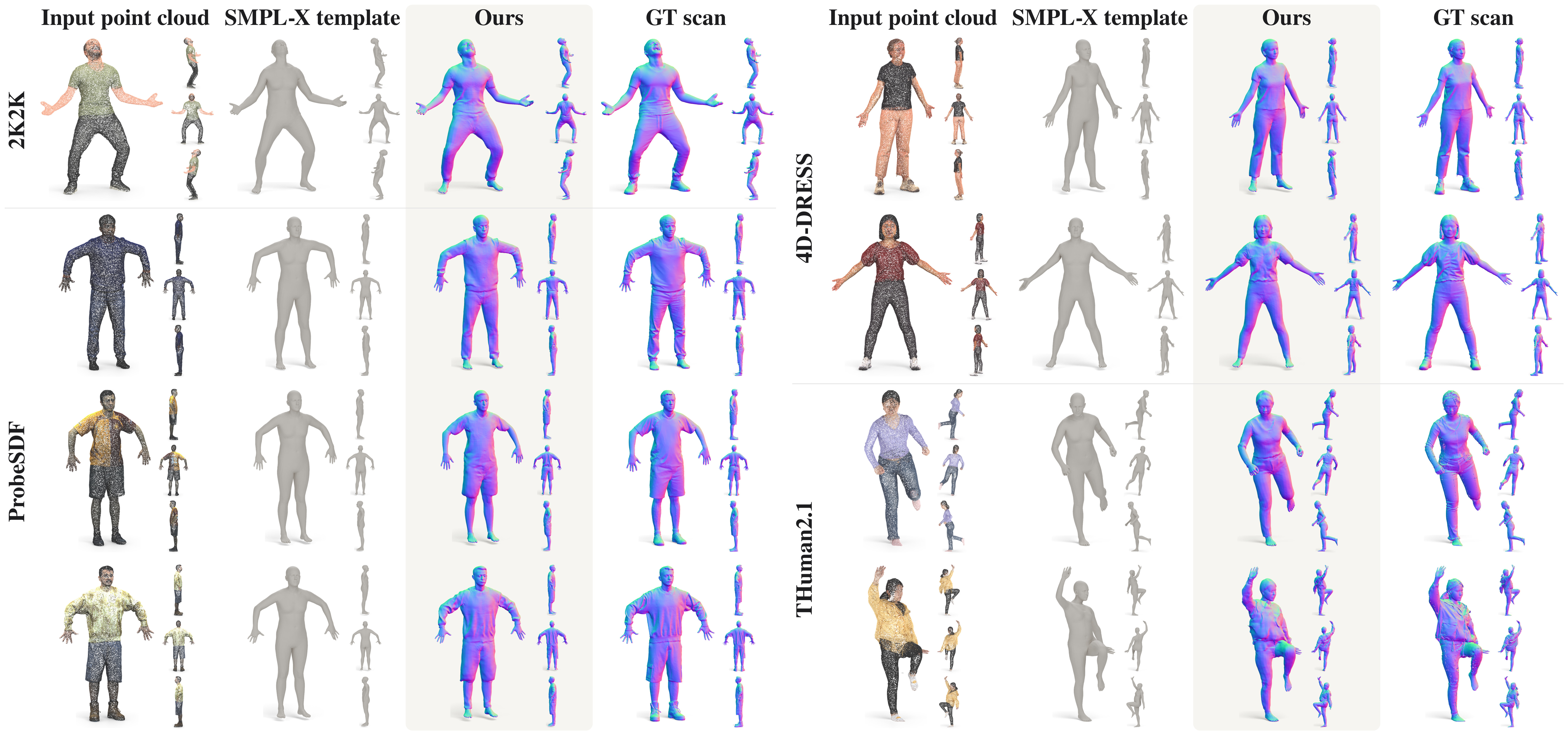}
  \caption{\textbf{Feed-forward registration of partial point clouds.} Each row shows the partial 8-camera input point cloud, the conditioning SMPL-X fit, our feed-forward SMPL-X-D prediction across four views, and the ground-truth scan (rendered as camera-space normals).}
  \label{fig:ff-registration}
\end{figure}

While the optimization-based registration in \proteus (\cref{sec:registration}) accurately fits individual assets, it requires per-sample iterative optimization.
We distill this registration process into a feed-forward generator that aligns SMPL-X-D to a raw, potentially partial point cloud, conditioned on an initial SMPL-X body estimate.
We retain the UV-space PixelDiT denoiser from \cref{sec:app-image} and replace the image conditioning tokens with localized point-cloud features.
Specifically, we place a $64\times64$ grid of UV anchors on the posed SMPL-X surface; each anchor pools features from its nearest input points in local posed coordinates via learned attention, and the aggregated features are added to the corresponding UV texture tokens.
To make the network robust to imperfect off-the-shelf SMPL-X fits and sensor noise, we perturb the conditioning SMPL-X parameters during training, re-expressing target offsets relative to the perturbed body, and augment input point clouds with synthetic noise, outliers, and cutouts.

As no existing method directly predicts template-aligned clothed meshes from partial point clouds, we demonstrate this application qualitatively in \cref{fig:ff-registration} on partial 8-camera point clouds of real scans (conditioned on their Route~2 SMPL-X fits).
The generated SMPL-X-D surfaces conform closely to the observed points while plausibly completing occluded clothing and hair regions absent from the parametric body.
Trading off only the finest high-frequency wrinkles, this feed-forward model registers raw point clouds in 50 denoising steps, orders of magnitude faster than optimization-based non-rigid registration (further analysis in \cref{supp:models-pointcloud}).
\subsection{Annotation of Raw Point Clouds}
\label{sec:app-annotation}
\begin{table}[t]
  \centering
  \caption{\textbf{3D point-cloud segmentation on human-annotated real scans.}
  We report overall accuracy (OA), mean IoU (mIoU), and mean F1 (mF1) across grouped semantic classes (skin, hair, shoes, and upper/lower clothing, with skin, hair, and shoes merged on MGN).
  CloSeNet is evaluated both unconstrained and with an oracle list of garment classes present in each scan.}
  \label{tab:segmentation-baselines}
  \footnotesize
  \setlength{\tabcolsep}{2.5pt}
  \fitwidth{%
    \begin{tabular}{lrrrrrrrrr}
      \toprule
      & \multicolumn{3}{c}{\thead{4D-DRESS}} & \multicolumn{3}{c}{\thead{CloSe-Di}} & \multicolumn{3}{c}{\thead{MGN}} \\
      \cmidrule(lr){2-4} \cmidrule(lr){5-7} \cmidrule(lr){8-10}
      \thead{Method} & \thead{OA} & \thead{mIoU} & \thead{mF1} & \thead{OA} & \thead{mIoU} & \thead{mF1} & \thead{OA} & \thead{mIoU} & \thead{mF1} \\
      \midrule
      CloSeNet~\cite{antic2024close} & .796 & .668 & .798 & .944 & .886 & .939 & .851 & .724 & .831 \\
      CloSeNet (oracle) & \trd{.869} & \trd{.747} & \trd{.855} & \fst{.972} & \fst{.937} & \fst{.967} & \trd{.967} & \trd{.925} & \trd{.960} \\
      Sonata~\cite{wu2025sonata} probe & .865 & .735 & .845 & .859 & .736 & .844 & .912 & .822 & .901 \\
      MV Sapiens2~\cite{khirodkar2026sapiens2} & \fst{.964} & \fst{.910} & \fst{.951} & \trd{.947} & \trd{.891} & \trd{.941} & \fst{.973} & \fst{.937} & \fst{.967} \\
      \midrule
      \textbf{Ours} & \snd{.947} & \snd{.877} & \snd{.934} & \snd{.948} & \snd{.892} & \snd{.942} & \snd{.971} & \snd{.936} & \fst{.967} \\
      \bottomrule
    \end{tabular}%
  }
\end{table}

To annotate its assets, \proteus renders each mesh from 50 views and aggregates predictions from a large 2D Sapiens2 model (\cref{sec:annotation}), which is accurate but computationally heavy.
Using the resulting per-vertex labels in \proteus as supervision, we distill this multi-view 2D teacher into a single-pass 3D point transformer based on Utonia~\cite{zhang2026utonia} that predicts all 29 Sapiens2~\cite{khirodkar2026sapiens2} semantic classes directly from a raw point cloud.
We evaluate this 3D segmentation model zero-shot on human-annotated scans from 4D-DRESS~\cite{wang20244ddress}, CloSe-Di~\cite{antic2024close}, and MGN~\cite{bhatnagar2019mgn}.
We compare against three baselines: CloSeNet~\cite{antic2024close}, which is trained on real annotated scans; a linear probe trained on frozen Sonata~\cite{wu2025sonata} features using \proteus; and the 50-view Sapiens2 aggregation pipeline itself, which serves as a reference.

As shown in \cref{tab:segmentation-baselines}, our model matches the 50-view Sapiens2 teacher on CloSe-Di and MGN and trails only slightly on 4D-DRESS (primarily on footwear).
CloSeNet surpasses our zero-shot model only on its native CloSe-Di benchmark when given an oracle list of present garment classes as input.
Furthermore, the substantial margin over the Sonata linear probe highlights the benefit of fine-tuning the point-cloud backbone end-to-end on \proteus human geometry (qualitative comparisons are in \cref{fig:supp-segmentation-comparison}).
\paragraph{SMPL-X estimation.}
Following a similar setup to ETCH-X~\cite{li2026etchxrobustifyexpressivebody}, we also train a Utonia-based regressor to predict SMPL-X pose and shape directly from a clothed, potentially partial point cloud in a single forward pass.
Evaluated on real scans, our feed-forward regressor is competitive with ETCH-X: it recovers more accurate skeletal poses, fits exposed skin slightly less tightly, and runs orders of magnitude faster (see \cref{supp:models-scan-smplx} for detailed benchmarks).
\finding[find:5]{\proteus effectively supervises a family of 3D annotation models, enabling feed-forward semantic segmentation, parametric body estimation, and non-rigid registration of raw point clouds in place of multi-view and optimization-based pipelines.}

\subsection{Dense Correspondence 2D}
\label{sec:app-dense}
DensePose~\cite{guler2018densepose} introduced dense pixel-to-surface correspondence for human bodies, and recent face models such as Pixel3DMM~\cite{thies2016face2face,giebenhain2026pixel3dmm} have shown that dense template predictions provide a powerful signal for monocular 3D tracking.
As \proteus registers clothing onto the SMPL-X surface, we can extend dense canonical correspondence to full-body clothed humans, mapping every foreground pixel to its corresponding coordinate on the SMPL-X template across both exposed skin and garments.

We fine-tune a DINOv3~\cite{simeoni2026dinov3} backbone paired with a multi-scale convolutional feature pyramid to jointly estimate per-pixel canonical template coordinates, relative depth, and 3D body joints.
Given camera intrinsics, these dense correspondences and joint predictions drive a monocular SMPL-X optimization, extending NLF~\cite{sarandi2024nlf} by allowing clothed pixels to deviate from the minimally clothed body surface.
Evaluated on frontal renders of the real scans, the resulting fits align more closely with the image than NLF-L and recover more accurate 3D poses on 4D-DRESS, although NLF-L remains more accurate in 3D across the full benchmark.
We showcase monocular video tracking results on our project page and provide the full comparison against NLF in \cref{supp:models-dense}.
\finding[find:6]{Registered clothed geometry from \proteus extends dense template correspondence from the body to the clothes: every foreground pixel, on skin or garment, maps onto SMPL-X, turning a single image into a parametric body fit.}

\subsection{Composing the Models}
\label{sec:app-composition}
Taken together, the models trained on \proteus form a self-contained, feed-forward system that converts raw 2D or 3D observations into registered SMPL-X-D avatars.
Given a raw 3D scan, our point-cloud networks predict semantic labels and an SMPL-X body fit, which then condition the feed-forward point-cloud registration model to produce a complete SMPL-X-D mesh, replacing the multi-view rendering, optimization-based body fitting, and non-rigid registration stages of \proteus with learned forward passes.
Likewise, given a single RGB image, our dense correspondence network enables monocular SMPL-X fitting, which in turn conditions our image-to-SMPL-X-D generator.

\section{Limitations and Future Work}
\label{sec:limitations}
\proteus represents each clothed person as a single surface deformed from the SMPL-X template, keeping every sample in dense correspondence and directly animatable.
Garments with topologies that depart from the body, such as skirts, dresses, and long open coats, cannot be modeled by offsets alone and would require an additional garment layer on top of the template.
In addition, the engine currently synthesizes assets in relaxed A-poses and obtains posed training data via linear blend skinning, which does not capture pose-dependent cloth dynamics.
Since our annotation, SMPL-X fitting, and registration stages are pose-agnostic, incorporating pose-conditioned image generation would extend \proteus to diverse poses without modifying the rest of the pipeline.
Finally, our downstream models use compact architectures to demonstrate the breadth of tasks a single \proteus corpus can supervise; each can be further specialized, for example, by improving feed-forward point-cloud registration to recover the finest high-frequency wrinkles captured by our optimization-based registration.
\section{Conclusion}
\label{sec:conclusion}
We presented \proteus, a scalable synthetic data engine that converts text descriptions into textured clothed humans sharing the topology and skeletal rig of SMPL-X.
Through multi-view semantic annotation and coarse-to-fine non-rigid registration, \proteus aligns unconstrained 3D generative outputs to a high-resolution SMPL-X-D template while preserving fine clothing detail as faithfully as on real scans.
Trained exclusively on \proteus and evaluated zero-shot on real-world scans, our downstream models generate template-aligned clothed avatars from images, matching or outperforming dedicated reconstruction methods across regression and generative paradigms, register partial point clouds in a feed-forward pass, segment and fit SMPL-X bodies to raw point clouds, and predict dense pixel-to-template correspondences in images.
Composed together, these models turn raw 2D images and 3D scans directly into animatable SMPL-X-D meshes, establishing template-aligned synthetic humans as an effective, scalable source of supervision for 3D human modeling.

\section*{Acknowledgments}
This work was funded by the Saarbr\"ucken Research Center for Visual Computing and Artificial Intelligence (VIA).

\bibliographystyle{ieeenat_fullname}
\bibliography{main}

\beginappendix
\addtocontents{toc}{\protect\setcounter{tocdepth}{2}}
\tableofcontents
\clearpage
\section{The Name \proteus}
\label{supp:name}
In Greek mythology, Proteus is a sea god capable of changing his shape at will.
In Homer's \emph{Odyssey}, Menelaus holds fast to Proteus while he transforms into a lion, a serpent, a leopard, a boar, running water, and a tree, before returning to his true form.
Our data engine performs a similar transformation: a single SMPL-X body template is deformed to match every clothed human asset it generates.
All samples share the same underlying template topology and remain in strict dense correspondence.

\section{Dataset Details}
\label{supp:dataset}
\subsection{Prompt Attributes and Diversity}
\begin{figure}[!tbp]
  \centering
  \includegraphics[width=\textwidth]{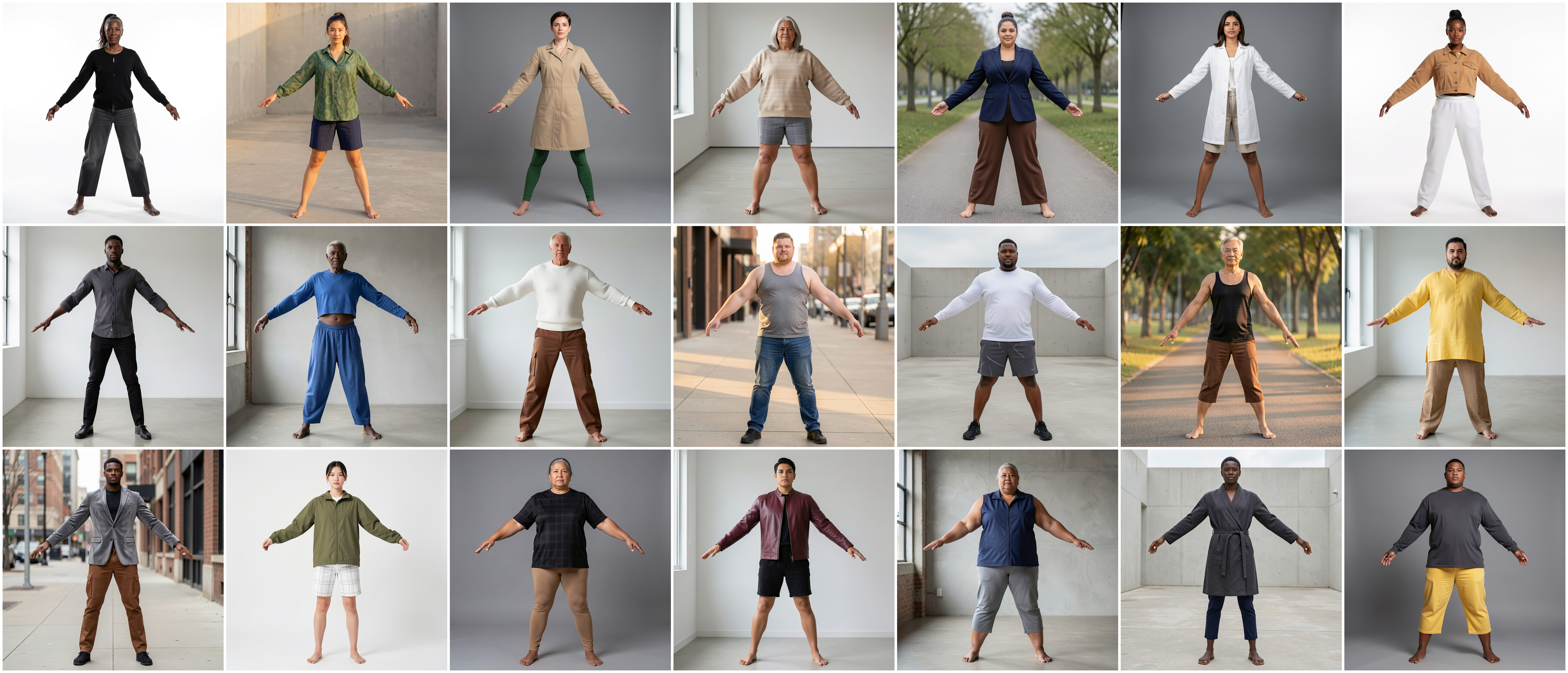}
  \caption{\textbf{Demographic diversity of generated text-to-image samples in \proteus.} Stratified prompt sampling spans diverse ages, genders, ethnicities, body compositions, and garment styles.}
  \label{fig:supp-t2i-demographic-diversity}
\end{figure}

Tables~\ref{tab:prompt-headline-distributions}, \ref{tab:prompt-ethnicity-distributions}, and \ref{tab:prompt-garment-types} summarize the attribute categories and sampling distributions of \proteus over a representative sample of $N=24$K prompts drawn from the same generator used for the full $156$K prompt bank, and Figure~\ref{fig:supp-t2i-demographic-diversity} shows 21 representative text-to-image outputs spanning gender, ethnicity, age, body composition, and outfit styles.
\begin{table}[!tbp]
  \centering
  \caption{\textbf{Subject and scene attribute distributions.} Counts and shares across a representative sample of $N=24$K prompts from the full $156$K prompt bank prior to downstream image and registration filtering.}
  \label{tab:prompt-headline-distributions}
  \scriptsize
  \setlength{\tabcolsep}{3pt}
  \begin{tabular}{@{}llrr@{\hspace{18pt}}llrr@{}}
    \toprule
    \thead{Attribute} & \thead{Value} & \thead{Count} & \thead{Share} & \thead{Attribute} & \thead{Value} & \thead{Count} & \thead{Share} \\
    \midrule
    Age & adult & 8,302 & 34.59\% & Gender & female & 10,774 & 44.89\% \\
     & senior adult & 7,151 & 29.80\% &  & male & 10,751 & 44.80\% \\
     & young adult & 4,868 & 20.28\% &  & neutral & 2,475 & 10.31\% \\
     & elderly & 3,679 & 15.33\% & & & & \\
    \midrule
    Body composition & average & 8,418 & 35.08\% & Scene family & studio & 14,496 & 60.40\% \\
      & athletic & 4,920 & 20.50\% &   & indoor & 5,802 & 24.18\% \\
      & slightly overweight & 3,658 & 15.24\% &   & outdoor & 3,702 & 15.43\% \\
      & skinny & 3,645 & 15.19\% & Background & seamless white studio & 10,191 & 42.46\% \\
      & overweight & 3,359 & 14.00\% &  & gray studio backdrop & 4,305 & 17.94\% \\
    & & & &  & plain room & 3,183 & 13.26\% \\
    & & & &  & concrete courtyard & 1,521 & 6.34\% \\
    & & & &  & gallery space & 1,443 & 6.01\% \\
    & & & &  & city sidewalk & 1,293 & 5.39\% \\
    & & & &  & loft wall & 1,176 & 4.90\% \\
    & & & &  & park path & 888 & 3.70\% \\
    & & & & Lighting & softbox & 12,291 & 51.21\% \\
    & & & &  & side-window light & 5,802 & 24.18\% \\
    & & & &  & overcast daylight & 2,745 & 11.44\% \\
    & & & &  & Rembrandt & 2,205 & 9.19\% \\
    & & & &  & golden-hour sunlight & 957 & 3.99\% \\
    \bottomrule
  \end{tabular}
\end{table}

\begin{table}[!tbp]
  \centering
  \caption{\textbf{Ethnicity distribution in the prompt catalog.} Counts and shares across $N=24$K sampled prompts.}
  \label{tab:prompt-ethnicity-distributions}
  \scriptsize
  \setlength{\tabcolsep}{3pt}
  \begin{tabular}{@{}p{0.25\textwidth}rr@{\hspace{30pt}}p{0.25\textwidth}rr@{}}
    \toprule
    \thead{Ethnicity} & \thead{Count} & \thead{Share} & \thead{Ethnicity} & \thead{Count} & \thead{Share} \\
    \midrule
    Central Asian & 1,098 & 4.58\% & East Asian & 1,149 & 4.79\% \\
    South Asian & 1,134 & 4.73\% & Southeast Asian & 1,205 & 5.02\% \\
    West Asian / Middle Eastern & 1,279 & 5.33\% & North African & 1,249 & 5.20\% \\
    West African & 1,164 & 4.85\% & East African & 1,140 & 4.75\% \\
    Central African & 1,188 & 4.95\% & Southern African & 1,218 & 5.08\% \\
    African American & 981 & 4.09\% & Afro-Caribbean & 936 & 3.90\% \\
    Latin American & 1,785 & 7.44\% & Indigenous North American & 672 & 2.80\% \\
    Indigenous Central / South American & 720 & 3.00\% & Inuit & 714 & 2.98\% \\
    Pacific Islander & 720 & 3.00\% & Indigenous Australian / Torres Strait Islander & 768 & 3.20\% \\
    European & 2,924 & 12.18\% & Mixed heritage & 1,956 & 8.15\% \\
    \bottomrule
  \end{tabular}
\end{table}

\begin{table}[!tbp]
  \centering
  \caption{\textbf{Upper- and lower-body garment distributions.} Counts and shares across $N=24$K sampled prompts.}
  \label{tab:prompt-garment-types}
  \scriptsize
  \setlength{\tabcolsep}{2pt}
  \begin{minipage}[t]{0.32\textwidth}
    \vspace{0pt}
    \begin{tabular*}{\linewidth}{@{\extracolsep{\fill}}p{0.46\linewidth}rr@{}}
      \toprule
      \multicolumn{3}{c}{\thead{Upper garment (1/2)}} \\
      \midrule
      \thead{Garment type} & \thead{Count} & \thead{Share} \\
      \midrule
      Blouse & 2,121 & 8.84\% \\
      Sweater & 2,016 & 8.40\% \\
      Button-up shirt & 1,965 & 8.19\% \\
      Cardigan & 1,806 & 7.52\% \\
      T-shirt & 1,638 & 6.83\% \\
      Vest & 1,439 & 6.00\% \\
      Tank top & 1,182 & 4.92\% \\
      Blazer & 1,155 & 4.81\% \\
      Biker jacket & 789 & 3.29\% \\
      Crop top & 771 & 3.21\% \\
      Camisole & 693 & 2.89\% \\
      Blanket coat & 537 & 2.24\% \\
      Suit jacket & 516 & 2.15\% \\
      Sweatshirt & 514 & 2.14\% \\
      Tunic & 513 & 2.14\% \\
      Hoodie & 489 & 2.04\% \\
      Halter top & 444 & 1.85\% \\
      Denim jacket & 444 & 1.85\% \\
      \bottomrule
    \end{tabular*}
  \end{minipage}\hfill%
  \begin{minipage}[t]{0.32\textwidth}
    \vspace{0pt}
    \begin{tabular*}{\linewidth}{@{\extracolsep{\fill}}p{0.46\linewidth}rr@{}}
      \toprule
      \multicolumn{3}{c}{\thead{Upper garment (2/2)}} \\
      \midrule
      \thead{Garment type} & \thead{Count} & \thead{Share} \\
      \midrule
      Peasant top & 426 & 1.77\% \\
      Bomber jacket & 396 & 1.65\% \\
      Pea coat & 354 & 1.47\% \\
      Track jacket & 342 & 1.43\% \\
      Trench coat & 342 & 1.43\% \\
      Wrap coat & 303 & 1.26\% \\
      Teddy coat & 297 & 1.24\% \\
      Puffer jacket & 279 & 1.16\% \\
      Parka & 279 & 1.16\% \\
      Polo & 273 & 1.14\% \\
      Cropped jacket & 261 & 1.09\% \\
      Puffer coat & 243 & 1.01\% \\
      Windbreaker & 234 & 0.97\% \\
      Varsity jacket & 207 & 0.86\% \\
      Kimono coat & 207 & 0.86\% \\
      Military coat & 186 & 0.78\% \\
      Duster coat & 177 & 0.74\% \\
      Shearling coat & 162 & 0.68\% \\
      \bottomrule
    \end{tabular*}
  \end{minipage}\hfill%
  \begin{minipage}[t]{0.32\textwidth}
    \vspace{0pt}
    \begin{tabular*}{\linewidth}{@{\extracolsep{\fill}}p{0.46\linewidth}rr@{}}
      \toprule
      \multicolumn{3}{c}{\thead{Lower garment}} \\
      \midrule
      \thead{Garment type} & \thead{Count} & \thead{Share} \\
      \midrule
      Trousers & 3,824 & 15.93\% \\
      Shorts & 3,288 & 13.70\% \\
      Jeans & 3,105 & 12.94\% \\
      Leggings & 2,517 & 10.49\% \\
      Bermuda shorts & 1,468 & 6.12\% \\
      Cropped pants & 1,388 & 5.78\% \\
      Lounge shorts & 1,362 & 5.67\% \\
      Culottes & 1,140 & 4.75\% \\
      Capri pants & 1,065 & 4.44\% \\
      Athletic shorts & 1,020 & 4.25\% \\
      Sweatpants & 954 & 3.98\% \\
      Cargo pants & 804 & 3.35\% \\
      Harem pants & 765 & 3.19\% \\
      Cargo shorts & 700 & 2.92\% \\
      Track pants & 600 & 2.50\% \\
      \bottomrule
    \end{tabular*}
  \end{minipage}
\end{table}

\FloatBarrier
\section{\proteus: Implementation Details}
\label{supp:impl}

\subsection{SMPL-X Fit}

We fit the SMPL-X body model to the cleaned outer surface $\mathcal{M}^{\mathrm{outer}}$ and triangulated 3D skeleton $\mathbf{J}^{\mathrm{op}}$.
We convert SMPL-X axis-angle joint rotations $\boldsymbol{\theta} \in \mathbb{R}^{55 \times 3}$ into a continuous 6D representation~\cite{zhou2019continuity} $\boldsymbol{\theta}_{\mathrm{6D}} \in \mathbb{R}^{55 \times 6}$ and jointly optimize pose and shape $\boldsymbol{\beta} \in \mathbb{R}^{10}$ under:
\begin{align}
\mathcal{L}_{\mathrm{smplx}}
= {}&
w_{\mathrm{joint}} \mathcal{L}_{\mathrm{joint}}
 + w_{\mathrm{cham}} \mathcal{L}_{\mathrm{cham}} + w_{\mathrm{sdf}} \mathcal{L}_{\mathrm{sdf}}
 + w_{\mathrm{hands}} \mathcal{L}_{\mathrm{hands}} + w_{\mathrm{shape}} \mathcal{L}_{\mathrm{shape}}.
\label{eq:smplx-fit}
\end{align}
We map the triangulated OpenPose keypoints $\mathbf{J}^{\mathrm{op}}$ to target SMPL-X joints $\mathbf{J}^\star\in\mathbb{R}^{55\times3}$ and penalize deviations from the posed SMPL-X joints $\mathbf{J}$ via a Huber loss $h_\delta$ with threshold $\delta$:
\begin{equation}
\mathcal{L}_{\mathrm{joint}}
=\sum_{j=1}^{55}\omega_j\,h_\delta
\!\left(\|\mathbf{J}_j-\mathbf{J}_j^\star\|_1\right),
\label{eq:smplx-joint-huber}
\end{equation}
where $\omega_j=1$ for joints with direct anatomical correspondence (shoulders, elbows, hips, knees, neck, and finger joints), $\omega_j=0$ for internal spine, collar, and pelvis joints lacking direct OpenPose counterparts, and head and toe targets are interpolated from OpenPose landmarks.

Fitting SMPL-X to 3D joints alone is under-constrained due to shape-skeleton ambiguity, so we incorporate surface terms relative to $\mathcal{M}^{\mathrm{outer}}$.
Let $\mathcal{M}^{\mathrm{body}}$ denote the posed SMPL-X mesh with vertices $\mathbf{V}^{\mathrm{body}}\in\mathbb{R}^{N_B\times3}$, and let $\mathcal{P}_B$ and $\mathcal{P}_T$ be point sets sampled uniformly on $\mathcal{M}^{\mathrm{body}}$ and $\mathcal{M}^{\mathrm{outer}}$, respectively.
With $d(\mathbf{x},\mathcal{M})$ denoting the point-to-mesh Euclidean distance, the bidirectional Chamfer term is
\begin{align}
\mathcal{L}_{\mathrm{cham}}
={}&\frac{1}{|\mathcal{P}_B|}\sum_{\mathbf{x}\in\mathcal{P}_B}d(\mathbf{x},\mathcal{M}^{\mathrm{outer}})^2 + \frac{1}{|\mathcal{P}_T|}\sum_{\mathbf{y}\in\mathcal{P}_T}d(\mathbf{y},\mathcal{M}^{\mathrm{body}})^2.
\label{eq:smplx-bidirectional-chamfer}
\end{align}
To keep the body inside the clothed surface without collapsing too far inward, let $\phi_i$ be the signed distance to $\mathcal{M}^{\mathrm{outer}}$ at body vertex $i$ (positive outside, in normalized coordinates with maximum target extent $2$) and $[z]_+=\max(z,0)$.
Using an outside margin $\tau_{\mathrm{out}}=0.01$ and an inside clearance $\tau_{\mathrm{in}}=0.01$, we penalize vertices that protrude past $-\tau_{\mathrm{out}}$ or lie deeper than $-\tau_{\mathrm{in}}$:
\begin{align}
\mathcal{L}_{\mathrm{sdf}}
={}&\frac{w_{\mathrm{out}}}{|\mathcal{V}_+|}\sum_{i\in\mathcal{V}_+}[\phi_i+\tau_{\mathrm{out}}]_+^2 + \frac{w_{\mathrm{in}}}{N_B}\sum_{i=1}^{N_B}[-\phi_i-\tau_{\mathrm{in}}]_+^2,
\label{eq:smplx-sdf}
\end{align}
where $\mathcal{V}_+=\{i:\phi_i>-\tau_{\mathrm{out}}\}$ is the set of active violating vertices, $w_{\mathrm{out}}=2\times 10^5$, and $w_{\mathrm{in}}=5\times 10^3$ (alongside $w_{\mathrm{joint}}=10^5$, $w_{\mathrm{cham}}=10^3$, $w_{\mathrm{sdf}}=1$, and $w_{\mathrm{shape}}=50$ for the $L_2$ shape prior $\mathcal{L}_{\mathrm{shape}} = \|\boldsymbol{\beta}\|_2^2$).

Finally, to prevent unnatural finger articulations when fitting noisy multi-view hand keypoints, we regularize the hand pose with three terms:
\begin{align}
\mathcal{L}_{\mathrm{hands}} = w_{\mathrm{hand-lap}}\mathcal{L}_{\mathrm{hand-lap}} + w_{\mathrm{bmc}}\mathcal{L}_{\mathrm{bmc}} + w_{\mathrm{euler}}\mathcal{L}_{\mathrm{euler}}.
\label{eq:smplx-hand}
\end{align}
First, a uniform graph Laplacian $L_{\mathrm{hand}}$ over the hand vertices suppresses ``candy-wrapper'' twist artifacts:
\begin{equation}
\mathcal{L}_{\mathrm{hand-lap}} = \|L_{\mathrm{hand}}\mathbf{V}^{\mathrm{body}}\|_F^2.
\label{eq:smplx-hand-lap}
\end{equation}
Second, a biomechanical constraint $\mathcal{L}_{\mathrm{bmc}}$~\cite{spurr2020weakly} keeps the hand skeleton within an anatomically feasible convex hull, and third, $\mathcal{L}_{\mathrm{euler}}$ penalizes joint rotations exceeding plausible Euler angle limits (with $w_{\mathrm{hand-lap}}=100$, $w_{\mathrm{bmc}}=10^{-3}$, and $w_{\mathrm{euler}}=0.1$).
\subsection{Non-Rigid Registration}
\paragraph{Optimization Schedule.}
The coarse ($10$K), mid ($41$K), and fine ($167$K) registration stages run for $1000$, $750$, and $500$ iterations with initial learning rates of $9\times10^{-3}$, $4\times10^{-3}$, and $2\times10^{-3}$, respectively.
Surface correspondences are updated every $2$, $2$, and $5$ iterations, and multi-view rendering losses are evaluated every $5$, $5$, and $10$ iterations with normal and alpha weights $\lambda_{\mathbf{N}}=0.9$ and $\lambda_{\mathbf{A}}=0.1$.
We optimize each stage using AdamW~\cite{loshchilov2017decoupled} with plateau-based learning-rate decay and early stopping.
During the coarse stage, canonical offsets are linearly ramped up over the first $500$ iterations to stabilize large initial deformations.
Correspondence search retains the nearest target point along each source vertex normal within $1\%$ of the active bounding-box diagonal, provided the source and target normals have a positive dot product.

\paragraph{Skin and Hair Regularization.}
Deviations from the underlying SMPL-X body should primarily reflect clothing rather than tangential drift over exposed skin.
We therefore penalize canonical offsets on template vertices whose matched target point $\mathbf{y}_i$ carries a skin or hair label $\ell(\mathbf{y}_i)$:
\begin{align}
\mathcal{L}_{\mathrm{sel}}
&=\frac{\sum_i a_i\|\Delta_i\|_2^2}{\sum_i a_i+\varepsilon},
\label{eq:target-aware-offset-loss}\\
a_i&=\mathbf{1}[\ell(\mathbf{y}_i)\in\{\mathrm{skin},\mathrm{hair}\}],\notag
\end{align}
where $\varepsilon>0$ avoids division by zero.
This keeps exposed skin anchored near the SMPL-X body while allowing garments to deform freely.
We additionally restrict hair correspondences to the head region of the template.

\paragraph{ARAP Rotation Estimation.}
Evaluating the ARAP loss requires a per-vertex local rotation $R_i$, classically computed via SVD of the edge covariance matrix
\(
\sum_{j \in \mathcal{N}(i)} w_{ij}
\mathbf{e}_{ij}\tilde{\mathbf{e}}_{ij}^\top
\)
from reference edges $\mathbf{e}_{ij}$, deformed edges $\tilde{\mathbf{e}}_{ij}$, and cotangent weights $w_{ij}$.
Running per-vertex SVD at every iteration is prohibitive at $167$K vertices, so we instead extract the orthogonal factor of the polar decomposition using Higham's scaled Heron iteration~\cite{higham1986polar}.

\subsection{Registration Scoring and Filtering}
\label{supp:reg-filter}
To automatically remove registration outliers at scale, we evaluate ten per-sample quality statistics spanning five complementary groups (\cref{tab:supp-reg-filter-metrics}) before assembling the final \proteus dataset.

\begin{table}[!tbp]
  \centering
  \caption{\textbf{Registration filtering statistics and thresholds.} Each rule rejects the indicated tail of a per-sample statistic using a shared calibrated rank cutoff (\cref{eq:supp-worst-tail}). Medians are computed over the $60$K scored registrations of the six generation chunks in Part~2 of \proteus, and thresholds report the cutoff range across these chunks. Surface distances use coordinates normalized to a maximum target extent of $2$, and offsets are normalized by the mean active edge length $\bar\ell$.}
  \label{tab:supp-reg-filter-metrics}
  \scriptsize
  \setlength{\tabcolsep}{3pt}
  \fitwidth{\begin{tabular}{@{}lllp{6.2cm}rr@{}}
    \toprule
    \thead{Group} & \thead{Statistic} & \thead{Tail} & \thead{Typical artifact} & \thead{Median} & \thead{Threshold} \\
    \midrule
    Surface distance & Source-to-target Chamfer, $p95$ & high & Template surface where the target has none: missing or truncated target limbs, pose mismatch & 0.012 & 0.086--0.088 \\
     & Target-to-source Chamfer, $p95$ & high & Uncovered target geometry: loose, long or open garments; unfollowed limbs & 0.014 & 0.112--0.114 \\
    \midrule
    Edge deformation & Edge log-stretch, median & both & High: globally inflated clothing; low: registration stays at the template & 0.157 & $<$0.079--0.083, $>$0.211--0.214 \\
     & Edge log-stretch, $p95$ & high & Local tearing or bridging, \eg between legs or under arms & 0.668 & 0.755--0.761 \\
    \midrule
    Offset deformation & Offset norm $\|\Delta_i\|_2/\bar\ell$, $p95$ & high & Large displacements from the body & 15.1 & 20.1--20.3 \\
     & Cotangent offset energy $E_{\cot}$ & high & Crumpled or folded offset field close to the target & 0.050 & 0.103--0.106 \\
    \midrule
    Correspondence & Valid correspondence fraction & low & Global misfit: wrong scale or misaligned SMPL-X fit & 0.786 & 0.511--0.524 \\
     & Normal angle, $p95$ (degrees) & high & Inverted or folded surface patches & 77.5 & 82.8--83.0 \\
    \midrule
    Multiview & Normal-map SSIM & low & Extended shape mismatch, \eg missing sleeves or wrong garment outline & 0.873 & 0.705--0.711 \\
     & Silhouette IoU & low & Coverage mismatch & 0.921 & 0.766--0.774 \\
    \bottomrule
  \end{tabular}}
\end{table}

\begin{figure}[!tbp]
  \centering
  \includegraphics[width=0.60\linewidth]{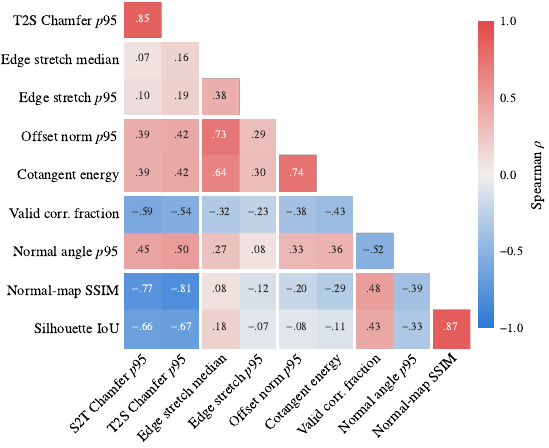}
  \caption{\textbf{Rank correlations across registration filtering statistics.} Spearman correlations among the ten filtering statistics over the $60$K scored registrations of Part~2 of \proteus. White lines separate the metric groups of \cref{tab:supp-reg-filter-metrics}: surface distance, edge deformation, offset deformation, correspondence quality, and multi-view quality (S2T: source-to-target; T2S: target-to-source).}
  \label{fig:supp-reg-filter-corr}
\end{figure}

\paragraph{Surface distance.}
We sample $10^5$ points uniformly by triangle area on the active region (excluding hands and feet) of both the registered mesh and the target surface $\mathcal{M}^{\mathrm{outer}}$, normalized to a maximum bounding-box extent of $2$.
We record the 95th-percentile ($p95$) closest-point distances in both directions, as $p95$ captures localized failures that mean Chamfer distance dilutes.
A high source-to-target distance indicates template geometry placed where no target surface exists (\eg, a missing or truncated limb in the lifted asset, or a pose mismatch), whereas a high target-to-source distance indicates uncovered target geometry (\eg, loose garments or unfollowed limbs).
\paragraph{Deformation.}
To detect intrinsic mesh distortion independent of target proximity, we evaluate three metrics directly on the canonical offset field.
Letting $\bar\ell$ denote the mean edge length of the active canonical template, we measure the $p95$ normalized offset magnitude $\|\Delta_i\|_2/\bar\ell$ and the per-vertex edge log-stretch:
\begin{equation}
s_i=\frac{1}{|\mathcal{N}(i)|}\sum_{j\in\mathcal{N}(i)}\left|\log\frac{\|(\mathbf{v}_j+\Delta_j)-(\mathbf{v}_i+\Delta_i)\|_2}{\|\mathbf{v}_j-\mathbf{v}_i\|_2}\right|.
\label{eq:supp-edge-log-stretch}
\end{equation}
We filter both tails of the per-sample median of $s_i$: excessively high values indicate globally inflated garments, while abnormally low values indicate a registration that failed to deform away from the base template.
The $p95$ of $s_i$ captures localized tearing or bridging (\eg, triangles stretched across the legs or armpits).
Finally, the area-normalized cotangent offset Dirichlet energy,
\begin{equation}
E_{\cot}=\frac{1}{A}\sum_{(i,j)\in\mathcal{E}}w_{ij}\|\Delta_i-\Delta_j\|_2^2,
\label{eq:supp-cot-energy}
\end{equation}
where $A$ is the total active surface area, detects high-frequency crumpling and self-folding that may otherwise achieve low point-to-surface distance.

\paragraph{Correspondence quality.}
We measure the fraction of active template vertices that find a valid normal-compatible target match within $1\%$ of the active bounding-box diagonal, along with the $p95$ angle between matched source and target normals.
A low valid match fraction flags global scale or pose misfits, while a high $p95$ normal angle detects locally inverted or folded surface patches.

\paragraph{Multi-view quality.}
We render $1024\times1024$ normal maps and silhouettes of the registered and target meshes from 8 azimuths ($45^\circ$ apart at zero elevation), cropped to the target bounding box with hands and feet masked out, and average normal-map SSIM and silhouette IoU across views.

\paragraph{Reporting subset.}
All filtering statistics in this section are computed on a subset of \proteus: the six generation chunks of Part~2 (\cref{supp:models-data}), comprising $60$K scored registrations.
Chunks generated after them are filtered with the same procedure, calibrated within each chunk as described below.

\paragraph{Correlations.}
\Cref{fig:supp-reg-filter-corr} shows the Spearman rank correlations among all ten statistics across these $60$K registrations.
Surface distance and multi-view metrics strongly correlate ($|\rho|$ up to $0.81$), as both measure target fidelity.
In contrast, intrinsic deformation metrics correlate strongly with one another ($\rho=0.64\text{--}0.74$) but weakly with target agreement ($|\rho|\le 0.42$), indicating that a registration can lie close to the target surface while still suffering from severe topological distortion.

\paragraph{Calibrated tail rejection.}
The ten statistics have different units and tail distributions, so we combine them via empirical rank calibration within each generation chunk ($8\text{--}10$K samples from a single 3D-lifting backend).
Let $u_{ri}\in(0,1)$ be the empirical CDF of statistic $r$ for sample $i$ within its chunk.
Depending on whether rule $r$ rejects the upper, lower, or both tails, we define its tail score as $b_{ri}=u_{ri}$, $1-u_{ri}$, or $2|u_{ri}-\tfrac12|$, and assign each sample its worst score across all rules:
\begin{equation}
b_i=\max_r b_{ri}.
\label{eq:supp-worst-tail}
\end{equation}
Rejecting the top $5\%$ of samples with the largest $b_i$ in each chunk is equivalent to thresholding every statistic at a shared per-rule tail mass $\tau \approx 0.73\text{--}0.75\%$.
Across chunks, the resulting metric cutoffs vary by at most $4\%$ (\cref{tab:supp-reg-filter-metrics}).

\paragraph{Rejection budget and outcome.}
We set the $5\%$ rejection budget after visually inspecting boundary samples at $2.5\%$, $5\%$, and $10\%$ cutoffs on an initial $30$K-sample run.
The $5\%$ cutoff isolates severe failures (truncated limbs, pose misfits, bridged geometry), whereas samples just above it are visually comparable to accepted registrations.
Across the $60$K scored registrations of the reporting subset, filtering removes $3$K samples ($5.0\%$) and retains $57$K.
In total, $82\%$ of rejected assets are flagged by only one of the five metric groups (824 by edge stretch, 818 by correspondence quality, 770 by offset distortion, 657 by surface distance, and 580 by multi-view agreement), indicating that the five criteria capture distinct failure modes.

\FloatBarrier
\section{\proteus Benchmarks}
\label{supp:data-bench}

This section presents the quantitative and qualitative benchmarks motivating our model choices across the stages of \proteus: text-to-image generation, image-to-3D lifting, SMPL-X fitting, and non-rigid registration.

While multi-view diffusion and video generation models can also synthesize 3D human priors, open-source multi-view models still suffer from cross-view geometric inconsistencies, and state-of-the-art video generators remain prohibitively expensive at the scale of $100$K+ assets.
We therefore adopt single-image 3D lifting models, which directly output explicit 3D meshes suitable for multi-view annotation and surface registration.

For 2D keypoint detection and semantic parsing, we build directly on established foundation models without separate benchmarking:
\paragraph{2D/3D keypoint estimation.}
Whole-body 2D pose estimation has converged on the OpenPose/COCO-WholeBody keypoint layout~\cite{cao2018openpose,hidalgo2019singlenetwork,jin2020whole}.
We run the ViT-based OpenPose-134 detector from RTMlib~\cite{jiang2023rtmlib} across multi-view renders of each cleaned 3D asset and triangulate the 2D detections into 3D using known camera parameters and RANSAC outlier rejection.
\paragraph{Human segmentation.}
While general-purpose promptable segmenters such as SAM~\cite{kirillov2023segany,ravi2024sam2segmentimages,carion2025sam3segmentconcepts} require separate text-prompted passes per garment category, Sapiens2~\cite{khirodkar2024sapiens,khirodkar2026sapiens2} predicts all 28 human body and clothing classes plus background in a single forward pass, making it well suited for dense 50-view mesh annotation.
\subsection{Image Generation}
\begin{figure}[!tbp]
  \centering
  \includegraphics[width=\linewidth]{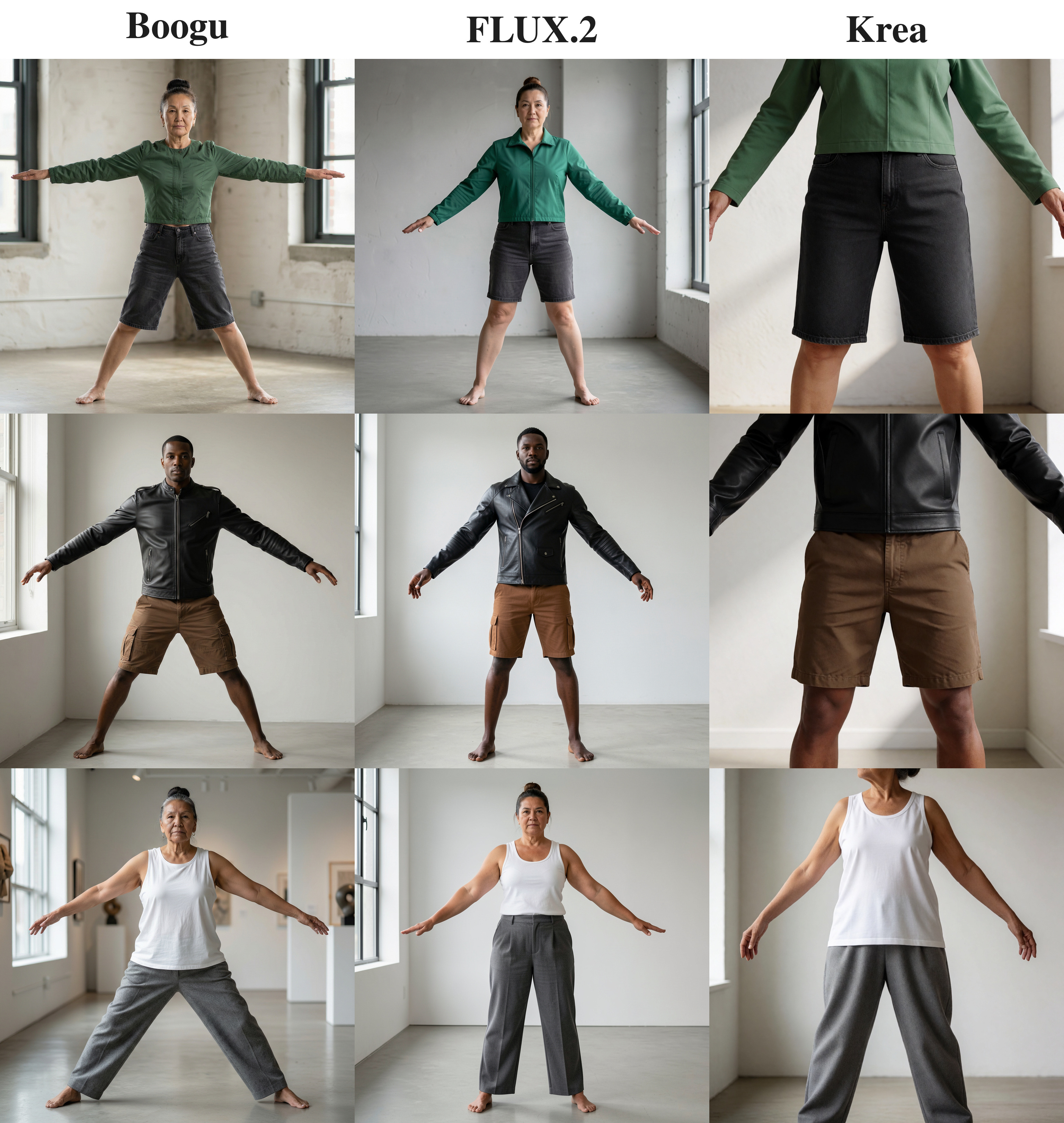}
  \caption{\textbf{Qualitative comparison of text-to-image backends.} Samples generated from identical structured prompts by Boogu-Image-Turbo, FLUX.2-klein-9B, and Krea-2-Turbo.}
  \label{fig:supp-t2i-bench-examples}
\end{figure}

We benchmark the three highest-ranking open-source text-to-image models on the \href{https://imagebench.ai/}{ImageBench} leaderboard (as of August 2026) capable of generating a full-body image in under 8 seconds on a single modern GPU: FLUX.2-klein-9B~\cite{flux-2-2025}, Krea-2-Turbo~\cite{krea2026krea2}, and Boogu-Image-Turbo~\cite{chen2026booguimage}.
For \proteus, the generator must faithfully render the prompted identity and outfit attributes while producing a full-body figure in a relaxed A-pose without cropped limbs or severe self-occlusions.
\paragraph{Evaluation set.}
We sample 500 diverse prompts from the \proteus attribute distribution and generate 4 images per prompt (with fixed random seeds) for each candidate model.
\paragraph{Metrics.}
We evaluate the generated images across three complementary criteria:
\begin{itemize}
    \item \textbf{VQA-based prompt adherence:} TIFA~\cite{hu2023tifa} decomposes each prompt into atomic question-answer pairs verified by BLIP-VQA~\cite{li2022blip}. GenEval2~\cite{kamath2025geneval} applies the same decomposition using Qwen3-VL~\cite{Qwen3-VL} log-probabilities, and we additionally evaluate a Gemma4~\cite{gemmateam2026gemma4} VQA judge to reduce single-VLM bias.
    \item \textbf{Global prompt-image alignment:} We report the CLIPScore~\cite{hessel2021clipscore,radford2021clip} (CLIP ViT-B/32 cosine similarity) from T2I-CompBench++~\cite{huang2025t2icompbenchpp} and the CLIP-T score from DreamBench++~\cite{peng2024dreambench}.
    \item \textbf{Human preference:} We compute HPSv3~\cite{Ma_2025_ICCV}, a learned Qwen2-VL~\cite{Qwen2-VL} reward model trained to predict human visual preference.
\end{itemize}
\paragraph{Results.}
\Cref{tab:supp-t2i-aggregate} reports aggregate metrics, \cref{tab:supp-t2i-structured} provides per-attribute VQA breakdowns, and \cref{fig:supp-t2i-bench-examples} shows qualitative examples.
\begin{table}[!tbp]
  \centering
  \caption{\textbf{Aggregate text-to-image benchmark scores.} Comparison of FLUX.2-klein-9B, Boogu-Image-Turbo, and Krea-2-Turbo across prompt adherence and aesthetic reward metrics.}
  \label{tab:supp-t2i-aggregate}
  \setlength{\tabcolsep}{3.2pt}
  \fitwidth{%
    \begin{tabular}{llccc}
      \toprule
      \thead{Benchmark} & \thead{Metric} & \thead{FLUX.2} & \thead{Boogu} & \thead{Krea} \\
      \midrule
      TIFA & exact-answer fraction & 0.9660 & \fst{0.9822} & 0.9628 \\
      TIFA atomic & exact-answer fraction & 0.7634 & \fst{0.7817} & 0.7308 \\
      CompBench component & CLIP cosine & \fst{0.3218} & 0.3208 & 0.3092 \\
      DreamBench++ component & CLIP-T & \fst{32.1776} & 32.0780 & 30.9222 \\
      HPSv3 & reward $\mu$ & \fst{8.8356} & 7.8857 & 7.1491 \\
      GenEval2 & Soft-TIFA AM & 75.3421 & \fst{78.4643} & 60.1633 \\
      GenEval2 atomic, Qwen3 & Soft-TIFA AM & 93.4381 & \fst{93.8693} & 86.8165 \\
      GenEval2 atomic, Gemma4 & Soft-TIFA AM & \fst{92.3074} & 92.1781 & 81.7964 \\
      \bottomrule
    \end{tabular}%
  }
\end{table}

Despite strong prompt adherence on upper-body clothing, Krea-2-Turbo frequently generates torso close-ups with cropped legs, rendering its outputs unusable for full-body 3D lifting.
Boogu-Image-Turbo and FLUX.2-klein-9B achieve comparable overall performance: Boogu-Image-Turbo edges ahead on atomic VQA checks, whereas FLUX.2-klein-9B leads in CLIP alignment and human preference (HPSv3) while introducing natural pose variation around the A-pose. We therefore select FLUX.2-klein-9B for \proteus.
\begin{table}[!tbp]
  \centering
  \caption{\textbf{Per-attribute TIFA-atomic scores.} Exact-answer rates for human body and garment attributes queried via binary visual questions.}
  \label{tab:supp-t2i-structured}
  \setlength{\tabcolsep}{3.2pt}
  \fitwidth{%
    \begin{tabular}{lccc}
      \toprule
      \thead{Question} & \thead{FLUX.2} & \thead{Boogu} & \thead{Krea} \\
      \midrule
      Full body & 1.000 & 1.000 & 0.999 \\
      Front facing & \fst{1.000} & 0.977 & 0.850 \\
      Two arms & 0.978 & \fst{0.992} & 0.990 \\
      Two legs & 0.189 & \fst{0.546} & 0.112 \\
      Hands and feet & 0.614 & \fst{0.748} & 0.326 \\
      Both garments visible & \fst{0.516} & 0.187 & 0.326 \\
      Upper garment color & 0.998 & 0.995 & \fst{1.000} \\
      Upper garment type & 0.871 & \fst{0.891} & 0.822 \\
      Upper garment pattern & 0.647 & \fst{0.815} & 0.622 \\
      Lower garment color & \fst{0.981} & 0.981 & 0.967 \\
      Lower garment type & 0.492 & 0.450 & \fst{0.515} \\
      Lower garment pattern & 0.297 & \fst{0.359} & 0.286 \\
      \bottomrule
    \end{tabular}%
  }
\end{table}

\subsection{3D Lifting}
We evaluate three open-source image-to-3D generators: TRELLIS.2~\cite{xiang2025trellis2}, Pixal3D~\cite{li2026pixal3d}, and Hunyuan3D-2.1~\cite{hunyuan3d2025hunyuan3d}.
TRELLIS.2 conditions on DINOv3 image tokens to generate sparse structure, shape, and texture latents decoded into an o-voxel representation with PBR materials.
This produces detailed non-watertight meshes that often contain internal geometry layers.
Pixal3D refines TRELLIS.2 with pixel-aligned image conditioning; both models bake their o-voxel colors and PBR materials into a UV texture.
In contrast, Hunyuan3D-2.1 decodes a shape latent into a neural SDF extracted via marching cubes~\cite{lorensen1987marchingcubes} to produce watertight meshes, textured via multi-view diffusion.
\paragraph{Evaluation protocol.}
\begin{table}[!tbp]
  \centering
  \caption{\textbf{Quantitative comparison of image-to-3D lifting backends.} Results averaged over all 290 real scans across five datasets. Chamfer distance and F1 are computed in normalized coordinates, and multi-view metrics are averaged over 8 rendered views. We highlight the \ffst{best}, \fsnd{second-best}, and \ftrd{third-best} scores.}
  \label{tab:lifting_backend_comparison}
  \setlength{\tabcolsep}{3.2pt}
  \fitwidth{%
    \begin{tabular}{lccccccccccccc}
      \toprule
      & \multicolumn{4}{c}{\thead{3D Metrics}}
      & \multicolumn{4}{c}{\thead{Multi-view RGB}}
      & \multicolumn{5}{c}{\thead{Multi-view Normals}} \\
      \cmidrule(lr){2-5} \cmidrule(lr){6-9} \cmidrule(lr){10-14}
      \thead{Lifting backend}
      & \thead{CD ($\times 10^4$) $\downarrow$}
      & \thead{F1@0.01 $\uparrow$}
      & \thead{NC $\uparrow$}
      & \thead{Voxel IoU $\uparrow$}
      & \thead{PSNR $\uparrow$}
      & \thead{SSIM $\uparrow$}
      & \thead{LPIPS $\downarrow$}
      & \thead{FID-DINOv2 $\downarrow$}
      & \thead{Silhouette IoU $\uparrow$}
      & \thead{PSNR $\uparrow$}
      & \thead{SSIM $\uparrow$}
      & \thead{LPIPS $\downarrow$}
      & \thead{FID-DINOv2 $\downarrow$} \\
      \midrule
      Pixal3D   & \snd{208} & \snd{0.7763} & \snd{0.8104} & \snd{0.3991} & \fst{21.998} & \snd{0.8996} & \fst{0.1117} & \fst{43.462} & \trd{0.8533} & \fst{20.274} & \snd{0.8835} & \fst{0.1125} & \fst{31.284} \\
      Hunyuan3D & \trd{224} & \trd{0.7666} & \fst{0.8224} & \trd{0.3845} & \trd{19.557} & \trd{0.8876} & \trd{0.1375} & \trd{74.218} & \snd{0.8589} & \trd{19.805} & \fst{0.8861} & \trd{0.1224} & \trd{49.746} \\
      TRELLIS.2 & \fst{195} & \fst{0.7879} & \trd{0.8045} &\fst{0.4025} & \snd{21.446} & \fst{0.9007} & \snd{0.1166} & \snd{50.490} & \fst{0.8640} & \snd{19.814} & \trd{0.8823} & \snd{0.1187} & \snd{38.399} \\
      \bottomrule
    \end{tabular}%
  }
\end{table}

\begin{figure}[!tbp]
  \centering
  \includegraphics[width=\textwidth]{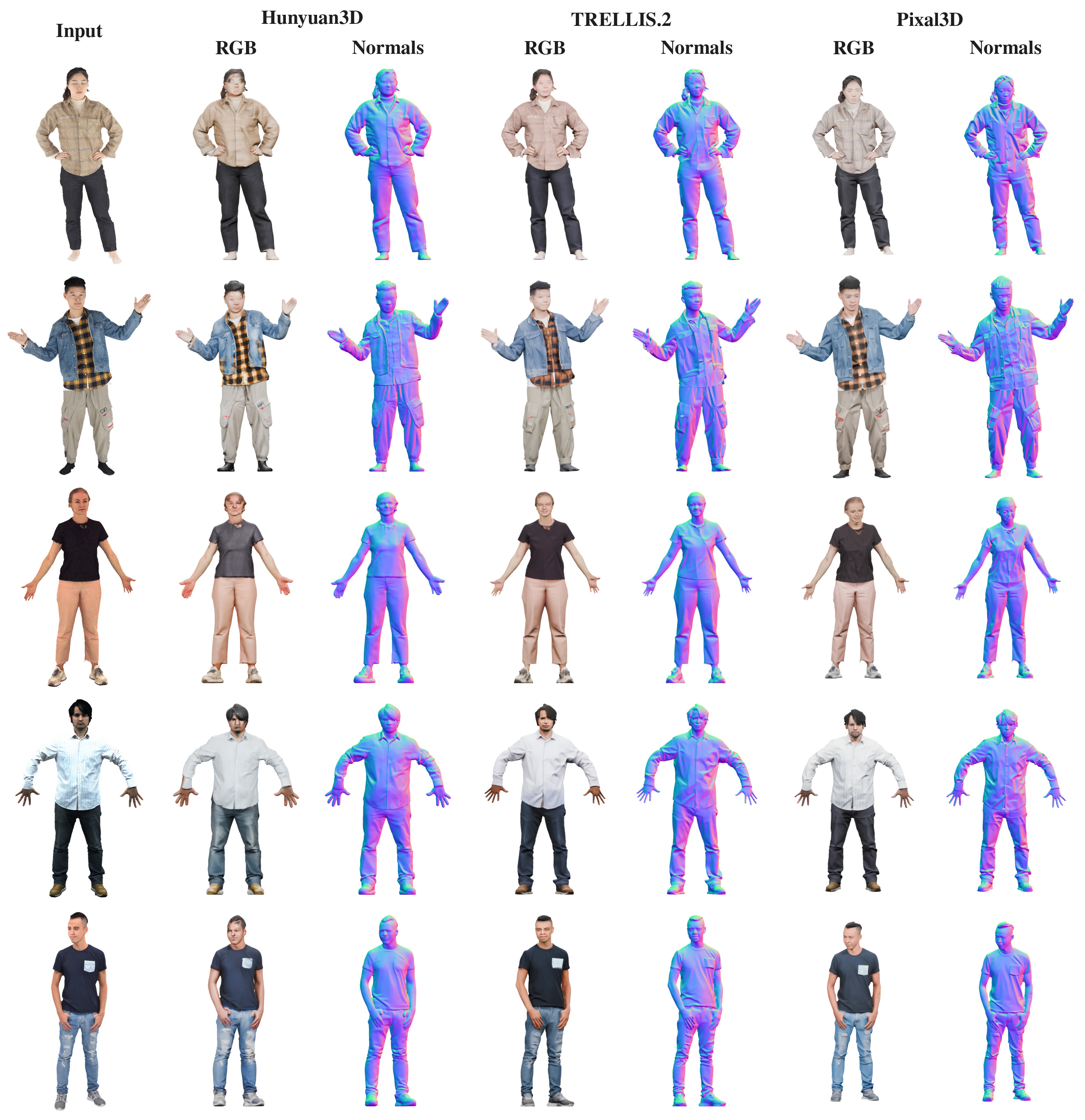}
  \caption{\textbf{Qualitative comparison of image-to-3D lifting backends.} Surface normal renders of raw 3D meshes reconstructed from frontal images by Pixal3D~\cite{li2026pixal3d}, Hunyuan3D-2.1~\cite{hunyuan3d2025hunyuan3d}, and TRELLIS.2~\cite{xiang2025trellis2}.}
  \label{fig:supp-3d-lifting-backends}
\end{figure}

We evaluate 3D reconstruction quality across 290 real clothed human scans curated from five datasets:
\begin{itemize}
    \item \textbf{4D-DRESS}~\cite{wang20244ddress}: 36 first-frame scans (typically in a relaxed A-pose) featuring layered everyday outfits, excluding dresses, skirts, and long coats.
    \item \textbf{THuman2.1}~\cite{tao2021function4d}: 46 high-resolution textured scans covering diverse, non-canonical poses.
    \item \textbf{CloSe-Di}~\cite{antic2024close}: 78 A-pose scans with one randomly selected outfit per subject.
    \item \textbf{2K2K}~\cite{han2023high}: 34 high-fidelity adult scans spanning diverse ages and ethnicities without object interactions.
    \item \textbf{ProbeSDF-Humans}: 96 in-house multi-view captures reconstructed with ProbeSDF~\cite{toussaint2025probesdf} in calibration A/T-poses across tight and loose garments.
\end{itemize}
Each scan is rendered from a frontal camera with a narrow field of view (so the subject spans $95\%$ of the $1024\times1024$ frame) and lifted to 3D by each model.
Ground-truth scans are normalized to $[-1, 1]^3$, and each lifted asset is normalized to the same bounding box and aligned via ICP~\cite{besl1992icp}.
We report three sets of metrics:
\begin{itemize}
    \item \textbf{3D geometry:} Chamfer distance (CD), F1-score, normal consistency (NC), and volumetric IoU on voxelized meshes.
    \item \textbf{Multi-view rendering:} PSNR, SSIM~\cite{wang2004ssim}, LPIPS~\cite{zhang2018lpips}, silhouette IoU, and DINOv2-based~\cite{oquab2023dinov2} FID~\cite{heusel2017fid} over 8 fixed views for both RGB appearance and surface normal maps.
    \item \textbf{Image-to-3D alignment:} ULIP-2~\cite{xue2023ulip2} and Uni3D~\cite{zhou2023uni3d} embedding similarities between the 3D asset and input image.
\end{itemize}
\paragraph{Results.}
\Cref{tab:lifting_backend_comparison} summarizes the geometric and multi-view benchmark, \cref{tab:lifting_encoder_comparison} reports ULIP-2 and Uni3D alignment, and \cref{fig:supp-3d-lifting-backends} compares qualitative renders.
\begin{table}[!tbp]
  \centering
  \caption{\textbf{Image-to-3D perceptual alignment.} Cosine similarity between conditioning images and lifted point clouds under Uni3D and ULIP-2 encoders.}
  \label{tab:lifting_encoder_comparison}
  \small
  \setlength{\tabcolsep}{5pt}
  \begin{tabular}{lccc}
    \toprule
    \thead{Encoder} & \thead{Pixal3D} & \thead{Hunyuan3D} & \thead{TRELLIS.2} \\
    \midrule
    Uni3D  & 0.213 & 0.217 & 0.214 \\
    ULIP-2 & 0.265 & 0.275 & 0.275 \\
    \bottomrule
  \end{tabular}
\end{table}

The three generators recover 3D human geometry with complementary strengths.
Hunyuan3D-2.1 produces watertight surfaces and preserves front-view lighting, though marching-cubes extraction and low-resolution multi-view texture baking smooth out fine geometric wrinkles and facial texture detail.
Conversely, the o-voxel models (TRELLIS.2 and Pixal3D) capture sharper geometric creases and garment folds, with Pixal3D exhibiting slightly stronger alignment to fine input-image features.
Because all three models synthesize plausible clothed geometry while differing in triangulation and texture characteristics, we use an equal $1/3$ split across all three backends in \proteus to maximize diversity and mitigate single-model bias.

\subsection{SMPL-X Fit}
\begin{figure}[!tbp]
  \centering
  \includegraphics[width=\textwidth]{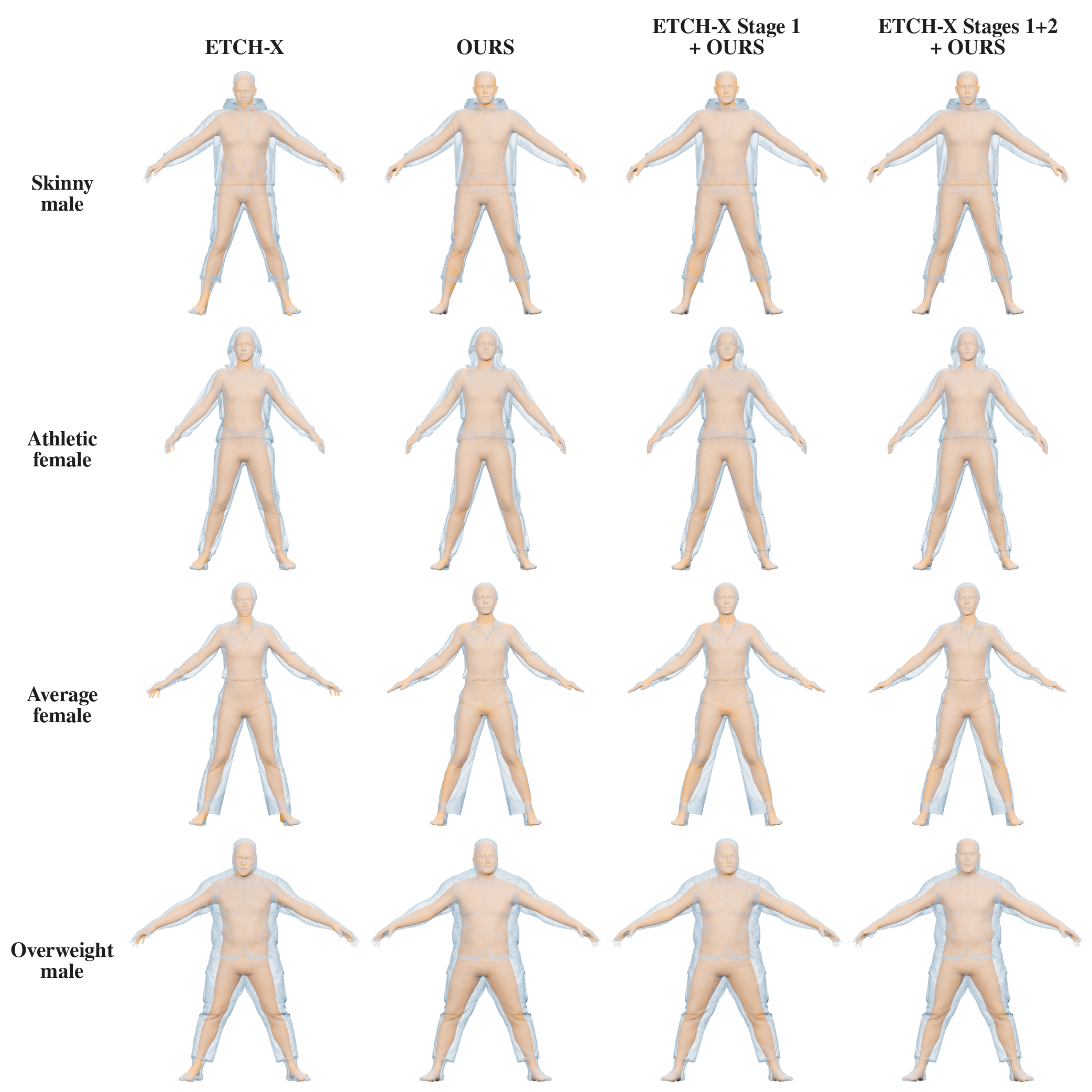}
  \caption{\textbf{Qualitative comparison of SMPL-X fitting across body compositions and genders.} Columns compare ETCH-X~\cite{li2026etchxrobustifyexpressivebody}, our standalone fitting pipeline used in \proteus, and our pipeline initialized with ETCH-X Stage~1 or Stages~1+2. Each fitted body (orange) is overlaid on the semi-transparent clothed target mesh (blue).}
  \label{fig:supp-smplx-fit-comparison}
\end{figure}

Fitting a parametric body to 3D skeletons alone~\cite{pavlakos2019smplx,pymafx2023,sarandi2024nlf,easymocap,cheng2023dna} leaves body shape under-constrained, whereas fitting directly to a clothed 3D surface~\cite{wang20244ddress,ho2024sith,ho2023custom} must contend with garment volume.
Recent methods such as ETCH~\cite{li2025etchgeneralizingbodyfitting} and ETCH-X~\cite{li2026etchxrobustifyexpressivebody} address clothed body fitting by predicting dense body-to-surface anchors and enforcing tightness-aware constraints.
\paragraph{Evaluation protocol.}
On a validation set of 120 \proteus assets spanning diverse genders, body shapes, and garment styles, we compare four configurations: standalone ETCH-X, our SMPL-X fitting procedure (\cref{supp:impl}), and two hybrid variants where our optimization is initialized or guided by ETCH-X Stage~1 or Stage~2 predictions.
We report bidirectional point-to-surface distances over the full mesh and over exposed skin regions, the fraction and depth of body vertices protruding outside the target surface, and the magnitude of the estimated shape vector $\boldsymbol\beta$.
\paragraph{Results.}
\Cref{tab:supp-smplx-fit} and \cref{fig:supp-smplx-fit-comparison} present quantitative and qualitative comparisons across the four fitting setups.
Our standalone fitting achieves the lowest full-surface and exposed-skin distances while maintaining low shape-parameter norms $\|\boldsymbol\beta\|_2$.
Incorporating ETCH-X Stage~1 guidance yields nearly identical metrics, while initializing from ETCH-X Stage~2 slightly reduces surface protrusion at the expense of larger skin distances and inflated shape norms.
We therefore use our standalone SMPL-X fitting pipeline throughout \proteus.
\begin{table}[!tbp]
  \centering
  \caption{\textbf{SMPL-X body fitting comparison.} Evaluation on 120 \proteus assets with distances and protrusion depths in normalized coordinates ($\times 10^3$). BD denotes the bidirectional nearest-neighbor distance, $P+$ is the fraction of vertices with positive target-mesh SDF, and $>1$ / $>5$ report exceedance rates at those thresholds. Skin metrics are restricted to visible skin, and shape statistics summarize the fitted $\boldsymbol{\beta}$ vector (lower is better).}
  \label{tab:supp-smplx-fit}
  \setlength{\tabcolsep}{3.2pt}
  \fitwidth{%
    \begin{tabular}{lccccccccc}
      \toprule
      & \multicolumn{2}{c}{\thead{Fit distance}}
      & \multicolumn{4}{c}{\thead{Containment / protrusion}}
      & \multicolumn{3}{c}{\thead{Shape code}} \\
      \cmidrule(lr){2-3} \cmidrule(lr){4-7} \cmidrule(lr){8-10}
      \thead{Configuration}
      & \thead{\shortstack{Full BD\\($\times 10^3$)}}
      & \thead{\shortstack{Skin BD\\($\times 10^3$)}}
      & \thead{\shortstack{Full protrusion\\depth ($\times 10^3$)}}
      & \thead{\shortstack{Full rate\\$P+$ / $>1$ / $>5$ (\%)}}
      & \thead{\shortstack{Skin protrusion\\depth ($\times 10^3$)}}
      & \thead{\shortstack{Skin rate\\$P+$ / $>1$ / $>5$ (\%)}}
      & \thead{\shortstack{$\|\boldsymbol{\beta}\|_2$\\mean / median}}
      & \thead{\shortstack{$\max|\beta_i|$\\mean / median}}
      & \thead{\shortstack{$\geq 3$ coeffs.\\mean / samples}} \\
      \midrule
      ETCH-X
        & 29.49
        & 11.70
        & 8.53
        & 15.65 / 13.88 / 8.29
        & 3.74
        & 18.01 / 13.84 / 4.85
        & 4.86 / 4.64
        & 2.96 / 2.67
        & 0.45 / 42/120 (35\%) \\
      OURS
        & \fst{25.80}
        & \fst{9.84}
        & 2.73
        & 11.73 / 8.71 / 2.26
        & 2.16
        & 16.08 / 10.95 / 2.02
        & \fst{1.37 / 1.32}
        & \fst{0.93 / 0.91}
        & 0.00 / 0/120 \\
      ETCH-X Stage 1 + OURS
        & 25.80
        & 9.88
        & 2.72
        & 11.70 / 8.66 / 2.23
        & 2.16
        & 16.11 / 10.95 / 2.01
        & 1.37 / 1.32
        & 0.93 / 0.91
        & 0.00 / 0/120 \\
      ETCH-X Stage 1 + Stage 2 + OURS
        & 26.79
        & 10.80
        & \fst{2.39}
        & \fst{9.13 / 6.36 / 1.29}
        & \fst{1.79}
        & \fst{12.10 / 7.79 / 1.01}
        & 2.46 / 2.37
        & 1.70 / 1.66
        & 0.00 / 0/120 \\
      \bottomrule
    \end{tabular}%
  }
\end{table}

\subsection{Non-Rigid Registration}
\begin{figure}[!tbp]
  \centering
  \includegraphics[width=\textwidth]{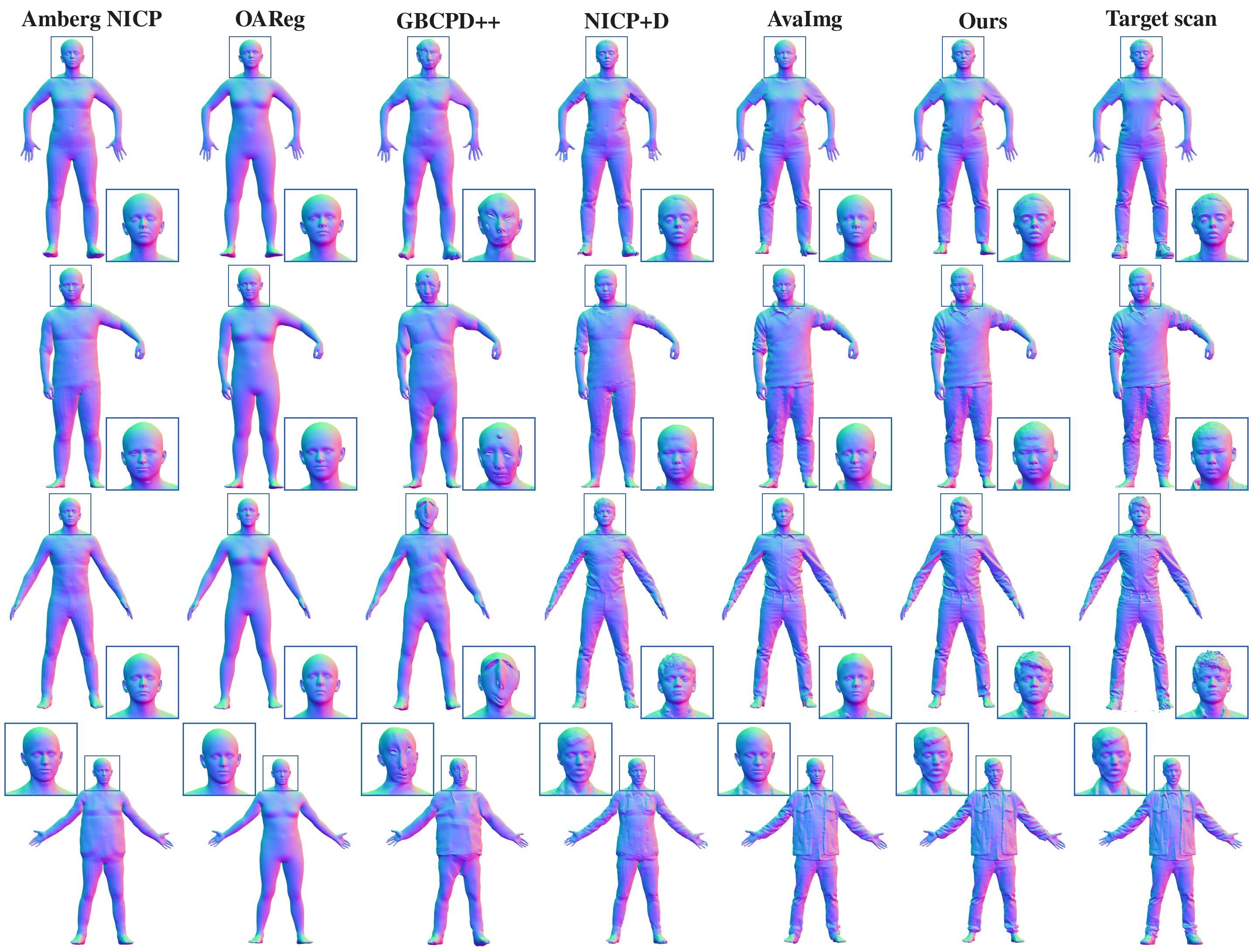}
  \caption{\textbf{Qualitative comparison of registration methods.} Surface normal renders across the four real scans shown in \cref{fig:registration-evaluation} for Amberg NICP, OAReg, GBCPD++, NICP+D, AvaImg, and our method, followed by the ground-truth scan. Insets magnify the head and collar region indicated on each body.}
  \label{fig:supp-registration-comparison}
\end{figure}

\Cref{fig:supp-registration-comparison} qualitatively compares our method against five registration baselines on the real scans of \cref{sec:results}: Amberg et al.~\cite{amberg2007optimal}, OAReg~\cite{zhao2025oareg}, GBCPD++~\cite{hirose2023geodesicbcpd}, NICP+D~\cite{marin2024nicp}, and AvaImg~\cite{kostyrko2026avaimg}.
Because the official AvaImg code is not publicly available, we evaluate our own reimplementation.

\begin{figure}[!tbp]
  \centering
  \includegraphics[width=\textwidth]{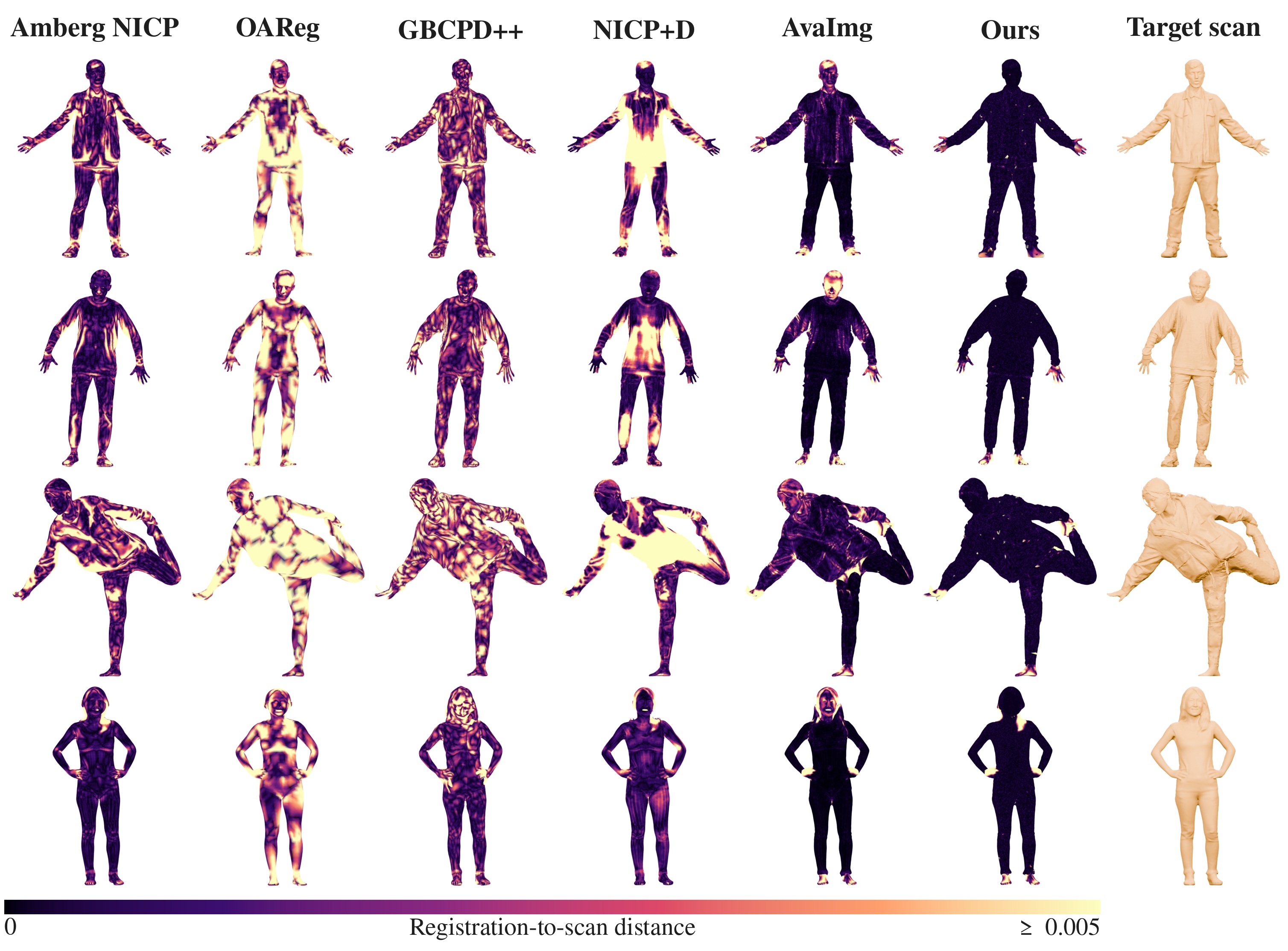}
  \caption{\textbf{Per-vertex registration error maps on real scans.} Results on 4D-DRESS, ProbeSDF, THuman2.1, and 2K2K (top to bottom), colored by unsquared Euclidean distance to the nearest target triangle in coordinates normalized by the maximum target extent (clipped at $0.005$). The rightmost column shows the untextured target scan.}
  \label{fig:supp-registration-error-maps}
\end{figure}

\Cref{fig:supp-registration-error-maps} visualizes per-vertex Euclidean registration error maps on real scans from 4D-DRESS, ProbeSDF, THuman2.1, and 2K2K.

\subsection{End-to-End Analysis}
\Cref{tab:real-scan-routes-full} extends the three-route real-scan evaluation of \cref{sec:results} with multi-view RGB and normal rendering metrics, and \cref{tab:real-scan-routes-supp} provides the full per-dataset breakdown.
\begin{table}[!tbp]
  \centering
  \caption{\textbf{Aggregate real-scan evaluation with rendered metrics.}
  Full version of \cref{tab:real-scan-routes-main}: R1 registers SMPL-X-D to a 3D-lifted frontal render of each scan, R2 to the scan itself, and R3 to the scan starting from the dataset-native SMPL-X fit.
  The \textbf{3D lifting} row compares the lifted assets to the scans, averaged over 3 lifting models.}
  \label{tab:real-scan-routes-full}
  \setlength{\tabcolsep}{3.0pt}
  \fitwidth{%
    \begin{tabular}{lllrrrrrrrrrr}
      \toprule
      & & & & \multicolumn{3}{c}{\thead{3D Metrics}} & \multicolumn{3}{c}{\thead{Multi-view RGB}} & \multicolumn{3}{c}{\thead{Multi-view Normals}} \\
      \cmidrule(lr){5-7} \cmidrule(lr){8-10} \cmidrule(lr){11-13}
      \thead{Route}
      & \thead{Target}
      & \thead{SMPL-X}
      & \thead{$N$}
      & \multicolumn{1}{c}{\thead{CD ($\times 10^4$) $\downarrow$}}
      & \multicolumn{1}{c}{\thead{F1@0.01 $\uparrow$}}
      & \multicolumn{1}{c}{\thead{NC $\uparrow$}}
      & \thead{PSNR $\uparrow$}
      & \thead{SSIM $\uparrow$}
      & \thead{LPIPS $\downarrow$}
      & \thead{PSNR $\uparrow$}
      & \thead{SSIM $\uparrow$}
      & \thead{LPIPS $\downarrow$} \\
      \midrule
      \textbf{R1} & Lifted surface & \proteus & 290 & 18.25 & 0.9562 & 0.9218 & 27.455 & 0.9422 & 0.0655 & 18.631 & 0.8754 & 0.1261 \\
      \textbf{R2} & GT scan        & \proteus & 290 & 14.67 & 0.9582 & 0.9150 & 24.603 & 0.9329 & 0.0569 & 21.282 & 0.8706 & 0.0963 \\
      \textbf{R3} & GT scan        & Native    & 116 & 13.47 & 0.9626 & 0.9237 & 24.388 & 0.9401 & 0.0526 & 21.831 & 0.8839 & 0.0881 \\
      \midrule
      \textbf{3D lifting} & GT scan & n/a & $290 \times 3$ & 81.14 & 0.7636 & 0.8124 & 21.000 & 0.8960 & 0.1219 & 19.961 & 0.8840 & 0.1179 \\
      \bottomrule
    \end{tabular}%
  }
\end{table}

\begin{table}[!tbp]
  \centering
  \caption{\textbf{Per-dataset breakdown of the real-scan evaluation.}
  Metrics and aggregation follow Tab.~\ref{tab:real-scan-routes-full}.
  R1 evaluates against the lifted outer surface; R2 and R3 evaluate against the prepared ground-truth scan.
  R3 is available only when the dataset provides native SMPL-X initialization.
  Highlighting marks the better of the directly comparable R2/R3 scan-target routes for datasets that support both.}
  \label{tab:real-scan-routes-supp}
  \setlength{\tabcolsep}{2.7pt}
  \fitwidth{%
    \begin{tabular}{llllrrrrrrrrrr}
      \toprule
      & & & & & \multicolumn{3}{c}{\thead{3D Metrics}} & \multicolumn{3}{c}{\thead{Multi-view RGB}} & \multicolumn{3}{c}{\thead{Multi-view Normals}} \\
      \cmidrule(lr){6-8} \cmidrule(lr){9-11} \cmidrule(lr){12-14}
      \thead{Dataset}
      & \thead{Route}
      & \thead{Target}
      & \thead{Initialization}
      & \thead{$N$}
      & \multicolumn{1}{c}{\thead{CD ($\times 10^4$) $\downarrow$}}
      & \multicolumn{1}{c}{\thead{F1@0.01 $\uparrow$}}
      & \multicolumn{1}{c}{\thead{NC $\uparrow$}}
      & \thead{PSNR $\uparrow$}
      & \thead{SSIM $\uparrow$}
      & \thead{LPIPS $\downarrow$}
      & \thead{PSNR $\uparrow$}
      & \thead{SSIM $\uparrow$}
      & \thead{LPIPS $\downarrow$} \\
      \midrule
      2K2K      & R1 & Lifted surface & Estimated & 34 & 19.48 & 0.9468 & 0.9142 & 24.804 & 0.9322 & 0.0767 & 17.381 & 0.8467 & 0.1501 \\
                 & R2 & GT scan        & Estimated & 34 & 18.37 & 0.9425 & 0.9010 & 23.811 & 0.9309 & 0.0564 & 21.018 & 0.8790 & 0.0983 \\
                 & R3 & GT scan        & Native    & 34 & \fst{15.39} & \fst{0.9552} & \fst{0.9056} & \fst{23.945} & \fst{0.9355} & \fst{0.0532} & \fst{21.483} & \fst{0.8816} & \fst{0.0929} \\
      \addlinespace
      4D-DRESS  & R1 & Lifted surface & Estimated & 36 & 17.94 & 0.9505 & 0.9206 & 25.645 & 0.9369 & 0.0671 & 19.069 & 0.8763 & 0.1205 \\
                 & R2 & GT scan        & Estimated & 36 & 19.01 & 0.9417 & 0.9257 & 24.140 & 0.9359 & 0.0534 & 21.232 & 0.8943 & 0.0780 \\
                 & R3 & GT scan        & Native    & 36 & \fst{16.39} & \fst{0.9513} & \fst{0.9312} & \fst{24.449} & \fst{0.9398} & \fst{0.0492} & \fst{21.436} & \fst{0.8980} & \fst{0.0745} \\
      \addlinespace
      CloSe-Di  & R1 & Lifted surface & Estimated & 78 & 15.70 & 0.9577 & 0.9145 & 25.073 & 0.9303 & 0.0747 & 19.372 & 0.8727 & 0.1259 \\
                 & R2 & GT scan        & Estimated & 78 & 14.24 & 0.9616 & 0.9019 & 22.544 & 0.9173 & 0.0716 & 21.168 & 0.8568 & 0.1119 \\
      \addlinespace
      ProbeSDF  & R1 & Lifted surface & Estimated & 96 & 22.00 & 0.9554 & 0.9316 & 31.907 & 0.9563 & 0.0519 & 18.181 & 0.8843 & 0.1221 \\
                 & R2 & GT scan        & Estimated & 96 & 13.21 & 0.9609 & 0.9223 & 26.769 & 0.9417 & 0.0471 & 21.054 & 0.8703 & 0.0883 \\
      \addlinespace
      THuman2.1 & R1 & Lifted surface & Estimated & 46 & 14.14 & 0.9671 & 0.9205 & 25.632 & 0.9446 & 0.0682 & 18.877 & 0.8824 & 0.1216 \\
                 & R2 & GT scan        & Estimated & 46 & 12.32 & 0.9709 & 0.9244 & 24.564 & 0.9402 & 0.0549 & 22.188 & 0.8705 & 0.0990 \\
                 & R3 & GT scan        & Native    & 46 & \fst{9.76} & \fst{0.9770} & \fst{0.9312} & \fst{24.667} & \fst{0.9437} & \fst{0.0547} & \fst{22.396} & \fst{0.8746} & \fst{0.0953} \\
      \bottomrule
    \end{tabular}%
  }
\end{table}

\subsection{Runtime}
\label{supp:pipeline-runtime}
\Cref{tab:supp-pipeline-runtime} reports the average time per sample for each \proteus stage, measured on the six generation chunks of Part~2 (\cref{supp:models-data}).
Each stage runs in parallel over sample batches on a heterogeneous compute cluster; we compute per-sample time as task wall time divided by batch size, averaged across the GPU types used for that stage (GPU runs only for GPU/CPU stages).
3D lifting accounts for $44\%$ of the $100.5$ GPU-seconds per sample, followed by registration ($23\%$) and mesh annotation ($12\%$).
Generating the $60$K raw samples required $1.7$K GPU-hours, or $106$ GPU-seconds per sample passing the final filter.
\begin{table}[!tbp]
  \centering
  \caption{\textbf{\proteus runtime per stage.} Average time per sample (seconds), including model loading and I/O. Different GPU types are used (H100, L40S, A40, RTX~8000). Registration runs $3$--$4$ samples concurrently per GPU. Stages marked GPU/CPU also run on CPUs; the reported times are from GPU runs.}
  \label{tab:supp-pipeline-runtime}
  \setlength{\tabcolsep}{4pt}
  \fitwidth{%
    \begin{tabular}{lllr}
      \toprule
      \thead{Stage} & \thead{Model} & \thead{Device} & \thead{Time (s)} \\
      \midrule
      Prompt sampling & -- & CPU & $<0.1$ \\
      Image generation & FLUX.2-klein-9B & GPU & 4.0 \\
      Image filtering & Grounded-SAM2 & GPU & 0.3 \\
      3D lifting & Pixal3D & GPU & 44.2 \\
      Outer surface & -- & GPU/CPU & 2.7 \\
      Mesh annotation & Sapiens2, RTMlib & GPU & 11.7 \\
      Triangulation & -- & GPU/CPU & 9.0 \\
      SMPL-X initialization & -- & GPU/CPU & 0.8 \\
      Registration & -- & GPU & 22.9 \\
      Registration scoring & -- & GPU & 4.8 \\
      Statistical filtering & -- & CPU & $<0.1$ \\
      \midrule
      Total & & & 100.5 \\
      \bottomrule
    \end{tabular}%
  }
\end{table}

\FloatBarrier
\section{Applications built on \proteus}
\label{supp:models}
This section provides architectural, training, and evaluation details for the downstream models presented in \cref{sec:application}.
We first describe the training data preparation, our UV-space conditional flow generator, the same-data and external reconstruction baselines, and additional results (\cref{supp:models-data,supp:models-ours,supp:models-samedata,supp:models-external,supp:models-eval,supp:models-results}).
We then detail the feed-forward point-cloud registration, 3D point-cloud segmentation, point-cloud SMPL-X regression, and dense image correspondence models (\cref{supp:models-pointcloud,supp:models-segmentation,supp:models-scan-smplx,supp:models-dense}).
\subsection{Training Data}
\label{supp:models-data}
\paragraph{Data splits.}
We generate the \proteus corpus in two parts.
\textbf{Part~1} comprises 99K assets (94K training, 2.5K validation, 2.5K test) with $512\times512$ random-view image renders.
It supervises all models in the same-data comparison (\cref{sec:app-architectures}), and a 50K-asset subset supervises point-cloud registration (\cref{supp:models-pointcloud}).
\textbf{Part~2} comprises 57K assets (51K training, 2.8K validation, 2.8K test) and supervises our model in the comparison against external reconstruction baselines (\cref{sec:app-image}) as well as all pose-varying tasks.
\paragraph{Canonical offset parameterization.}
Our registration optimizes canonical vertex offsets $\Delta$ relative to the shaped template $T(\hat{\boldsymbol\beta})$.
For learning, we re-express these offsets relative to the mean-shape template $T(\mathbf{0})$:
\begin{equation}
\Delta^{0} = \Delta + T(\hat{\boldsymbol\beta}) - T(\mathbf{0}),
\label{eq:zero-shape-offsets}
\end{equation}
where $T(\boldsymbol\beta)$ denotes the twice-subdivided SMPL-X rest-pose template under shape $\boldsymbol\beta$.
A single offset field $\Delta^0$ thus jointly encodes body shape and clothing geometry, and predicted offsets are posed via linear blend skinning using the SMPL-X pose parameters alone.
We normalize each offset coordinate channel by its training mean and three standard deviations; offsets on the eyes, hands, and feet are zeroed out and excluded from the normalization statistics and from all per-vertex regression losses.
\paragraph{UV baking.}
For UV-space architectures, per-vertex offsets are rasterized into a $256\times256$ displacement texture in the SMPL-X UV layout during data loading, and predicted displacement textures are bilinearly sampled back at the template vertex UV coordinates at inference time.
\paragraph{Point-cloud sampling and augmentation.}
To supervise point-cloud models, we pre-sample 128K points on the outer surface $\mathcal{M}^{\mathrm{outer}}$ of each asset and record their visibility across 32 virtual cameras.
During training, we draw 64K points (with RGB colors and normals) visible from a randomly chosen subset of 2, 4, 8, 16, or 32 cameras (the variable-density model in \cref{supp:models-pointcloud} instead samples between 8K and 64K points per batch).
To simulate real-world sensor noise and occlusions, we apply the following augmentations:
\begin{itemize}
    \item isotropic Gaussian point jitter with standard deviation up to 5\,mm;
    \item up to 5\% random outlier points;
    \item spherical cutouts removing up to 4 regions of radius 15\,cm;
    \item independent dropout of RGB colors and surface normals with probability 0.3.
\end{itemize}
To make point-cloud registration tolerant to imperfect body fits, we also perturb the conditioning SMPL-X parameters with Gaussian noise on body pose ($\sigma=0.03$), global orientation ($0.02$), translation ($0.01$), shape ($0.25$), and log-scale ($0.01$).
We then recompute the target offsets relative to the perturbed body so the ground-truth clothed surface remains unchanged.
\paragraph{Posed dataset.}
To supervise pose-varying tasks, we re-pose each of the 57K assets in Part~2 with 5 poses sampled from distinct clusters of an AMASS~\cite{mahmood2019amass} pose library (with random left-right mirroring), yielding 283K posed samples.
All poses of a given asset remain in the same data split, with validation and test poses drawn from held-out motion clusters.
We render each posed SMPL-X-D mesh from 4 views, storing RGB, alpha masks, and rasterized triangle indices and barycentric coordinates that map every foreground pixel to its canonical SMPL-X surface point.
Per-vertex Sapiens2~\cite{khirodkar2026sapiens2} semantic labels transferred from $\mathcal{M}^{\mathrm{outer}}$ to the registered SMPL-X-D mesh are inherited across all poses of an asset.

\FloatBarrier
\subsection{Conditional Generator}
\label{supp:models-ours}
\begin{table}[!tbp]
  \centering
  \caption{\textbf{Hyperparameters of our conditional generator.}
  Values are shared by the image- and point-cloud-conditioned models unless noted.}
  \label{tab:supp-pixeldit-config}
  \footnotesize
  \setlength{\tabcolsep}{4pt}
  \fitwidth{%
  \begin{tabular}{ll}
    \toprule
    \thead{Setting} & \thead{Value} \\
    \midrule
    Output & $256\times256\times3$ UV displacement texture \\
    Patch size / token grid & $8\times8$ / $32\times32$ \\
    Joint-attention blocks & 15, width 512, 8 heads, MLP ratio 4 \\
    Pixel-level blocks & 2, texel width 32 \\
    Image encoder & DINOv3 ViT-B/16, frozen, $512\times512$ input \\
    Shape condition & $\boldsymbol\beta\in\mathbb{R}^{10}$, added to timestep embedding \\
    Point-cloud condition & $64^2$ anchors, 128-ch.\ descriptor (+2 coverage), pooled to $32^2$ \\
    \midrule
    Objective & Rectified flow, uniform timesteps \\
    Optimizer & AdamW, lr $10^{-4}$, weight decay $10^{-4}$ \\
    Warm-up / gradient clipping & 10K steps / norm 1.0 \\
    Steps / batch size & 200K / 72 (image models) \\
    Condition dropout & 0.1 per condition \\
    Precision / hardware & bf16 / one NVIDIA H200 \\
    Sampling & Euler, 50 steps, no guidance \\
    \bottomrule
  \end{tabular}%
  }
\end{table}

\begin{figure}[!tbp]
  \centering
  \fitwidth{\begin{tikzpicture}[arch]
    \node[data] (xt) at (1.2,0) {Noisy UV texture $\mathbf{x}_t$\\[-1pt]{\scriptsize $256^2\times3$}};
    \node[module] (patch) at (4.1,0) {Patch embed\\[-1pt]{\scriptsize $8\times8$}};
    \node[module, minimum height=15mm, minimum width=30mm] (joint) at (8.3,0) {Joint-attention block\\[-1pt]{\scriptsize patch $\leftrightarrow$ image tokens}\\[-1pt]{\scriptsize AdaLN, 2D RoPE}};
    \node[dims, anchor=north east] at (joint.south east) {$\times15$};
    \node[module] (texel) at (11.2,0) {Texel-level\\[-1pt]block};
    \node[dims, anchor=north east] at (texel.south east) {$\times2$};
    \node[module] (linear) at (12.8,0) {Linear};
    \node[data] (vel) at (14.4,0) {Velocity $\mathbf{v}_\theta$\\[-1pt]{\scriptsize $256^2\times3$}};
    \node[loss] (lflow) at (16.4,0) {$\mathcal{L}_{\mathrm{flow}}$\\[-1pt]{\scriptsize Eq.~\eqref{eq:supp-flow-loss}}};
    \draw[flow] (xt) -- (patch);
    \draw[flow] (patch) -- node[dims, above] {$32^2$ tokens} (joint);
    \draw[flow] (joint) -- (texel);
    \draw[flow] (texel) -- (linear);
    \draw[flow] (linear) -- (vel);
    \draw[flow] (vel) -- (lflow);
    \node[data] (img) at (1.2,1.95) {Image $\mathbf{I}$\\[-1pt]{\scriptsize $512^2$}};
    \node[frozen] (dino) at (3.7,1.95) {DINOv3 ViT-B/16};
    \node[module] (proj) at (5.9,1.95) {Linear};
    \draw[flow] (img) -- (dino);
    \draw[flow] (dino) -- (proj);
    \draw[cond] (proj) -| node[dims, pos=0.75, right] {image tokens} (7.4,0.75);
    \node[op] (addc) at (11.2,1.2) {$+$};
    \node[data] (t) at (11.2,1.95) {$t$};
    \node[module] (mlp) at (12.5,1.2) {MLP};
    \node[data] (beta) at (13.6,1.2) {$\boldsymbol\beta$};
    \draw[cond] (t) -- (addc);
    \draw[flow] (beta) -- (mlp);
    \draw[cond] (mlp) -- (addc);
    \draw[cond] (addc) -| node[dims, pos=0.25, above] {AdaLN} (9.2,0.75);
    \draw[cond] (addc) -- (texel);
  \end{tikzpicture}}
  \caption{\textbf{Architecture of our conditional UV-space generator.} A PixelDiT denoiser predicts the rectified-flow velocity of a noisy $256\times256$ SMPL-X-D displacement texture. Frozen DINOv3 image tokens enter joint attention, and the timestep and body shape $\boldsymbol\beta$ modulate every block. Dashed modules are frozen.}
  \label{fig:supp-generator-arch}
\end{figure}
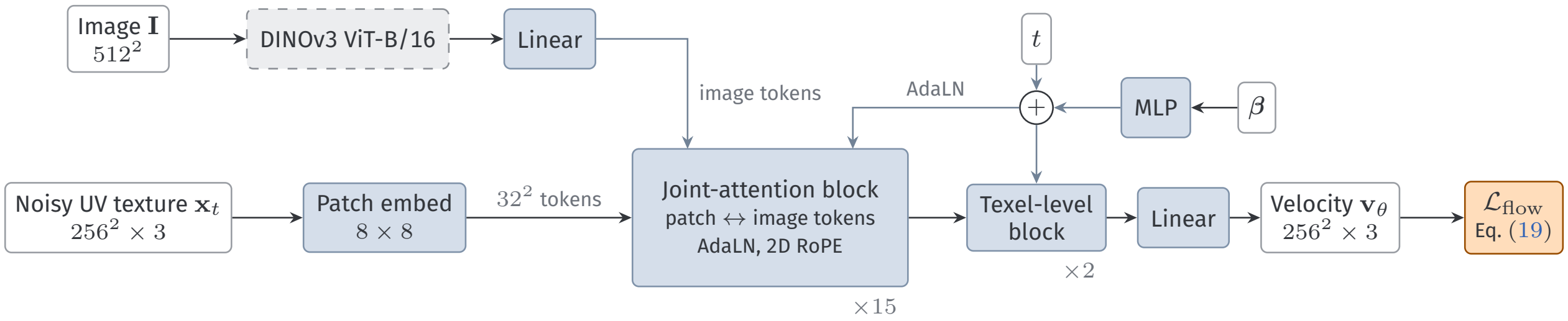

\paragraph{Architecture.}
Our UV displacement generator builds on PixelDiT~\cite{yu2026pixeldit} using the hyperparameters listed in \cref{tab:supp-pixeldit-config} (\cref{fig:supp-generator-arch}).
A patch embedder maps $8\times8$ texel patches of the noisy $256\times256$ displacement texture onto a $32\times32$ token grid.
A stack of joint-attention blocks then processes the patch tokens alongside conditioning tokens following MM-DiT~\cite{esser2024scaling}, with stream-specific adaptive layer normalization (AdaLN) and 2D rotary position embeddings (RoPE) on the patch tokens.
Two texel-level transformer blocks subsequently refine individual texel tokens modulated by their parent patch tokens, followed by a linear output layer that predicts the flow velocity at each texel.
\paragraph{Image and shape conditioning.}
A frozen DINOv3 ViT-B/16~\cite{simeoni2026dinov3} encodes the $512\times512$ input image into patch tokens that are linearly projected to the model dimension and passed to the joint-attention MM-DiT blocks.
A lightweight MLP embeds the 10 SMPL-X shape coefficients $\boldsymbol\beta$ and adds them to the diffusion timestep embedding that modulates every block.
\paragraph{Training and sampling.}
Let $\mathbf{x}_0$ denote the normalized $256\times256\times3$ displacement texture and $\mathbf{c}$ the conditioning signals.
We train with the rectified-flow objective~\cite{liu2023rectifiedflow,lipman2023flowmatching}, which interpolates linearly between data and Gaussian noise $\boldsymbol\epsilon\sim\mathcal{N}(\mathbf{0},\mathbf{I})$ and regresses the constant velocity of this path:
\begin{align}
\mathbf{x}_t&=(1-t)\,\mathbf{x}_0+t\,\boldsymbol\epsilon,\notag\\
\mathcal{L}_{\mathrm{flow}}
&=\mathbb{E}_{\mathbf{x}_0,\boldsymbol\epsilon,t}\big\|\mathbf{v}_\theta(\mathbf{x}_t,t,\mathbf{c})-(\boldsymbol\epsilon-\mathbf{x}_0)\big\|^2,
\label{eq:supp-flow-loss}
\end{align}
where $t\sim\mathcal{U}[0,1]$ and $\|\cdot\|^2$ denotes the mean squared error over texels and channels.
Texels of inactive regions and texels outside the UV charts carry zero targets.
We optimize with AdamW~\cite{loshchilov2017decoupled} (learning rate $10^{-4}$, weight decay $10^{-4}$, 10K-step linear warm-up, gradient norm clipping at 1.0, and bf16 mixed precision).
Each image-conditioned model is compiled with \texttt{torch.compile} and trained for 200K steps with batch size 72 on a single NVIDIA H200 GPU, replacing each conditioning signal independently with zeros with probability $0.1$.
At inference, we integrate the learned velocity field from noise at $t=1$ to data at $t=0$ with 50 uniform Euler steps, $\mathbf{x}_{t-1/50}=\mathbf{x}_t-\tfrac{1}{50}\mathbf{v}_\theta(\mathbf{x}_t,t,\mathbf{c})$, without classifier-free guidance.
\subsection{Same-Data Baselines}
\label{supp:models-samedata}
\begin{figure}[!tbp]
  \centering
  \includegraphics[width=\textwidth]{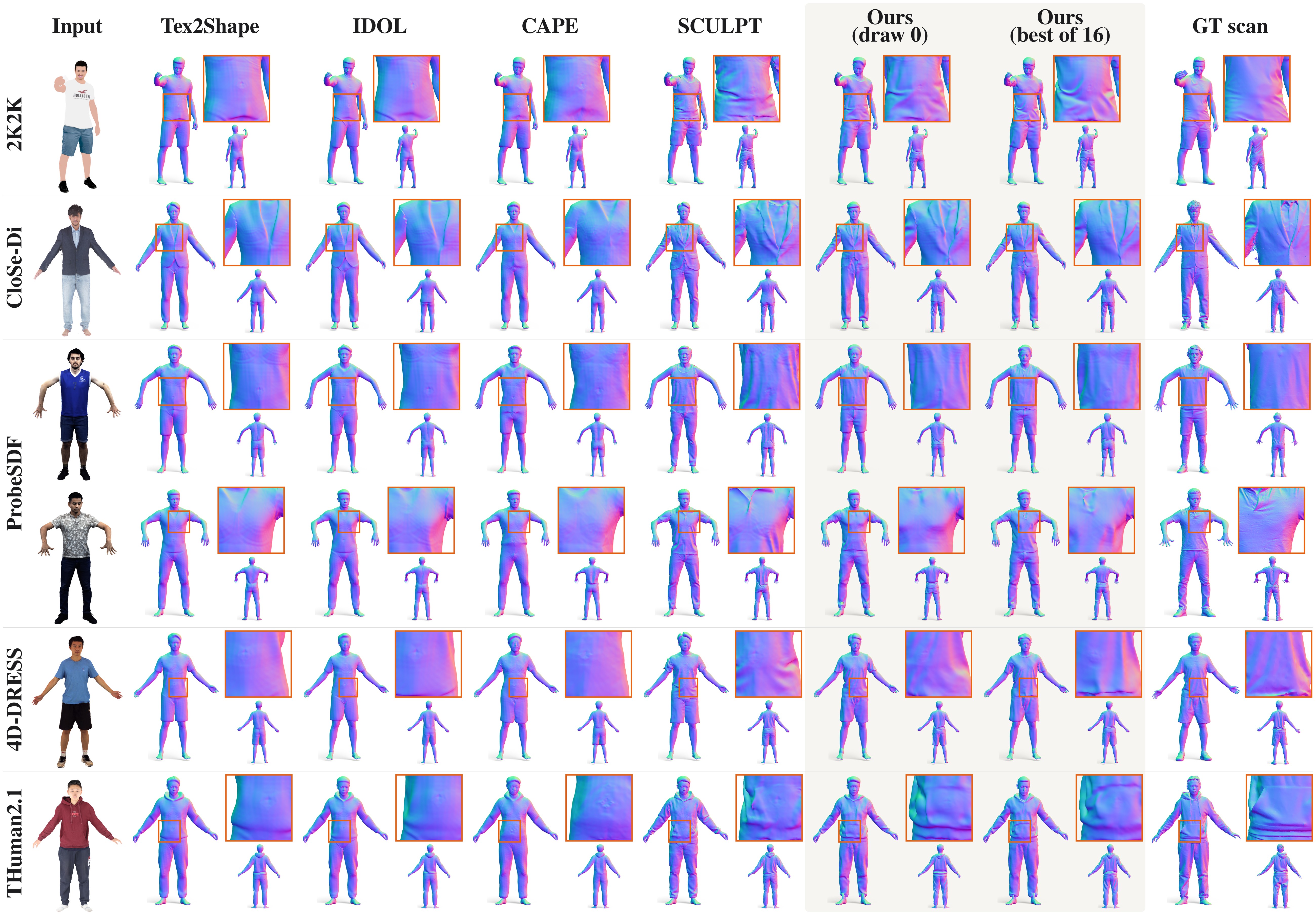}
  \caption{\textbf{Qualitative comparison of same-data baselines on real scans.} Surface normal renders of Tex2Shape, IDOL, CAPE, SCULPT, and our model (first draw and best of 16 by Chamfer distance), followed by the ground-truth scan; the stochastic CAPE and SCULPT are shown at their best of 16. Insets magnify the patch, indicated on each front view, where high-frequency surface detail varies most across methods.}
  \label{fig:supp-samedata}
\end{figure}

\begin{figure}[!tbp]
  \centering
  \includegraphics[width=\textwidth]{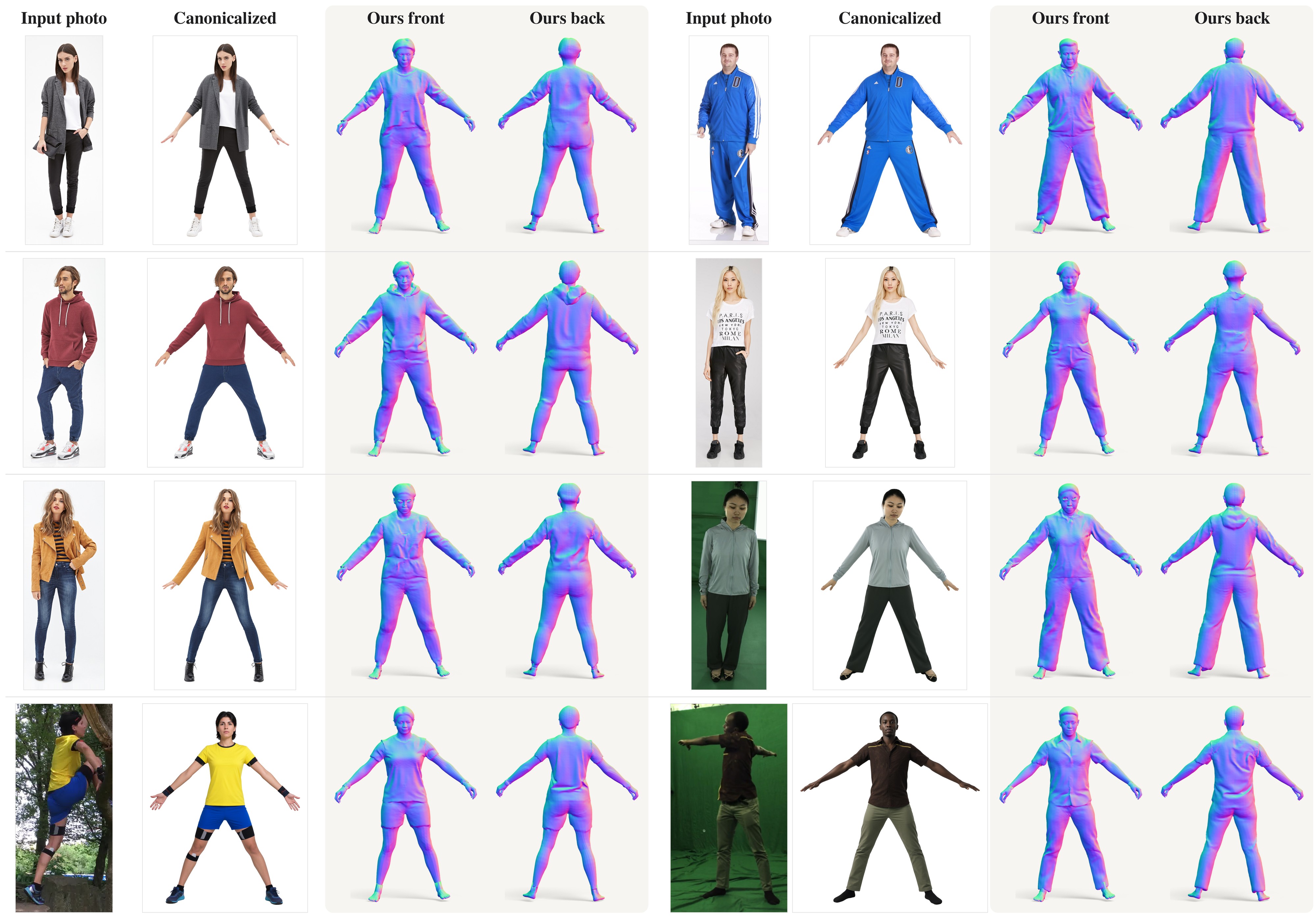}
  \caption{\textbf{Image-conditioned SMPL-X-D generation on in-the-wild photos.} Each example shows the input photo, its pose-canonicalized version obtained with Qwen-Image~2.1~\cite{wu2025qwenimagetechnicalreport} editing, and our SMPL-X-D decoded in the canonical A-pose (front and back normals); the edit may hallucinate content unseen in the photo. SMPL-X estimates come from our dense correspondence-guided fitting.}
  \label{fig:supp-wild}
\end{figure}

\begin{table}[!tbp]
  \centering
  \caption{\textbf{Same-data models.}
  All models use DINOv3 image tokens at $512\times512$, the SMPL-X shape parameters, Part~1 of \proteus, and 200K training steps on one NVIDIA H200 GPU.
  Huber thresholds $\delta$ are given in normalized offset units.}
  \label{tab:supp-samedata-config}
  \footnotesize
  \setlength{\tabcolsep}{2.5pt}
  \fitwidth{%
    \begin{tabular}{llllrr}
      \toprule
      \thead{Method} & \thead{Output} & \thead{Image / shape input} & \thead{Objective} & \thead{Batch} & \thead{LR} \\
      \midrule
      Tex2Shape & UV $256^2$ & projected grid / FiLM & Huber, $\delta{=}0.05$ & 24 & $10^{-4}$ \\
      IDOL & UV $256^2$ & cross-attention / FiLM & Huber, $\delta{=}0.05$ & 24 & $10^{-4}$ \\
      CAPE & mesh graph & cross-attention / global & Huber, $\delta{=}0.025$ + KL & 256 & $10^{-4}$ \\
      SCULPT & UV $256^2$ & global / global & hinge GAN & 144 & $2.5\cdot10^{-3}$ \\
      \textbf{Ours} & UV $256^2$ & joint attention / global & rectified flow & 72 & $10^{-4}$ \\
      \bottomrule
    \end{tabular}%
  }
\end{table}

To isolate architectural differences under identical supervision, we reimplement four representative template-aligned methods (\cref{tab:supp-samedata-config}).
All baselines share our training split, twice-subdivided SMPL-X template and UV layout, frozen DINOv3 image encoder, $\boldsymbol\beta$ shape conditioning, and 200K-step training budget, while retaining their native network topologies and training objectives (with Huber losses on normalized active-vertex offsets for all regression terms):
\begin{itemize}
    \item \textbf{Tex2Shape}~\cite{alldieck2019tex2shape}: Projects DINOv3 image tokens onto a $32\times32$ spatial grid processed by a 12-layer transformer, followed by a convolutional upsampling decoder that regresses the $256\times256$ UV displacement map.
    \item \textbf{IDOL}~\cite{zhuang2025idol}: Uses a $16\times16$ grid of learnable UV query tokens that cross-attend to DINOv3 image tokens across a 6-layer transformer, decoded into the $256\times256$ UV displacement map via the same convolutional upsampling head.
    \item \textbf{CAPE}~\cite{ma2020cape}: Implements a conditional graph-convolutional VAE directly on the subdivided SMPL-X mesh, using 256 global tokens that cross-attend to the DINOv3 image tokens and a 512-dimensional latent code, trained with a Huber reconstruction loss and a KL divergence weight of $10^{-6}$ (warmed up over 25K steps).
    \item \textbf{SCULPT}~\cite{sanyal2024sculpt}: Adapts a StyleGAN3-T~\cite{karras2021stylegan3} generator to synthesize the $256\times256$ UV displacement map conditioned on global image and shape embeddings, trained with a spectrally normalized hinge-loss discriminator.
\end{itemize}
\subsection{External Baselines}
\label{supp:models-external}
We evaluate seven external single-image clothed human reconstruction baselines, SiTH~\cite{ho2024sith}, SIFU~\cite{zhang2024sifu}, SHERT~\cite{zhan2024shert} (initialized from ECON~\cite{xiu2023econ}), VS~\cite{liu2024vs}, Human3Diffusion~\cite{xue2024human3diffusion}, HumanNOVA~\cite{hu2026humannova}, and SyncHuman~\cite{chen2025synchuman}, using their official repositories, pretrained checkpoints, and default inference settings.
All methods receive the exact same frontal RGB render and foreground mask for each of the 290 evaluation scans.
Body-guided methods additionally receive our Route~2 SMPL-X fit converted into each baseline's expected body parameterization and camera coordinate frame (verified to overlap at least $95\%$ of the foreground silhouette).
In total, we evaluate roughly 2K scan-method pairs.
\subsection{Evaluation Protocol}
\label{supp:models-eval}
\paragraph{Inputs and sampling.}
Our models take the same frontal render and Route~2 SMPL-X fit as the body-guided baselines.
For stochastic generators, we draw 16 samples per scan using fixed random seeds, using two notions of the best sample.
\Cref{tab:same-data-models} reports the 16-sample mean (\emph{Mean}) and the best sample selected independently per metric (\emph{Best}).
\Cref{tab:image-to-smplxd} and best-of-16 visualizations instead use the single sample with the lowest CD per scan.
\paragraph{3D geometric accuracy.}
Predictions and ground-truth scans are normalized to the scan's bounding box, and 100K surface points are sampled on each mesh to compute symmetric Chamfer distance (CD), F1 scores at thresholds of $0.005$ and $0.01$, and normal consistency (NC).
\paragraph{Distributional realism.}
We render $1024\times1024$ surface normal maps across 8 fixed views for all predictions and ground-truth scans, computing FID~\cite{heusel2017fid} and, in the same-data comparison, CMMD~\cite{jayasumana2024cmmd} in CLIP~\cite{radford2021clip} feature space.
\subsection{Additional Results}
\label{supp:models-results}
\paragraph{Qualitative results.}
\Cref{fig:supp-samedata} compares the same-data models on real scans: the deterministic regressors and CAPE produce smooth garments, SCULPT synthesizes sharp but less faithful wrinkles, and our samples recover folds, seams, and pockets closer to the scan.
\Cref{fig:supp-wild} applies our image-conditioned model to in-the-wild photos, whose poses we first canonicalize to the training A-pose with Qwen-Image~2.1~\cite{wu2025qwenimagetechnicalreport} image editing.

\FloatBarrier
\subsection{Feed-Forward Registration of Point Clouds}
\label{supp:models-pointcloud}
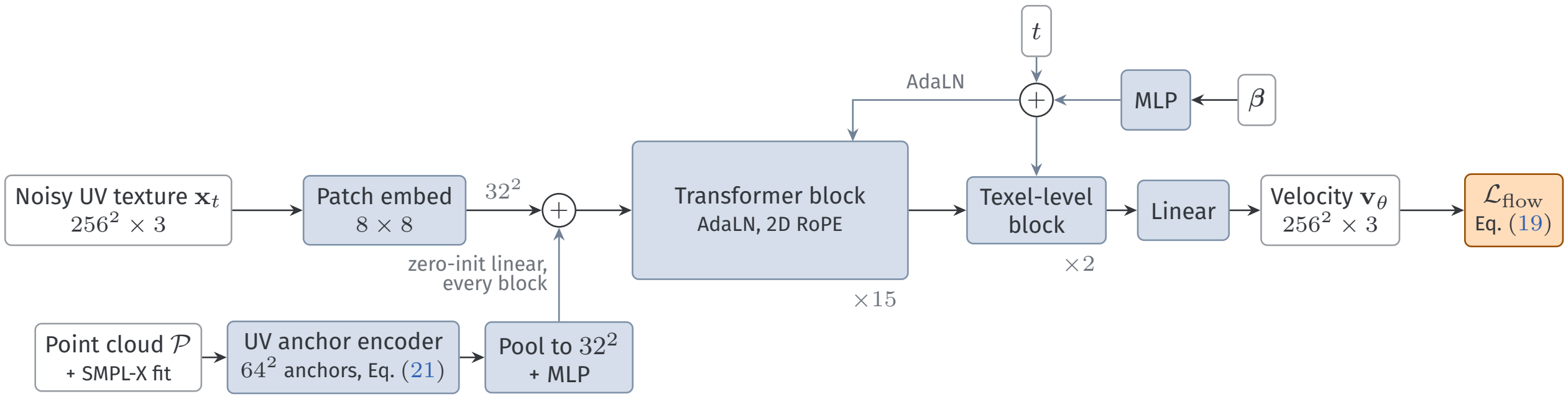
\begin{figure}[!tbp]
  \centering
  \fitwidth{\begin{tikzpicture}[arch]
    \node[data] (xt) at (1.2,0) {Noisy UV texture $\mathbf{x}_t$\\[-1pt]{\scriptsize $256^2\times3$}};
    \node[module] (patch) at (4.1,0) {Patch embed\\[-1pt]{\scriptsize $8\times8$}};
    \node[op] (add) at (6.0,0) {$+$};
    \node[module, minimum height=15mm, minimum width=30mm] (block) at (8.3,0) {Transformer block\\[-1pt]{\scriptsize AdaLN, 2D RoPE}};
    \node[dims, anchor=north east] at (block.south east) {$\times15$};
    \node[module] (texel) at (11.2,0) {Texel-level\\[-1pt]block};
    \node[dims, anchor=north east] at (texel.south east) {$\times2$};
    \node[module] (linear) at (12.8,0) {Linear};
    \node[data] (vel) at (14.4,0) {Velocity $\mathbf{v}_\theta$\\[-1pt]{\scriptsize $256^2\times3$}};
    \node[loss] (lflow) at (16.4,0) {$\mathcal{L}_{\mathrm{flow}}$\\[-1pt]{\scriptsize Eq.~\eqref{eq:supp-flow-loss}}};
    \draw[flow] (xt) -- (patch);
    \draw[flow] (patch) -- node[dims, above] {$32^2$} (add);
    \draw[flow] (add) -- (block);
    \draw[flow] (block) -- (texel);
    \draw[flow] (texel) -- (linear);
    \draw[flow] (linear) -- (vel);
    \draw[flow] (vel) -- (lflow);
    \node[op] (addc) at (11.2,1.2) {$+$};
    \node[data] (t) at (11.2,1.95) {$t$};
    \node[module] (mlp) at (12.5,1.2) {MLP};
    \node[data] (beta) at (13.6,1.2) {$\boldsymbol\beta$};
    \draw[cond] (t) -- (addc);
    \draw[flow] (beta) -- (mlp);
    \draw[cond] (mlp) -- (addc);
    \draw[cond] (addc) -| node[dims, pos=0.25, above] {AdaLN} (9.2,0.75);
    \draw[cond] (addc) -- (texel);
    \node[data] (pc) at (1.2,-1.6) {Point cloud $\mathcal{P}$\\[-1pt]{\scriptsize + SMPL-X fit}};
    \node[module] (anchor) at (3.65,-1.6) {UV anchor encoder\\[-1pt]{\scriptsize $64^2$ anchors, Eq.~\eqref{eq:supp-anchor-encoder}}};
    \node[module] (pool) at (6.0,-1.6) {Pool to $32^2$\\[-1pt]+ MLP};
    \draw[flow] (pc) -- (anchor);
    \draw[flow] (anchor) -- (pool);
    \draw[cond] (pool) -- node[dims, left, align=right] {zero-init linear,\\[-2pt]every block} (add);
  \end{tikzpicture}}
  \caption{\textbf{Architecture of the feed-forward point-cloud registration model.} The generator of \cref{fig:supp-generator-arch} without image tokens. A UV anchor encoder gathers the input points around a $64\times64$ grid of anchors on the posed SMPL-X surface, and the pooled anchor features are added to the patch tokens before every block through zero-initialized linear projections. The timestep and body shape $\boldsymbol\beta$ modulate every block as before.}
  \label{fig:supp-pointcloud-generator-arch}
\end{figure}

\begin{figure}[!tbp]
  \centering
  \includegraphics[width=\textwidth]{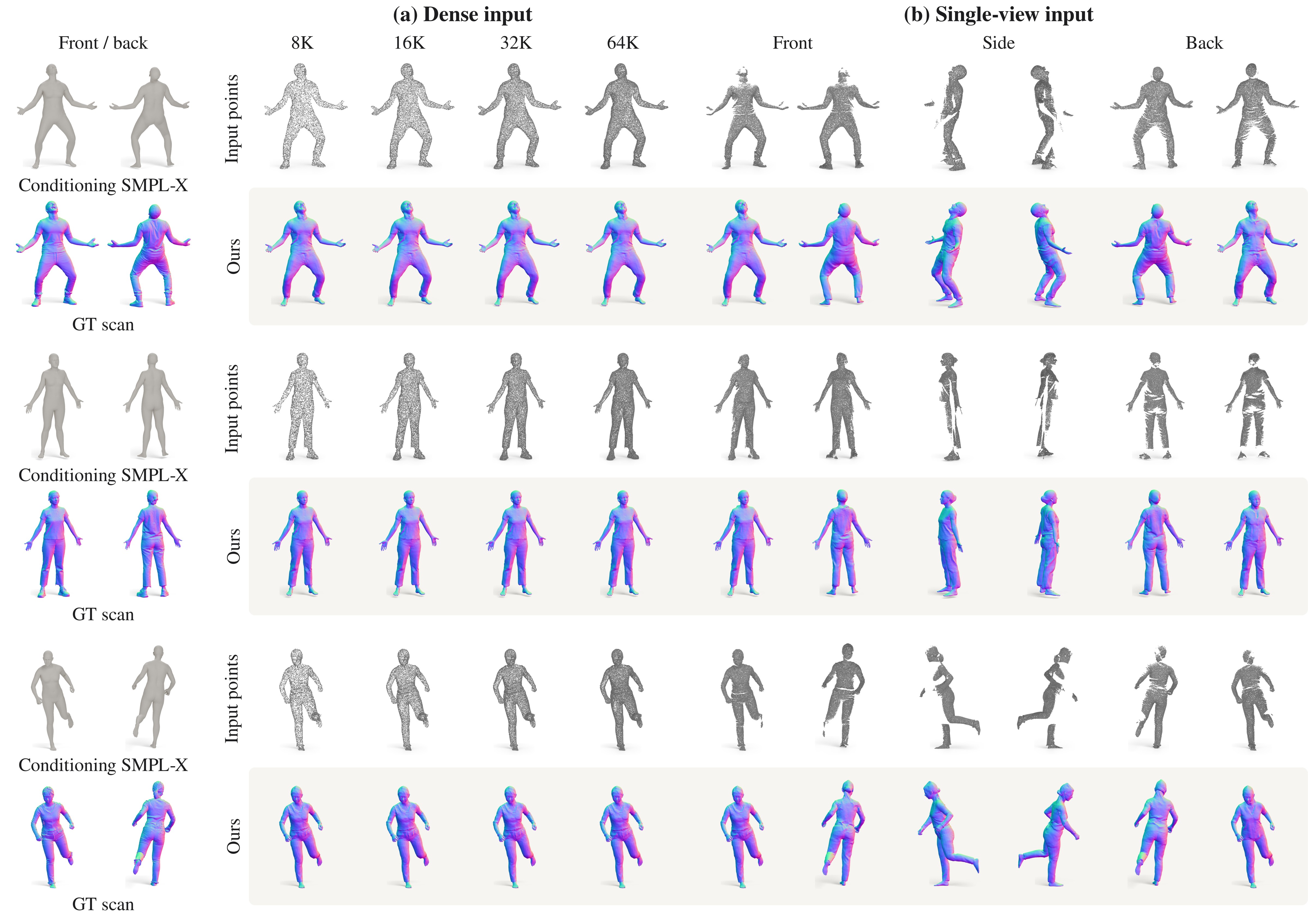}
  \caption{\textbf{Feed-forward registration across input densities and viewpoints.} For three real scans, (a)~dense point clouds of 8K to 64K points from 8 virtual cameras and (b)~single-view point clouds from the front, left side, and back, shown from the input camera and the opposite side. Each prediction (draw~0 of the variable-density model) appears below its input; the left column shows the conditioning Route~2 SMPL-X fit and the ground-truth scan.}
  \label{fig:supp-ffreg}
\end{figure}

\begin{figure}[!tbp]
  \centering
  \includegraphics[width=0.60\linewidth]{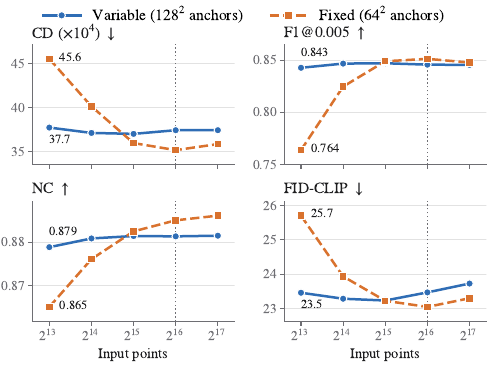}
  \caption{\textbf{Point-cloud conditioning across input point counts.} Evaluation across the 290 real scans with 4 samples per scan, where FID-CLIP is computed on normal renders. The variable-count model is trained on 8K to 64K points and the fixed-count model on 64K points (dotted line). Arrows indicate the direction of better performance.}
  \label{fig:supp-pointcloud-scaling}
\end{figure}

\paragraph{Point-cloud conditioning.}
For feed-forward point-cloud registration, we gather the input point cloud onto a $64\times64$ UV anchor grid on the posed SMPL-X surface.
Each anchor at posed position $\mathbf{a}$ retrieves its 16 nearest input points $\mathbf{p}_k$ with normals $\mathbf{n}_k$ and colors $\mathbf{c}_k$.
Their relative displacements and normals are mapped into canonical rest space via the inverse of the anchor's blended linear skinning transform $\mathbf{F}_{\mathbf{a}}$, making the local geometric encoding invariant to skeletal pose:
\begin{equation}
\boldsymbol\delta_k = \mathbf{F}_{\mathbf{a}}^{-1}(\mathbf{p}_k-\mathbf{a}),
\qquad
\tilde{\mathbf{n}}_k = \frac{\mathbf{F}_{\mathbf{a}}^{-1}\mathbf{n}_k}{\|\mathbf{F}_{\mathbf{a}}^{-1}\mathbf{n}_k\|_2},
\qquad
d_k = \|\boldsymbol\delta_k\|_2.
\end{equation}
A shared MLP $\phi$ embeds each neighbor, and a learned scalar score $\mathbf{w}^\top\mathrm{GELU}(\cdot)$ aggregates the 16 neighbors into a 128-channel anchor descriptor:
\begin{align}
\mathbf{e}_k &= \mathrm{LN}(\mathbf{W}_c\mathbf{c}_k) + \phi\big(\boldsymbol\delta_k/\ell,\,\tilde{\mathbf{n}}_k,\,d_k/\ell\big),\notag\\
\mathbf{f}_{\mathbf{a}} &= \sum_{k}\operatorname{softmax}_k\!\big(\mathbf{w}^\top\mathrm{GELU}(\mathbf{e}_k)\big)\,\mathbf{e}_k,
\label{eq:supp-anchor-encoder}
\end{align}
where $\ell=5$\,cm.
To distinguish observed from occluded surface regions, we append two coverage channels, $\boldsymbol\kappa_{\mathbf{a}} = \big[e^{-d_{(1)}/\ell},\ \tfrac{1}{16}\sum_k\mathbf{1}[d_k<5\,\mathrm{cm}]\big]$, encoding the nearest-neighbor distance $d_{(1)}$ and the fraction of neighbors within 5\,cm.
We then pool the resulting anchor features onto the $32\times32$ patch grid.
An MLP embeds the pooled $32\times32$ grid and adds it to the patch tokens via zero-initialized linear projections prior to each transformer block (\cref{fig:supp-pointcloud-generator-arch}).
\paragraph{Training and inference.}
Our feed-forward point-cloud registration model uses the PixelDiT architecture described in \cref{supp:models-ours}, conditioned on the SMPL-X shape embedding $\boldsymbol\beta$ and the $64\times64$ UV point-cloud anchor features without image tokens.
We train the model on 45K samples for 100K steps with batch size 40 on a single GPU using the same optimizer and learning-rate schedule as the image-conditioned generator.
For evaluation on the 290 real scans, we synthesize partial point clouds from 8 virtual camera views, condition on the Route~2 SMPL-X fit, and draw 4 samples per scan using 50 Euler integration steps.
\paragraph{Robustness to point-cloud density.}
\Cref{fig:supp-pointcloud-scaling} compares our fixed-density model (trained on 64K input points) against a variable-density variant trained with input point counts sampled uniformly between 8K and 64K points.
While the fixed-density model is marginally more accurate near its 64K training density, its F1@0.005 drops to $0.764$ when evaluated on sparse 8K-point inputs.
In contrast, the variable-density model maintains stable accuracy across a $16\times$ range of input densities, with F1@0.005 varying by at most $0.005$ between 8K and 128K points.
\paragraph{Qualitative results.}
\Cref{fig:supp-ffreg} shows that the variable-density model yields consistent registrations from 8K to 64K input points and completes the unobserved side of single-view point clouds.

\FloatBarrier
\subsection{Point-Cloud Segmentation}
\label{supp:models-segmentation}
\begin{table}[!tbp]
  \centering
  \caption{\textbf{Per-class IoU and F1 of point-cloud segmentation.}
  MGN only annotates upper and lower garments, so all other surfaces form one class (S+H+Sh).
  We highlight the \ffst{best} score per class.}
  \label{tab:supp-segmentation-groups}
  \footnotesize
  \setlength{\tabcolsep}{2.5pt}
  \fitwidth{%
    \begin{tabular}{lrrrrrrrrrrrrr}
      \toprule
      & \multicolumn{5}{c}{\thead{4D-DRESS}} & \multicolumn{5}{c}{\thead{CloSe-Di}} & \multicolumn{3}{c}{\thead{MGN}} \\
      \cmidrule(lr){2-6} \cmidrule(lr){7-11} \cmidrule(lr){12-14}
      \thead{Method} & \thead{Skin} & \thead{Hair} & \thead{Shoes} & \thead{Upper} & \thead{Lower} & \thead{Skin} & \thead{Hair} & \thead{Shoes} & \thead{Upper} & \thead{Lower} & \thead{S+H+Sh} & \thead{Upper} & \thead{Lower} \\
      \midrule
      \multicolumn{14}{l}{\emph{IoU}} \\
      CloSeNet & .691 & .551 & .807 & .683 & .609 & .801 & .905 & .887 & .954 & .884 & .567 & .928 & .677 \\
      CloSeNet (oracle) & .787 & .692 & .726 & .786 & .744 & \fst{.908} & \fst{.953} & \fst{.905} & \fst{.970} & \fst{.949} & .872 & .957 & .945 \\
      Sonata probe & .760 & .737 & .606 & .801 & .770 & .568 & .782 & .718 & .853 & .759 & .735 & .853 & .877 \\
      MV Sapiens2 & \fst{.958} & .776 & \fst{.890} & \fst{.971} & \fst{.954} & .803 & .922 & .836 & .962 & .933 & .890 & \fst{.967} & \fst{.954} \\
      \midrule
      \textbf{Ours} & .923 & \fst{.875} & .767 & .932 & .892 & .835 & .920 & .833 & .954 & .918 & \fst{.900} & .960 & .948 \\
      \midrule
      \multicolumn{14}{l}{\emph{F1}} \\
      CloSeNet & .818 & .711 & .893 & .811 & .757 & .889 & .950 & .940 & .977 & .939 & .724 & .963 & .807 \\
      CloSeNet (oracle) & .881 & .818 & .841 & .880 & .853 & \fst{.952} & \fst{.976} & \fst{.950} & \fst{.985} & \fst{.974} & .931 & .978 & .972 \\
      Sonata probe & .864 & .848 & .755 & .889 & .870 & .724 & .878 & .836 & .920 & .863 & .848 & .920 & .935 \\
      MV Sapiens2 & \fst{.978} & .874 & \fst{.942} & \fst{.985} & \fst{.976} & .891 & .959 & .911 & .980 & .965 & .942 & \fst{.983} & \fst{.977} \\
      \midrule
      \textbf{Ours} & .960 & \fst{.933} & .868 & .965 & .943 & .910 & .958 & .909 & .976 & .958 & \fst{.947} & .980 & .973 \\
      \bottomrule
    \end{tabular}%
  }
\end{table}

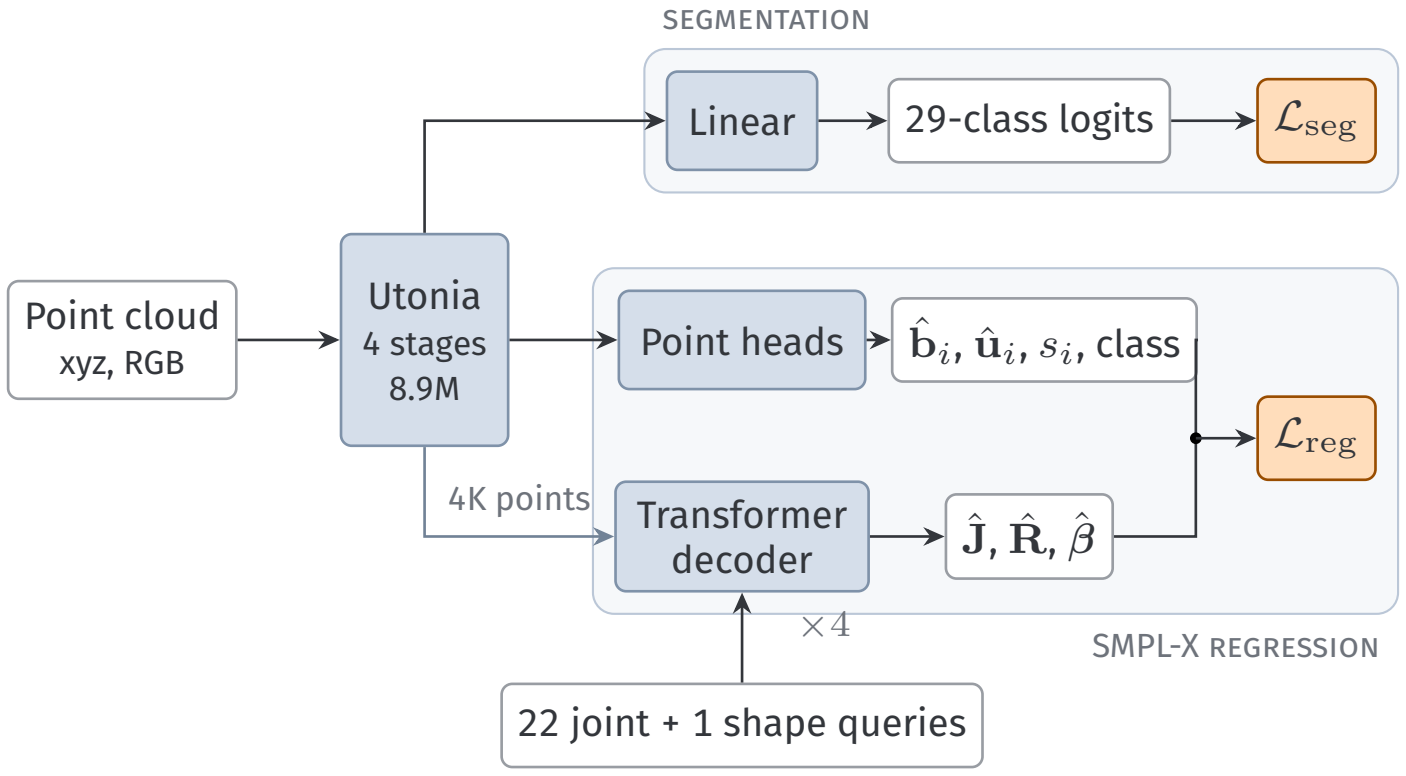
\begin{figure}[!tbp]
  \centering
  \fitwidth{%
  \begin{tikzpicture}[arch]
    \node[data] (pc) at (0.75,0) {Point cloud\\[-1pt]{\scriptsize xyz, RGB}};
    \node[module, minimum height=14mm] (utonia) at (2.75,0) {Utonia\\[-1pt]{\scriptsize 4 stages}\\[-1pt]{\scriptsize 8.9M}};
    \draw[flow] (pc) -- (utonia);
    \node[module] (seghead) at (4.85,1.45) {Linear};
    \node[data] (seglogits) at (6.75,1.45) {29-class logits};
    \node[loss] (lseg) at (8.65,1.45) {$\mathcal{L}_{\mathrm{seg}}$};
    \draw[flow] (utonia.north) |- (seghead);
    \draw[flow] (seghead) -- (seglogits);
    \draw[flow] (seglogits) -- (lseg);
    \node[module] (pointhead) at (4.85,0) {Point heads};
    \node[data] (pointout) at (6.85,0) {$\hat{\mathbf{b}}_i$, $\hat{\mathbf{u}}_i$, $s_i$, class};
    \node[module] (decoder) at (4.85,-1.3) {Transformer\\[-1pt]decoder};
    \node[dims, anchor=north east] at (decoder.south east) {$\times4$};
    \node[data] (queries) at (4.85,-2.55) {22 joint + 1 shape queries};
    \node[data] (jointout) at (6.75,-1.3) {$\hat{\mathbf{J}}$, $\hat{\mathbf{R}}$, $\hat{\boldsymbol\beta}$};
    \node[loss] (lreg) at (8.65,-0.65) {$\mathcal{L}_{\mathrm{reg}}$};
    \draw[flow] (utonia) -- (pointhead);
    \draw[flow] (pointhead) -- (pointout);
    \draw[flow] (queries) -- (decoder);
    \draw[cond] (utonia.south) |- node[dims, pos=0.75, above] {4K points} (decoder.west);
    \draw[flow] (decoder) -- (jointout);
    \coordinate (jreg) at ($(lreg.west)+(-0.4,0)$);
    \draw (pointout.east) -| (jreg) |- (jointout.east);
    \fill (jreg) circle (1.1pt);
    \draw[flow] (jreg) -- (lreg);
    \begin{scope}[on background layer]
      \node[group, fit=(seghead)(lseg)] (gseg) {};
      \node[group, fit=(pointhead)(decoder)(lreg)(jointout)] (greg) {};
    \end{scope}
    \node[stage, anchor=south west] at (gseg.north west) {segmentation};
    \node[stage, anchor=north east] at (greg.south east) {SMPL-X regression};
  \end{tikzpicture}%
  }
  \caption{\textbf{Point-cloud models.} A Utonia point transformer is trained from scratch for 29-class segmentation. Its checkpoint initializes the SMPL-X regressor, which adds per-point heads for the posed body point, canonical template coordinate with Laplace scale, and semantic class, and a transformer decoder whose joint and shape queries cross-attend to subsampled point features.}
  \label{fig:supp-pointcloud-arch}
\end{figure}

\begin{figure}[!tbp]
  \centering
  \includegraphics[width=\textwidth,height=0.82\textheight,keepaspectratio]{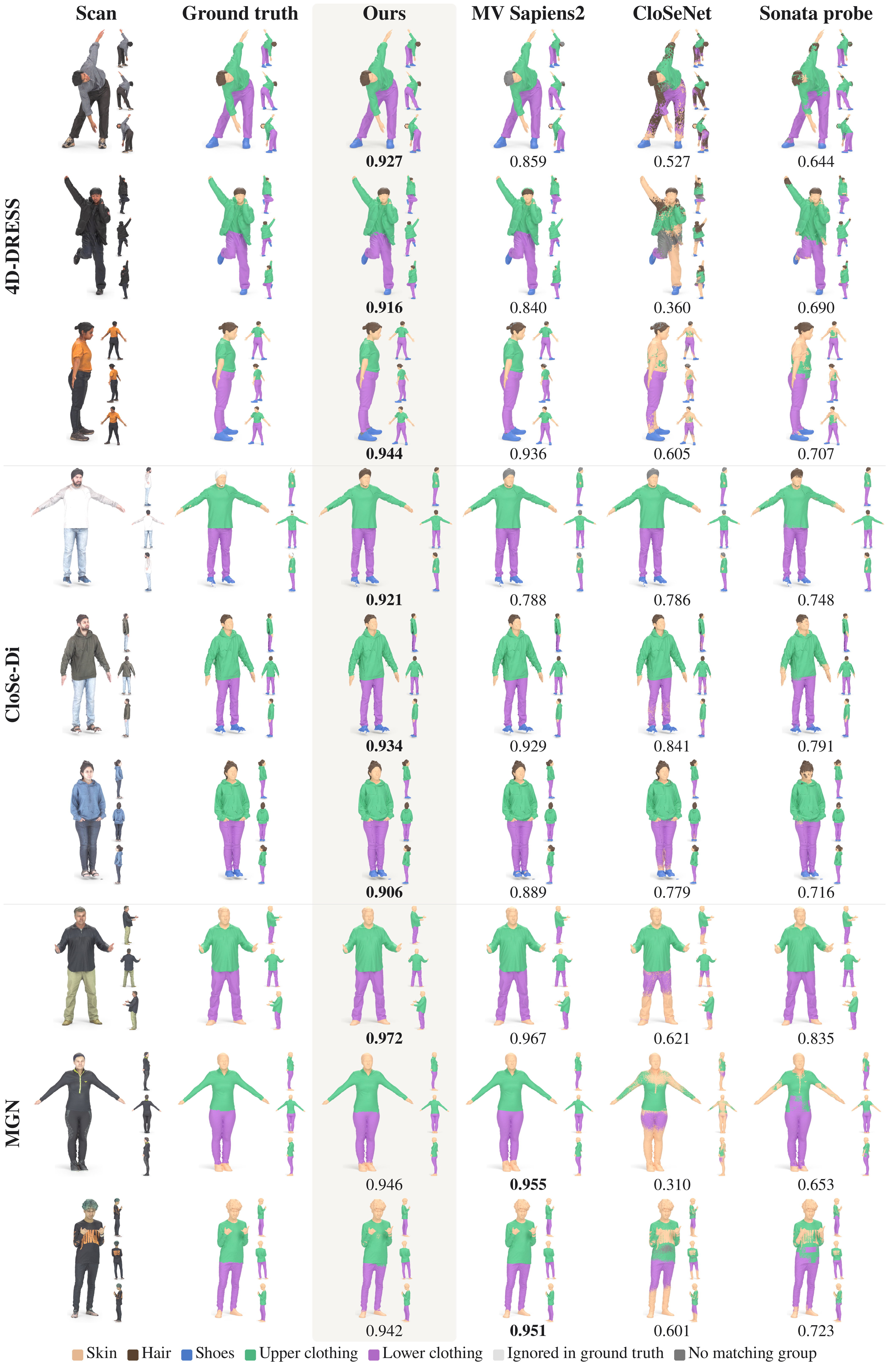}
  \caption{\textbf{Qualitative comparison of point-cloud segmentation on real scans.} Each method is shown from one primary and three auxiliary views with its scan-level grouped mIoU. Light gray indicates unannotated ground-truth surfaces and dark gray denotes predictions outside the evaluated groups. Skin, hair, and shoes are merged into a single class on MGN; CloSeNet (oracle) uses ground-truth garment labels.}
  \label{fig:supp-segmentation-comparison}
\end{figure}

\paragraph{Architecture and preprocessing.}
We train a Utonia~\cite{zhang2026utonia} point transformer from scratch followed by a linear classification head (\cref{fig:supp-pointcloud-arch}, top).
The network comprises 4 stages with 2 blocks per stage, channel widths increasing from 36 to 288, 3 to 24 attention heads, and 8.9M parameters in total.
Input point clouds are gravity-aligned ($+Z$ up) metric coordinates with RGB colors, centered at their centroid and scaled by the dataset-wide standard deviation of $0.2807$\,m (under which the 0.032 voxel size corresponds to $\sim 9$\,mm).
\paragraph{Training data and augmentations.}
Each training cloud contains 2K points sampled with equal probability from one of three sources:
\begin{itemize}
    \item the full outer surface $\mathcal{M}^{\mathrm{outer}}$ observed across all 32 virtual scanner cameras;
    \item a partial outer surface observed from 4 adjacent zero-elevation cameras;
    \item an AMASS-posed SMPL-X-D mesh observed across its 4 rendered camera views.
\end{itemize}
To simulate scanner artifacts and appearance variations, we apply the following augmentations:
\begin{itemize}
    \item normal-direction perturbations combining independent Gaussian noise and a smooth spatial wrinkle field (amplitudes up to 2.5\,mm on canonical outer clouds and 5\,mm on posed clouds, leaving 20\% of clouds unperturbed);
    \item up to 3 spherical cutouts of radius 5--15\,cm (applied with probability 0.5);
    \item random azimuth rotations around $+Z$ and uniform scaling in $[0.9, 1.1]$;
    \item photometric jitter, replacing RGB colors with neutral gray with probability 0.1.
\end{itemize}
\paragraph{Training and inference.}
We optimize a sum of class-weighted cross-entropy and the Lov\'asz-softmax loss~\cite{berman2018lovasz},
\begin{equation}
\mathcal{L}_{\mathrm{seg}}
=\frac{\sum_i w_{y_i}\big(-\log p_{i,y_i}\big)}{\sum_i w_{y_i}}
+\mathcal{L}_{\mathrm{Lov\acute{a}sz}},
\label{eq:supp-seg-loss}
\end{equation}
with class weights
\begin{equation}
w_c=\min\!\Big(\frac{f_c^{-1/2}}{\sum_{c'}f_{c'}^{1/2}},\,10\Big),
\end{equation}
where $p_{i,y_i}$ is the predicted probability of the ground-truth class $y_i$ of point $i$, $f_c$ is the training frequency of class $c$, and the normalization sets the frequency-weighted mean class weight to one before capping; unlabeled points are ignored.
We train for 100K steps with batch size 32 on a single GPU, using AdamW~\cite{loshchilov2017decoupled} (peak learning rate $2\times10^{-3}$, weight decay $0.01$, 2K-step linear warm-up, and cosine decay to $1\%$ of peak).
To segment arbitrary-resolution scans at test time while matching the 2K-point training density, we randomly partition the scan vertices into whole-body subsets of 2K points and average predicted logits across 3 independent partitions.
\paragraph{Evaluation protocol.}
We evaluate zero-shot on human-annotated scans from three benchmarks: 397 frames from 4D-DRESS~\cite{wang20244ddress}, 146 test scans from CloSe-Di~\cite{antic2024close}, and 58 scans from MGN~\cite{bhatnagar2019mgn}.
Fine-grained labels are mapped into five standard semantic groups (skin, hair, shoes, upper clothing, and lower clothing) as follows:
\begin{itemize}
    \item \textbf{4D-DRESS:} outer jackets and coats are grouped into upper clothing.
    \item \textbf{CloSe-Di:} the native body class maps to skin; dresses, swimwear, underwear, jumpsuits, and accessories are ignored in the ground truth (as \proteus excludes accessories, though any false-positive accessory predictions on evaluated regions are penalized).
    \item \textbf{MGN:} only upper and lower garments are annotated, so skin, hair, and shoes are merged into a single background/body class (\emph{S+H+Sh}) across the 58 fully annotated scans.
\end{itemize}
\Cref{tab:supp-segmentation-groups} reports per-class IoU and F1 scores.
\paragraph{Baselines.}
\emph{CloSeNet}~\cite{antic2024close} is evaluated using its official checkpoint both unconstrained (receiving all garment labels) and with an oracle list of the garment classes present in each scan; note that CloSeNet is trained on CloSe-D, which shares subject identities with CloSe-Di.
\emph{Multi-view Sapiens2}~\cite{khirodkar2026sapiens2} renders 50 views at $768\times768$ across elevations in $[-24^\circ, 24^\circ]$ and back-projects segmentation logits onto visible vertices using depth and view-angle weights, matching the annotation teacher used by \proteus.
Finally, the \emph{Sonata probe} trains a linear classifier for 20K steps on frozen Sonata~\cite{wu2025sonata} point-cloud features (using 2\,cm voxels, which outperformed 1\,cm voxels across all benchmarks) on the same \proteus training split.
\paragraph{Qualitative results.}
\Cref{fig:supp-segmentation-comparison} compares segmentation predictions across three scans per dataset: whereas CloSeNet and the Sonata linear probe frequently produce noisy, fragmented garment labels, our end-to-end point transformer yields spatially coherent boundaries matching the 50-view Sapiens2 teacher.

\FloatBarrier
\subsection{SMPL-X Regression from Point Clouds}
\label{supp:models-scan-smplx}
\begin{table}[!tbp]
  \centering
  \caption{\textbf{SMPL-X estimation from real scans.} Evaluation against Route~2 reference fits across 290 real scans (distances in mm). MPJPE and PVE are pelvis-aligned and PA-PVE is Procrustes-aligned. Surface metrics report the percentage of body vertices outside the scan, their mean and median protrusion depth, and bidirectional distances between exposed skin and the underlying body. Best full-scan estimates are \ffst{highlighted}.}
  \label{tab:supp-scan-smplx}
  \footnotesize
  \setlength{\tabcolsep}{4pt}
  \fitwidth{\begin{tabular}{llrrrrrrrr}
    \toprule
    & & \multicolumn{3}{c}{\thead{Pose error (mm)}} & \multicolumn{5}{c}{\thead{Surface}} \\
    \cmidrule(lr){3-5} \cmidrule(lr){6-10}
    \thead{Method} & \thead{Input} & \thead{MPJPE} & \thead{PVE} & \thead{PA-PVE} & \thead{Outside (\%)} & \thead{Protrusion} & \thead{Protrusion, median} & \thead{SMPL-X$\rightarrow$skin} & \thead{Skin$\rightarrow$SMPL-X} \\
    \midrule
    ETCH-X~\cite{li2026etchxrobustifyexpressivebody} & Full & 21.2 & 23.7 & 15.2 & 15.0 & \fst{5.3} & \fst{3.5} & \fst{8.5} & \fst{6.3} \\
    \textbf{Ours}, regressed & Full & 17.0 & 20.6 & \fst{11.5} & 13.6 & 6.3 & 4.2 & 13.0 & 8.0 \\
    \textbf{Ours}, refined & Full & \fst{15.9} & \fst{20.4} & 11.9 & 14.5 & 5.6 & 3.6 & 11.0 & 6.4 \\
    \midrule
    \textbf{Ours}, regressed & Single view & 21.5 & 24.0 & 15.6 & 18.2 & 10.9 & 5.9 & 13.9 & 9.9 \\
    \textbf{Ours}, refined & Single view & 21.3 & 24.2 & 16.2 & 20.3 & 9.7 & 5.3 & 11.9 & 8.2 \\
    \midrule
    Route~2 fit (reference) & -- & -- & -- & -- & 20.0 & 4.0 & 2.8 & 6.7 & 4.5 \\
    \bottomrule
  \end{tabular}}
\end{table}

\begin{figure}[!tbp]
  \centering
  \includegraphics[width=\textwidth,height=0.82\textheight,keepaspectratio]{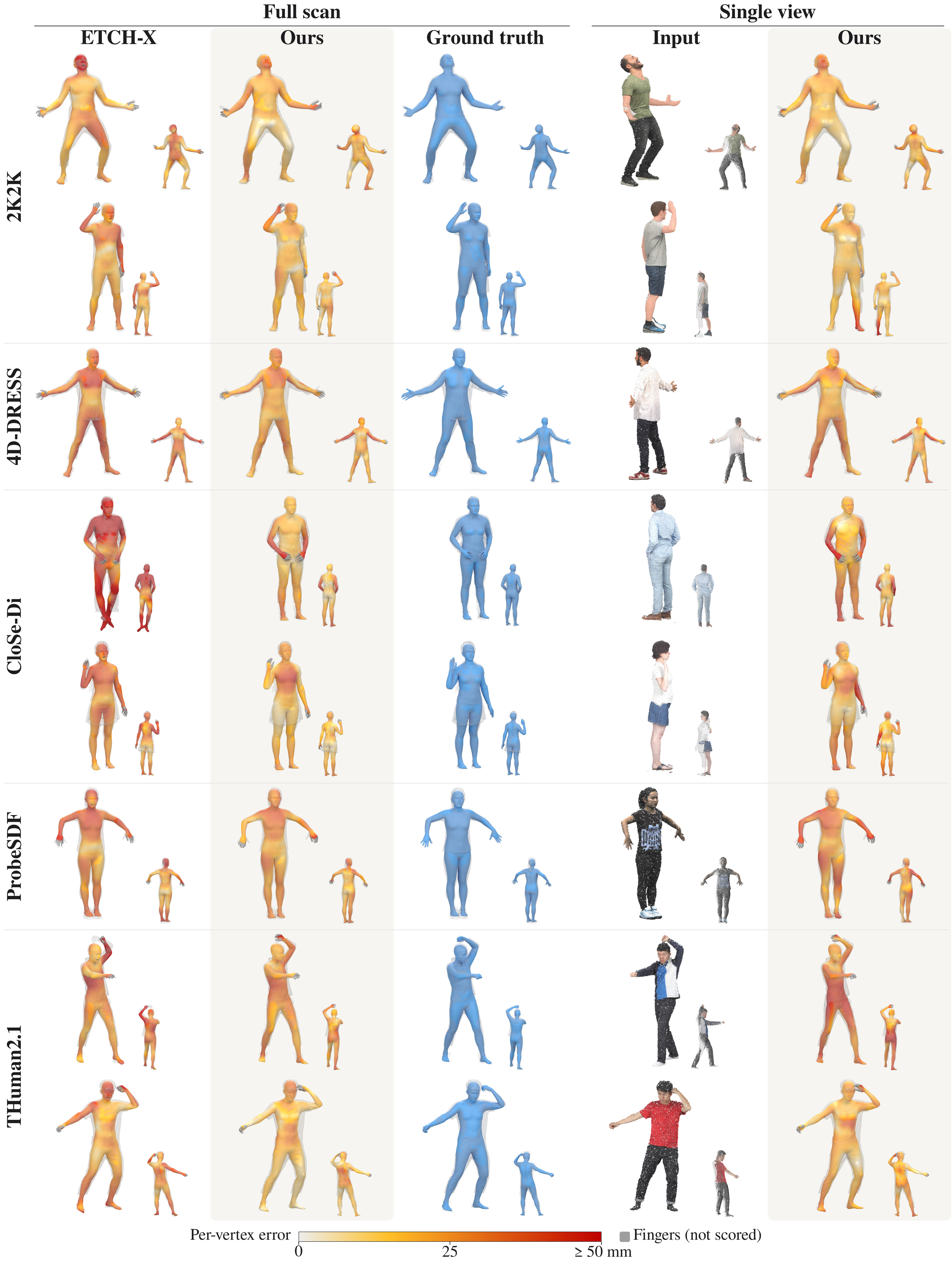}
  \caption{\textbf{Qualitative comparison of SMPL-X estimation from real scans.} Full-scan results compare ETCH-X~\cite{li2026etchxrobustifyexpressivebody} and our refined body against the Route~2 reference fit, while single-view results show the input point cloud alongside our refined body. Each panel displays a front three-quarter view with an opposite-side inset; bodies are rendered inside the semi-transparent scan and colored by pelvis-aligned per-vertex distance to the Route~2 fit.}
  \label{fig:supp-scan-smplx-comparison}
\end{figure}

\paragraph{Architecture.}
To regress SMPL-X parameters directly from a 16K-point cloud in the metric, gravity-aligned frame of \cref{supp:models-segmentation}, we attach dense per-point heads and a transformer decoder to the Utonia backbone (\cref{fig:supp-pointcloud-arch}, bottom).
For each input point $i$, pointwise heads predict: (1) the underlying posed unclothed body surface point $\hat{\mathbf{b}}_i$, (2) its canonical zero-shape template coordinate $\hat{\mathbf{u}}_i$ along with a Laplace log-scale $s_i$, and (3) its semantic class.
In parallel, a 4-layer transformer decoder (width 256, 8 heads) processes 22 learnable joint queries and 1 shape query that cross-attend to 4K subsampled point features augmented with 3D Fourier positional encodings.
Each joint query regresses its amodal 3D joint position and continuous 6D rotation~\cite{zhou2019continuity}, while the shape query predicts the 10 SMPL-X shape coefficients $\boldsymbol\beta$ relative to the \proteus mean (keeping jaw, eye, and finger rotations fixed to neutral).
\paragraph{Training data and objective.}
We use the point-cloud sources and augmentations of \cref{supp:models-segmentation} without scale augmentation, as metric point-cloud scale directly encodes body size; posed clouds are sampled from 1 to 4 camera views, and RGB colors are dropped with probability 0.5.
Let $\mathbf{V}$ and $\mathbf{J}$ denote the SMPL-X body vertices and joints posed with the predicted rotations $\hat{\mathbf{R}}_j$, shape $\hat{\boldsymbol\beta}$, and pelvis position, and $\hat{\mathbf{J}}^{\mathrm{dir}}$ the directly regressed joints.
The training objective combines Euclidean distances, a chordal rotation distance, and a Laplace negative log-likelihood on canonical template coordinates:
\begin{align}
\mathcal{L}_{\mathrm{reg}}
={}&10\big(\overline{\|\mathbf{V}-\mathbf{V}^{\mathrm{gt}}\|_2}
+\overline{\|\mathbf{J}-\mathbf{J}^{\mathrm{gt}}\|_2} + \overline{\|\hat{\mathbf{J}}^{\mathrm{dir}}-\mathbf{J}^{\mathrm{gt}}\|_2}
+\overline{\|\hat{\mathbf{b}}_i-\mathbf{b}_i\|_2}\big)
+\overline{\|\hat{\mathbf{R}}_j-\mathbf{R}_j\|_F^2}\notag\\
&+0.1\big(\overline{|\hat{\boldsymbol\beta}-\boldsymbol\beta|}+\mathcal{L}_{\mathbf{u}}+\mathcal{L}_{\mathrm{sem}}\big),
\label{eq:supp-scan-loss}\\
\mathcal{L}_{\mathbf{u}}
={}&\frac{1}{N}\sum_{i}\Big(\|\hat{\mathbf{u}}_i-\mathbf{u}_i\|_1\,e^{-s_i}+3\,s_i\Big),
\label{eq:supp-laplace-nll}
\end{align}
where overlines denote means over vertices, joints, points, or coefficients, $\mathcal{L}_{\mathrm{sem}}$ is the semantic cross-entropy, and $s_i$ is clamped below at $-7$.
We initialize the backbone from our point-cloud segmentation checkpoint and train for 100K steps (batch size 64) on an NVIDIA A40 GPU.
We use AdamW with head learning rate $3\times10^{-4}$, backbone learning rate $6\times10^{-5}$, weight decay $0.01$, gradient clipping at $1.0$, and a 1K-step warm-up; the point encoder runs in fp32 and the transformer decoder in bf16 for numerical stability.
\paragraph{Optional test-time refinement.}
At inference, the feed-forward prediction can optionally be refined with 10 Levenberg-Marquardt iterations over 4K subsampled points and the predicted 3D joints.
Each predicted template coordinate $\hat{\mathbf{u}}_i$ is snapped to the nearest template surface point, and $\mathbf{X}_i$ denotes that point on the posed body; we minimize
\begin{align}
E_{\mathrm{scan}}
={}&\sum_i \hat c_i\,\rho_{\sigma}\big(\|\mathbf{X}_i-\hat{\mathbf{b}}_i\|_2\big)
+\frac{1}{22}\sum_{j}\|\mathbf{J}_j-\hat{\mathbf{J}}^{\mathrm{dir}}_j\|_2^2\notag\\
&+\|\boldsymbol\beta-\boldsymbol\mu_{\boldsymbol\beta}\|^2_{\boldsymbol\Sigma_{\boldsymbol\beta}^{-1}}
+0.1\,\|\boldsymbol\theta-\boldsymbol\mu_{\boldsymbol\theta}\|^2_{\boldsymbol\Sigma_{\boldsymbol\theta}^{-1}},
\label{eq:supp-scan-fit}
\end{align}
with distances in centimeters, the Cauchy penalty $\rho_\sigma(x)=\sigma^2\log(1+x^2/\sigma^2)$ at scale $\sigma=3$\,cm, and confidences $\hat c_i\propto e^{-s_i}\,\mathbf{1}[\text{snap distance}<3\,\mathrm{cm}]$ normalized to sum to one.
During refinement, the Gaussian pose prior acts on the 21 relative joint rotations, and we keep wrist, ankle, and foot rotations fixed to their feed-forward predictions, as surface correspondences weakly constrain distal axial twist.
\paragraph{Evaluation protocol.}
We evaluate on the 290 real scans, sampling 16K colored points either uniformly across the full scan or from a single orthographic camera view at a random azimuth.
As a baseline, we evaluate the public 4D-DRESS checkpoint of ETCH-X~\cite{li2026etchxrobustifyexpressivebody} on full scans without hand or face parameters.
We report two sets of metrics:
\begin{itemize}
    \item \textbf{Pose accuracy:} Pelvis-aligned MPJPE and PVE, as well as Procrustes-aligned PA-PVE (in mm) over the 22 body joints and non-finger vertices, measured against our Route~2 SMPL-X fits.
    \item \textbf{Surface containment and skin proximity:} Using the generalized winding number on the full scan, we measure the percentage of non-finger body vertices protruding outside the scan along with their mean and median protrusion depth. On scan regions labeled as exposed skin (excluding hands), we also compute bidirectional distances between the SMPL-X body and the skin surface (\emph{SMPL-X$\rightarrow$skin} and \emph{Skin$\rightarrow$SMPL-X}). We include the Route~2 fit itself as a reference row.
\end{itemize}
\paragraph{Results.}
\Cref{tab:supp-scan-smplx} summarizes quantitative performance across full scans and single-view point clouds, and \cref{fig:supp-scan-smplx-comparison} shows qualitative examples.
On full scans, our feed-forward regressor outperforms ETCH-X across all pose metrics (achieving $20.6$\,mm vs.\ $23.7$\,mm PVE), while ETCH-X fits exposed skin more tightly and exhibits lower surface protrusion.
Adding 10 Levenberg-Marquardt refinement steps closes the skin-to-body gap ($6.4$\,mm vs.\ $6.3$\,mm \emph{Skin$\rightarrow$SMPL-X}) and further reduces MPJPE to $15.9$\,mm.
Even when restricted to a single partial view, our regressor achieves $24.0$\,mm PVE, matching full-scan ETCH-X accuracy.
On an NVIDIA RTX 3090, our feed-forward pass takes a median of $20$\,ms per scan ($57$\,ms with refinement), compared to $\sim 28$\,s for Stage~2 of ETCH-X alone (nearly $500\times$ faster).

\FloatBarrier
\subsection{Dense Correspondence and Monocular SMPL-X Fitting}
\label{supp:models-dense}
\begin{figure}[!tbp]
  \centering
  \includegraphics[width=\textwidth]{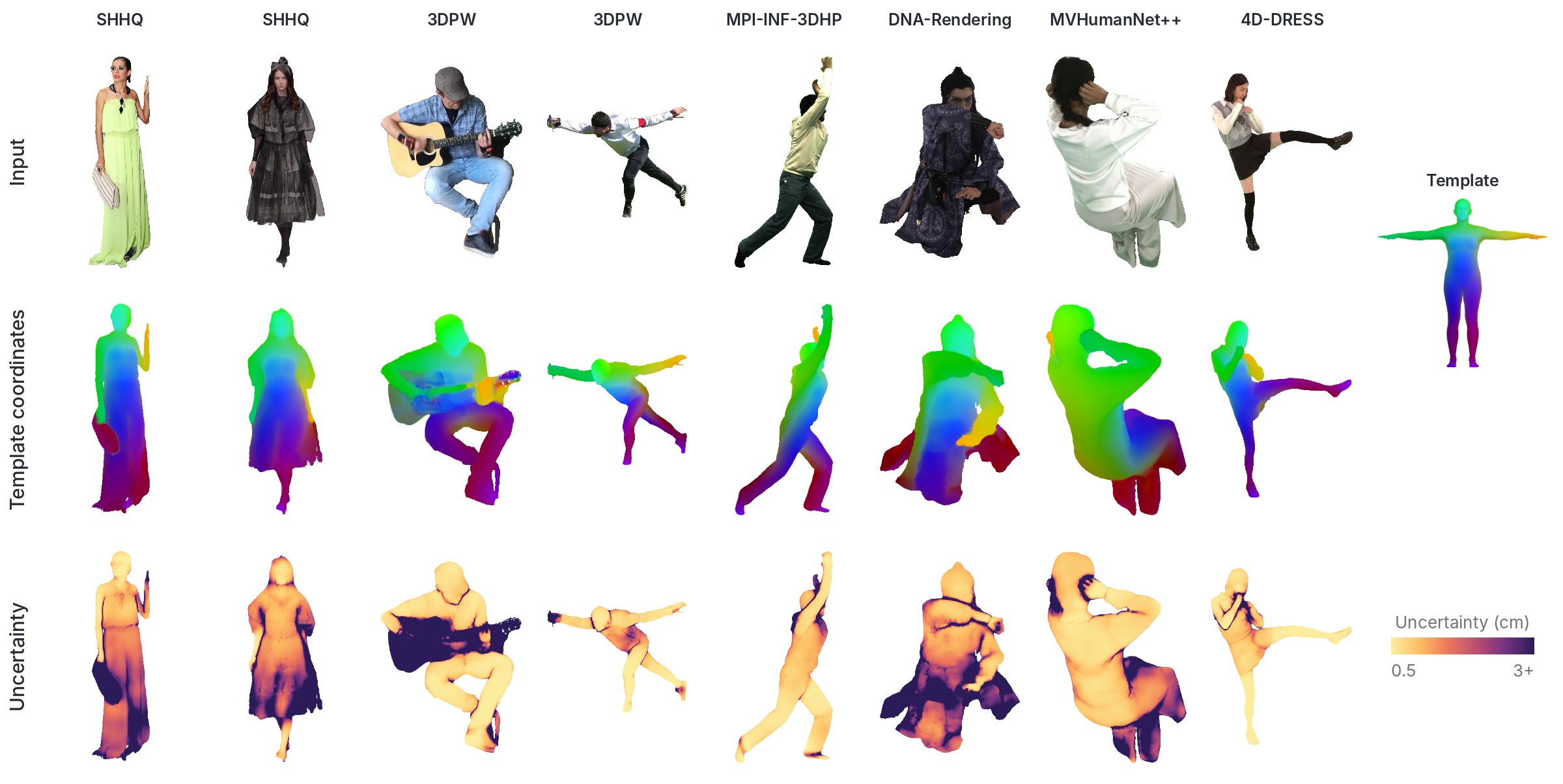}
  \caption{\textbf{Dense correspondence predictions on real images.} Predicted canonical template coordinates, colored as the T-pose template on the right, and their per-pixel Laplace uncertainty for in-the-wild photos and multi-view capture datasets. Uncertainty rises at garment hems, held objects, and self-occlusions.}
  \label{fig:supp-dense}
\end{figure}

\begin{figure}[!tbp]
  \centering
  \includegraphics[width=\textwidth]{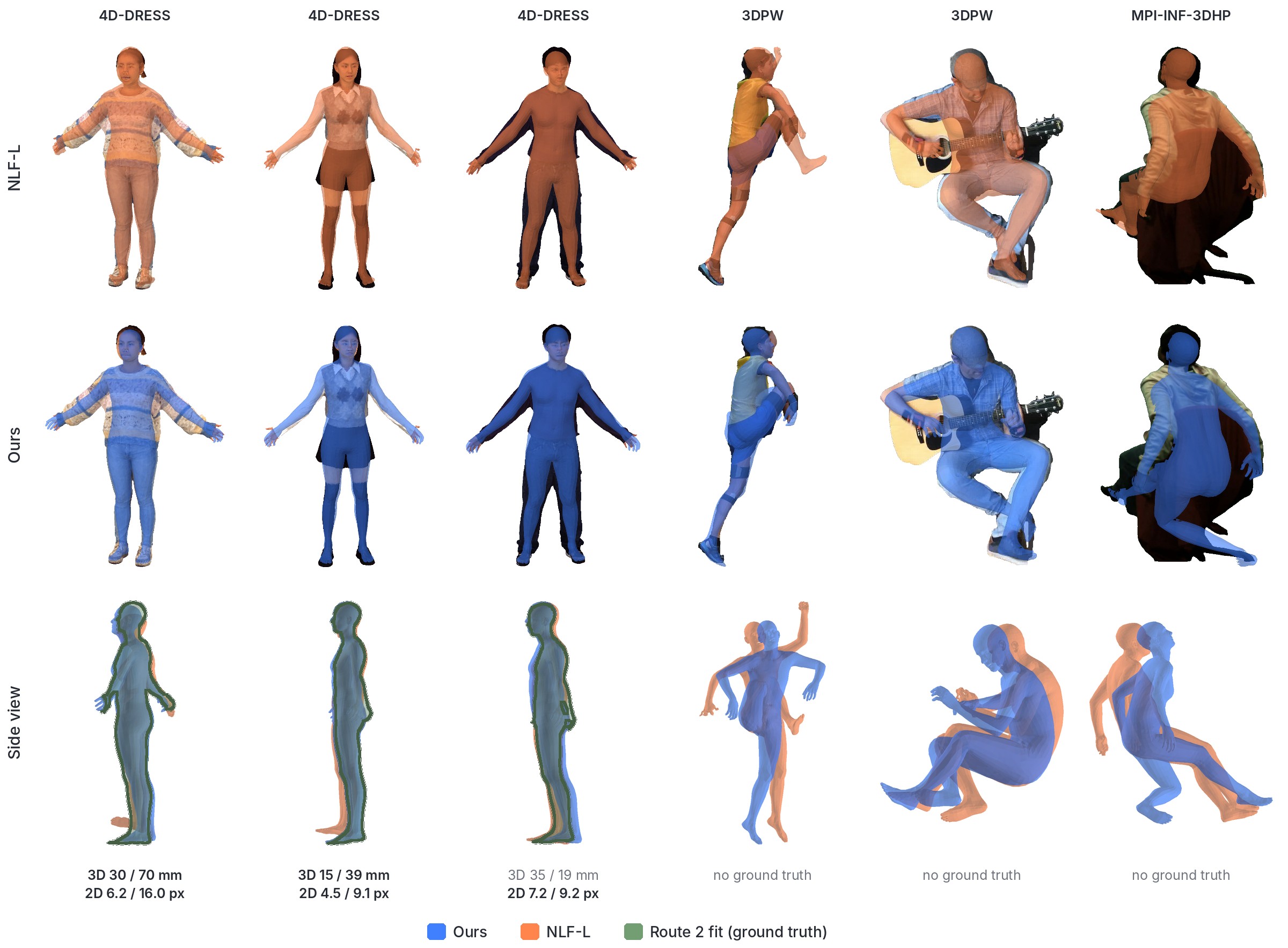}
  \caption{\textbf{Qualitative comparison of monocular SMPL-X fitting.} Image overlays of NLF-L~\cite{sarandi2024nlf} and our fit, with a pelvis-aligned side view of both bodies and the Route~2 fit (green) where available. Numbers give MPJPE and 2D joint error (ours\,/\,NLF-L) on 4D-DRESS renders; the third column and the occluded seated pose in the last column are failure cases.}
  \label{fig:supp-nlf}
\end{figure}

\begin{table}[!tbp]
  \centering
  \caption{\textbf{Monocular SMPL-X fitting on real scans.} Errors (mm) against Route~2 fits on frontal renders of the 290 real scans with known intrinsics. MPJPE and PVE are pelvis-aligned, PA denotes Procrustes alignment, and 2D reports joint reprojection error in pixels. Ground-truth joints (2D projections and pelvis-relative depths) provide an oracle upper bound for the joint term.}
  \label{tab:supp-dense-smplx}
  \footnotesize
  \setlength{\tabcolsep}{3pt}
  \fitwidth{%
    \begin{tabular}{lrrrrr}
      \toprule
      \thead{Method} & \thead{MPJPE} & \thead{PA-MPJPE} & \thead{PVE} & \thead{PA-PVE} & \thead{2D} \\
      \midrule
      NLF-L~\cite{sarandi2024nlf} & \fst{37.1} & \fst{23.9} & \fst{44.2} & \fst{22.8} & 11.1 \\
      \midrule
      Ours, correspondences & 66.4 & 38.2 & 75.1 & 31.6 & 10.3 \\
      Ours, + predicted joints & 47.1 & 31.0 & 52.5 & 27.1 & \fst{7.5} \\
      \midrule
      Ours, + ground-truth joints & 11.2 & 6.2 & 18.1 & 12.0 & 3.1 \\
      \bottomrule
    \end{tabular}%
  }
\end{table}

\begin{figure}[!tbp]
  \centering
  \fitwidth{\begin{tikzpicture}[arch]
    \node[data] (img) at (0.6,0) {Image\\[-1pt]{\scriptsize RGB}};
    \node[module, minimum height=14mm] (vit) at (2.35,0) {DINOv3\\[-1pt]ViT-L/16};
    \node[dims, anchor=south] at (vit.north) {blocks 6, 12, 18, 24};
    \node[module, minimum height=14mm] (fpn) at (4.6,0) {Feature\\[-1pt]pyramid\\[-1pt]{\scriptsize ConvNeXt}};
    \node[module] (stem) at (2.35,-1.15) {RGB stem};
    \node[module] (densehead) at (7.1,0.45) {Dense heads};
    \node[module] (jointhead) at (7.1,-0.45) {Joint head};
    \draw[flow] (img) -- (vit);
    \draw[flow] (vit) -- node[dims, above] {$\times4$} (fpn);
    \draw[flow] (img) |- (stem);
    \draw[flow] (stem) -| node[dims, pos=0.3, below] {strides 4, 2, 1} (fpn);
    \draw[flow] (fpn.east |- densehead) -- node[dims, above] {stride 1} (densehead);
    \draw[flow] (fpn.east |- jointhead) -- node[dims, below] {stride 4} (jointhead);
    \node[data] (denseout) at (9.5,0.45) {$\hat{\mathbf{u}}$, $s$, mask, depth};
    \node[data] (jointout) at (9.45,-0.45) {soft-argmax joints};
    \draw[flow] (densehead) -- (denseout);
    \draw[flow] (jointhead) -- (jointout);
    \node[loss] (ldense) at (11.65,0) {$\mathcal{L}_{\mathrm{dense}}$};
    \coordinate (jdense) at ($(ldense.west)+(-0.4,0)$);
    \draw (denseout.east) -| (jdense) |- (jointout.east);
    \fill (jdense) circle (1.1pt);
    \draw[flow] (jdense) -- (ldense);
    \node[module] (init) at (2.35,-2.55) {Closed-form init\\[-1pt]{\scriptsize Kabsch + ridge, 6 it.}};
    \node[module] (lm) at (5.6,-2.55) {Clothed LM\\[-1pt]{\scriptsize 10 it., Eq.~\eqref{eq:supp-dense-fit}}};
    \node[data] (smplx) at (8.6,-2.55) {SMPL-X\\[-1pt]{\scriptsize $\boldsymbol\theta$, $\boldsymbol\beta$, translation}};
    \draw[flow] (init) -- (lm);
    \draw[flow] (lm) -- (smplx);
    \draw[cond] (jointout.south) -- (jointout.south |- 0,-1.75) -| node[dims, pos=0.2, above] {weighted correspondences, joints} (init.north);
    \draw[cond] (jointout.south |- 0,-1.75) -| (lm.north);
    \begin{scope}[on background layer]
      \node[group, fit=(init)(lm)(smplx)] (gfit) {};
    \end{scope}
    \node[stage, anchor=north east] at (gfit.south east) {monocular SMPL-X fitting};
  \end{tikzpicture}}
  \caption{\textbf{Dense correspondence network and monocular SMPL-X fitting.} A fine-tuned DINOv3 backbone and convolutional feature pyramid predict per-pixel canonical template coordinates $\hat{\mathbf{u}}$ with Laplace scales $s$, a foreground mask, relative depth, and 3D joints. The weighted correspondences drive a closed-form initialization followed by clothed Levenberg-Marquardt refinement.}
  \label{fig:supp-dense-arch}
\end{figure}
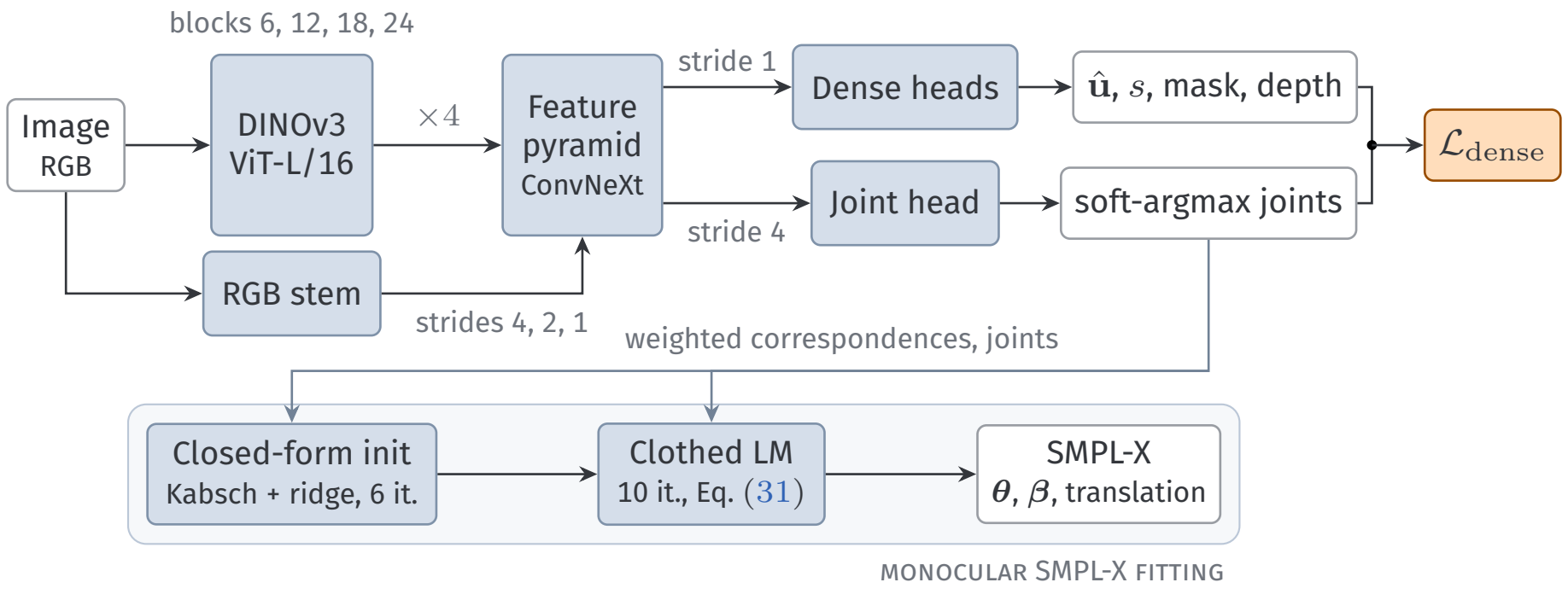

\paragraph{Architecture.}
Our dense image correspondence predictor fine-tunes a DINOv3 ViT-L/16~\cite{simeoni2026dinov3} backbone paired with a multi-scale convolutional feature pyramid (\cref{fig:supp-dense-arch}).
Tokens from transformer blocks 6, 12, 18, and 24 feed pyramid levels at strides 16, 8, 4, and 2, while a 3-layer RGB convolutional stem injects high-resolution features at strides 4, 2, and 1 to preserve sharp silhouette edges.
Feature channels are progressively halved from 256 down to 16 across pyramid levels and refined via ConvNeXt blocks.

At full resolution, the network predicts per-pixel canonical SMPL-X template coordinates, a foreground silhouette logit, a Laplace log-uncertainty scale, and person-relative depth.
In addition, a lightweight joint head attached to the stride-4 feature map predicts 2D heatmaps, pelvis-relative depths, and uncertainty scales for the 22 SMPL-X body joints, read out via spatial soft-argmax~\cite{sun2018integral}.
\paragraph{Training data and augmentations.}
We train progressively across three image resolutions: the $256\times256$ stage uses the 283K posed samples from Part~2 (\cref{supp:models-data}), while the $512\times512$ and $1024\times1024$ stages add the 99K canonical A-pose renders from Part~1 (382K samples total) to preserve coverage of loose garments.
To simulate imperfect foreground segmentation at test time, we composite training renders onto black backgrounds using alpha mattes perturbed by random morphological erosion/dilation (up to 4\,px at $512\times512$), contour noise, and random binarization (probability 0.5).
We sample square bounding-box crops scaled between $1.0\times$ and $1.4\times$ the body extent (with up to $10\%$ translation shift), and with probability 0.15 sample truncated partial-body crops ($0.4\text{--}0.8\times$ extent).
We further apply photometric jitter, Gaussian blur (0.2), sensor noise (0.2), JPEG compression (0.3), up to 3 rectangular erasing occluders (0.2), and random horizontal flipping (0.5, with corresponding left-right target reflection).
\paragraph{Training schedule.}
The training loss combines per-pixel, mask, and joint terms:
\begin{equation}
\mathcal{L}_{\mathrm{dense}}
=\mathcal{L}_{\mathbf{u}}
+0.5\,\mathcal{L}_{\nabla}
+\mathcal{L}_{\mathrm{BCE}}+\mathcal{L}_{\mathrm{Dice}}
+\mathcal{L}_{z}
+\mathcal{L}_{\mathrm{joint}}.
\label{eq:supp-dense-loss}
\end{equation}
Here, $\mathcal{L}_{\mathbf{u}}$ is the Laplace negative log-likelihood of \cref{eq:supp-laplace-nll} on per-pixel canonical template coordinates over foreground pixels $p$, and $\mathcal{L}_{\mathrm{BCE}}$ and $\mathcal{L}_{\mathrm{Dice}}$ supervise the foreground mask.
A spatial finite-difference loss preserves sharp correspondence boundaries,
\begin{equation}
\mathcal{L}_{\nabla}
=\frac{1}{2|S|}\sum_{\delta\in S}\ \sum_{\mathrm{axis}\in\{x,y\}}
\overline{\big\|\nabla_{\delta}\hat{\mathbf{u}}_p-\nabla_{\delta}\mathbf{u}_p\big\|_1},
\end{equation}
where $\nabla_\delta$ is the finite difference at stride $\delta$ along one image axis, averaged over pixel pairs inside the foreground, with $S=\{1\}$ at $256\times256$, $\{1,2,4\}$ at $512\times512$, and $\{1,2,4,8\}$ at $1024\times1024$.
Since person-relative depth is defined up to an offset, the shift-invariant depth loss over the foreground pixel set $\mathcal{F}$,
\begin{equation}
\mathcal{L}_z=\frac{1}{|\mathcal{F}|}\sum_{p\in\mathcal{F}}\big|(\hat z_p-z_p)-\mu\big|,
\end{equation}
subtracts the foreground mean $\mu$ of $\hat z_p-z_p$.
The joint term is a Laplace negative log-likelihood on 2D coordinates $\hat{\mathbf{j}}_j$ and pelvis-relative depths $\hat z_j$ with separate log-scales $s^{\mathrm{2D}}_j$ and $s^{z}_j$,
\begin{align}
\mathcal{L}_{\mathrm{joint}}
=\frac{1}{|\mathcal{J}|}\sum_{j\in\mathcal{J}}\Big(
&\|\hat{\mathbf{j}}_j-\mathbf{j}_j\|_1\,e^{-s^{\mathrm{2D}}_j}+2\,s^{\mathrm{2D}}_j + |\hat z_j-z_j|\,e^{-s^{z}_j}+s^{z}_j\Big),
\end{align}
where $\mathcal{J}$ is the set of joints inside the crop.
We optimize with AdamW (weight decay $0.05$, gradient clipping at 1.0, bf16 mixed precision, backbone learning rate $0.1\times$ the head learning rate with layer-wise decay $0.8$, and EMA weight decay $0.999$) across three progressive stages, scaling the head learning rate with the square root of the effective batch size (base $5\times10^{-4}$ at batch size 32):
\begin{itemize}
    \item $256\times256$: batch size 104, head learning rate $9.01\times10^{-4}$, 45K steps;
    \item $512\times512$: batch size 24, head learning rate $6.12\times10^{-4}$, 75K steps;
    \item $1024\times1024$: batch size 6, head learning rate $4.33\times10^{-4}$, 100K steps.
\end{itemize}
\paragraph{Monocular SMPL-X fitting.}
At test time, every foreground pixel at stride 4 provides a 2D-to-3D correspondence whose predicted canonical coordinate is snapped to the nearest vertex on the twice-subdivided SMPL-X template (within 3\,cm) and weighted by its predicted inverse Laplace scale.
We estimate the 22 SMPL-X body joint rotations, 10 shape coefficients $\boldsymbol\beta$, and 3D camera translation in two stages:
\begin{enumerate}
    \item \textbf{Closed-form initialization:} Following the alternating solver of NLF~\cite{sarandi2024nlf} for 6 iterations, a weighted Kabsch step aligns each body segment rotation to the camera rays of its assigned pixels (regularized toward its parent joint), followed by a ridge-regression update of body shape and global translation.
    \item \textbf{Clothed Levenberg-Marquardt refinement:} We run 10 Levenberg-Marquardt iterations with analytic Jacobians on a robust reprojection and depth objective with Gaussian priors on pose (fit to AMASS) and shape (fit to \proteus), defined below.
\end{enumerate}
Clothed pixels lie outside the minimally clothed body, so each pixel correspondence $i$ with snapped body point $\mathbf{X}_i$, outward normal $\mathbf{n}_i$, and observed pixel $\mathbf{x}_i$ may be explained by a non-negative outward offset $o_i\in[0,15\,\mathrm{cm}]$:
\begin{align}
E_{\mathrm{mono}}
={}&\sum_i \hat c_i\left[\rho_\sigma\!\left(\left\|\begin{bmatrix}\pi(\mathbf{X}_i+o_i\mathbf{n}_i)-\mathbf{x}_i\\ \lambda_z\big((\mathbf{X}_i)_z-a\,\tilde z_i-b\big)\end{bmatrix}\right\|_2\right)+\kappa\,o_i^2\right]\notag\\
&+\sum_j \omega_j\,\rho_\sigma\big(\|\mathbf{r}^{J}_j\|_2\big) + \|\boldsymbol\beta-\boldsymbol\mu_{\boldsymbol\beta}\|^2_{\boldsymbol\Sigma_{\boldsymbol\beta}^{-1}}
+0.1\,\|\boldsymbol\theta-\boldsymbol\mu_{\boldsymbol\theta}\|^2_{\boldsymbol\Sigma_{\boldsymbol\theta}^{-1}},
\label{eq:supp-dense-fit}
\end{align}
where $\pi$ projects to crop pixels, $\rho_\sigma$ is the Cauchy penalty of \cref{eq:supp-scan-fit} at scale $\sigma=8$\,px, and $\kappa=(5\,\mathrm{cm})^{-2}$ imposes a mild penalty on the offset.
In the depth residual, $(\mathbf{X}_i)_z$ is the camera-frame depth of the body point and $\tilde z_i$ is the mean-centered monocular depth at pixel $i$, aligned via a jointly optimized per-image scale and shift $(a,b)$ and scaled into pixel units by $\lambda_z$.
The depth map can be predicted by our network's depth head or supplied by an off-the-shelf monocular depth estimator such as MoGe-2~\cite{wang2025moge2}.
We linearize the projection in $o_i$ and solve for each offset in closed form before clamping it to $[0,15\,\mathrm{cm}]$.
Confidences $\hat c_i\propto e^{-s_i}$ follow the predicted Laplace scales and sum to one.
The joint residual $\mathbf{r}^{J}_j$ stacks the reprojection error of joint $j$ and its relative-depth error, and the weights $\omega_j\propto e^{-s^{\mathrm{2D}}_j}$ over joints inside the crop sum to one.
\paragraph{Evaluation and results.}
We evaluate monocular SMPL-X fitting on frontal renders of the 290 real scans (with known camera intrinsics) against NLF-L~\cite{sarandi2024nlf} (run with 5 test-time augmentations), using our Route~2 SMPL-X fits as ground truth.

\Cref{tab:supp-dense-smplx} summarizes the results.
Incorporating our joint head alongside dense surface correspondences reduces our pelvis-aligned MPJPE from $66.4$\,mm to $47.1$\,mm and achieves lower 2D reprojection error than NLF-L ($7.5$\,px vs.\ $11.1$\,px).
NLF-L achieves lower 3D error across the full five-dataset benchmark, whereas our method outperforms it on 4D-DRESS ($31.8$\,mm vs.\ $36.2$\,mm MPJPE).
Substituting ground-truth 2D joint projections and pelvis-relative joint depths reduces MPJPE to $11.2$\,mm, indicating that 3D error is primarily limited by the joint head rather than the dense correspondences.
Our two-stage solver runs in a median of $146$\,ms per image on an NVIDIA RTX 3090.
\Cref{fig:supp-dense} shows correspondence predictions on real images, and \cref{fig:supp-nlf} compares our fits with NLF-L, including failure cases.

\FloatBarrier
\section{Limitations and Future Work}
\label{supp:limitations}
\paragraph{Pose diversity.}
\proteus currently prompts the text-to-image generator for relaxed A-poses and synthesizes posed training data by skinning the registered SMPL-X-D meshes with AMASS~\cite{mahmood2019amass} poses via linear blend skinning.
While this produces articulated geometry across diverse poses, the clothing offsets remain static in canonical space and do not capture pose-dependent deformation such as dynamic creases at bent elbows or knees.

Neither our multi-view annotation, SMPL-X fitting, nor coarse-to-fine registration assumes an A-pose: keypoints are triangulated from multi-view renders and registration poses the template via the fitted skeletal rig.
Replacing unconstrained text-to-image generation with pose-conditioned synthesis (\eg, conditioned on skeleton or normal renders of posed SMPL-X bodies) would allow \proteus to generate assets directly in diverse poses without modifying downstream stages.
This would provide direct supervision for pose-dependent cloth deformation and further improve posed point-cloud and image tasks.
\paragraph{Garment topology.}
Representing the clothed human as a single surface in the twice-subdivided SMPL-X topology ensures universal vertex correspondence and direct skeletal skinning across all samples, covering garments that broadly follow the body's topology (from fitted tops to loose trousers and jackets).

However, garments whose topology departs from the body, such as skirts, dresses, and long open coats, span the space between the legs or hang freely away from the torso, causing a single body-aligned surface to either bridge across gaps or miss the garment boundary.
There is no limitation on the 2D or 3D generator side: such outfits can be generated in principle by extending the prompt catalog.
Supporting them in \proteus primarily requires pairing the body template with separate, template-anchored garment layers and using our multi-view semantic labels to route surface correspondences to the appropriate layer.
\paragraph{Downstream extensions.}
The downstream models in \cref{sec:application} use compact architectures to demonstrate the breadth of supervision provided by \proteus; each can be extended using modalities already present in our dataset:
\begin{itemize}
    \item \textbf{Textured and text-conditioned generation:} Our UV flow generator currently predicts geometry offsets alone, but can be directly extended to jointly synthesize RGB textures or condition on text prompts using the paired UV textures and structured descriptions in \proteus.
    \item \textbf{Hybrid feed-forward and optimization registration:} Chaining our point-cloud SMPL-X regressor with our feed-forward point-cloud generator yields an end-to-end, optimization-free scan registration pipeline. Its predictions can also serve as a warm start for a few iterations of fine-stage optimization to recover fine high-frequency wrinkles.
    \item \textbf{Fine-grained garment parsing:} Beyond the 29 Sapiens2 semantic classes, the structured prompt metadata in \proteus records fine-grained garment categories and attributes that can supervise open-vocabulary or fine-grained 3D garment parsing.
    \item \textbf{Expressive body and hand tracking:} Extending our point-cloud and dense-correspondence SMPL-X estimators to jointly predict finger and facial parameters, alongside stronger 2D joint detectors, would enable full expressive avatar capture from monocular video or raw scans.
\end{itemize}

\end{document}